\documentclass[leqno,twocolumn]{article}

\usepackage[letterpaper, total={6.5in, 8.5in}]{geometry}

\usepackage[pdftex]{graphicx}
\usepackage[cmex10]{amsmath}
\usepackage{amssymb}
\usepackage{caption,subcaption} % to use subfigures
\usepackage{xcolor} % to have TODO notes in red
\usepackage{bbm}		%indicator function
\usepackage{mathtools} 	%substack

\usepackage{hyperref}

\newtheorem{definition}{Definition}
\newtheorem{theorem}{Theorem}

\newcommand{\ID}{\ensuremath{\mathrm{ID}}}
\newcommand{\IDstar}{\ensuremath{\mathrm{ID}^*}}

\newcommand{\MLE}{\ensuremath{{\mathrm{MLE}}}}
\newcommand{\TLE}{\ensuremath{{\mathrm{TLE}}}}
\newcommand{\TLEplus}{\TLE_{\mathrm{c}}}
\newcommand{\TLEminus}{\TLE}
\newcommand{\TLEplusNR}{\TLE_{\mathrm{c}}^{\mathrm{n}}}
\newcommand{\TLEminusNR}{\TLE^{\mathrm{n}}}

\newcommand{\IDMLE}{\ensuremath{\widehat{\mathrm{ID}}_{\MLE}}}

\newcommand{\IDTLEplus}{\ensuremath{\widehat{\mathrm{ID}}_{\TLEplus}}}
\newcommand{\IDTLEminus}{\ensuremath{\widehat{\mathrm{ID}}_{\TLEminus}}}
\newcommand{\IDTLEplusNR}{\ensuremath{\widehat{\mathrm{ID}}_{\TLEplusNR}}}
\newcommand{\IDTLEminusNR}{\ensuremath{\widehat{\mathrm{ID}}_{\TLEminusNR}}}

\newenvironment{qedproof}{\noindent {\bf Proof:\,\ }}{\hfill\mbox{\ $\Box$}\medskip}

\begin{document}

%\setcounter{chapter}{2} % If you are doing your chapter as chapter one,
%\setcounter{section}{3} % comment these two lines out.

%\title{\Large Intrinsic Dimensionality Estimation
%within Tight Localities\thanks{
\title{\Large Intrinsic Dimensionality Estimation
within Tight Localities: \\ A Theoretical and Experimental Analysis\thanks{
M.\ E.\ H.\
supported by JSPS Kakenhi Kiban (B) Research Grant 18H03296.
K.\ K.\ supported by JST ERATO Kawarabayashi Large Graph Project JPMJER1201 and by JSPS Kakenhi JP18H05291.
M.\ R.\ supported by the Science Fund of the Republic of Serbia, \#6518241, AI -- GRASP.
}}
%\author{Laurent Amsaleg}
\author{
Laurent Amsaleg\thanks{CNRS-IRISA, France.} \\
laurent.amsaleg@irisa.fr
\and
Oussama Chelly\thanks{Amazon Web Services.} \\
ouchelly@amazon.com
\and
Michael E. Houle\thanks{The University of Melbourne, Australia. (Partially conducted while at the National Institute of Informatics, Japan.)} \\
mhoule@unimelb.edu.au
\and
Ken{-}ichi Kawarabayashi\thanks{National Institute of Informatics, Japan.} \\
k\_keniti@nii.ac.jp
\and
Milo\v{s} Radovanovi\'c\thanks{University of Novi Sad, Faculty of Sciences, Serbia.} \\
radacha@dmi.uns.ac.rs
\and
Weeris Treeratanajaru\thanks{Bank of Thailand, Thailand.} \\
weerist@bot.or.th
}
\date{}

\maketitle

% Copyright Statement
% When submitting your final paper to a SIAM proceedings, it is requested that you include
% the appropriate copyright in the footer of the paper.  The copyright added should be
% consistent with the copyright selected on the copyright form submitted with the paper.
% Please note that "20XX" should be changed to the year of the meeting.

% Default Copyright Statement
%\fancyfoot[R]{\scriptsize{Copyright \textcopyright\ 2022 by SIAM\\
%Unauthorized reproduction of this article is prohibited}}

% Depending on which copyright you agree to when you sign the copyright form, the copyright
% can be changed to one of the following after commenting out the default copyright statement
% above.

%\fancyfoot[R]{\scriptsize{Copyright \textcopyright\ 20XX\\
%Copyright for this paper is retained by authors}}

%\fancyfoot[R]{\scriptsize{Copyright \textcopyright\ 20XX\\
%Copyright retained by principal author's organization}}

\pagenumbering{arabic}
\setcounter{page}{1}%Leave this line commented out.

\begin{abstract} \small\baselineskip=9pt
Accurate estimation of Intrinsic Dimensionality (ID) is of crucial
importance in many data mining and machine learning tasks, including
dimensionality reduction, outlier detection, similarity search and subspace
clustering. However, since their convergence generally requires sample sizes (that
is, neighborhood sizes) on the order of hundreds of points, existing ID
estimation methods may have only limited usefulness for applications in
which the data consists of many natural groups of small size. In this
paper, we propose a local ID estimation strategy stable even for `tight'
localities consisting of as few as 20 sample points. The estimator applies
MLE techniques over all available pairwise distances among the members of
the sample, based on a recent extreme-value-theoretic model of intrinsic
dimensionality, the Local Intrinsic Dimension (LID). Our experimental
results show that our proposed estimation technique can achieve notably
smaller variance, while maintaining comparable levels of bias, at much
smaller sample sizes than state-of-the-art estimators.
\end{abstract}

%\keywords{Intrinsic Dimensionality, Estimation, Manifold Learning}

\vspace{-6pt}

%%%%%%%%%%%%%%%%
\section{Introduction}
\label{S:intro}

\vspace{-2pt}

% ID
In high-dimensional contexts
where data is represented by many features,
the performance of data analysis techniques often
greatly depends on the inherent complexity of the
data model. Although this complexity is often taken to
be the number of features themselves (that is, the
`representational dimension' or the `ambient dimension'),
simply counting the number of features does not take
into account the relationships among them:
some features may be redundant, while
others may be irrelevant, while yet others may exhibit
various degrees of dependency. A better measure of the
complexity of the data model is to determine the
intrinsic dimensionality (ID) of the data according to
some natural criterion, such as the number of
latent variables required to describe the data,
or the number of basis vectors needed to describe a
manifold that closely approximates the data set.

% Estimation of ID
Over the decades, many characterizations of intrinsic
dimensionality were proposed, each with
its own estimators~\cite{camastra2016intrinsic}.
%In order to obtain an ID measurement from a given model
%so as to use it in a learning algorithm,
%estimation techniques based on optimization and statistical approaches
%are needed.
Topological models estimate the basis dimension of the tangent space
of the data manifold~\cite{pearson1901,Jolliffe86,bruske1998intrinsic,verveer1995evaluation}.
This class of estimators includes Principal Component Analysis (PCA)
and its variants~\cite{pearson1901,Jolliffe86,KarhunenJ94,tipping1999probabilistic},
and multidimensional scaling
(MDS)~\cite{cox2000multidimensional,sammon1969nonlinear,kruskal1964multidimensional,bennett1969intrinsic,chen1974nonlinear,demartines1997curvilinear,kohonen1995learning}.
Graph-based methods
attempt to preserve the $k$-NN graph
~\cite{conf/SSC/Costa2004}.
Fractal models, popular in physics applications,
are used to estimate the dimension of nonlinear
%systems~\cite{hausdorff1918dimension,ott2002chaos}
systems~\cite{ott2002chaos}
--- these include popular estimators due to
Camastra \& Vinciarelli~\cite{camastra2002estimating},
Fan, Qiao \& Zhang~\cite{fan2009intrinsic},
Grassberger \& Procaccia~\cite{grassberger2004measuring},
Hein \& Audibert~\cite{hein2005intrinsic},
K\'egl~\cite{kegl2002intrinsic},
Raginsky \& Lazebnik~\cite{raginsky2005estimation}
and Takens~\cite{takens1985numerical}.
Statistical estimators such as IDEA~\cite{rozza2011idea}
and DANCo~\cite{ceruti2014danco}
estimate the dimension from the concentration of norms and angles.

% Global vs Local
The aforementioned estimators can be described as `global',
in that they provide a single ID measurement for the full data set,
as opposed to `local' ID estimators that assign a different dimensionality
to each point or region in the data set.
% Local ID estimators
Commonly-used local estimators of ID include:
topological methods that model
dimensionality as that of a locally tangent subspace to a manifold,
such as
PCA-based approaches~\cite{journal/toc/FukunagaO71,%
bruske1998intrinsic,fan2009intrinsic,little2009multiscale,little2016multiscale},
and ID estimation from Expected Simplex Skewness~\cite{johnsson2015low};
Local Multidimensional Scaling methods~\cite{cox2000multidimensional}
such as Isometric Mapping~\cite{isomap},
Locally Linear Embedding~\cite{LLE},
Laplacian and Hessian eigenmaps~\cite{donoho2003hessian,belkin2003laplacian},
and Brand's Method~\cite{brand2002charting};
distance-based measures such as the Expansion Dimension (ED),
that assess the rate of expansion of the neighborhood size
with increasing radius~\cite{KargerR02,ged,he2014intrinsic},
as well as other probabilistic methods
that view the data as a sample from a hidden distance distribution, such as
the Hill estimator~\cite{Hill75},
the Manifold-Adaptive Dimension~\cite{farahmand2007manifold},
Levina and Bickel's algorithm~\cite{mleid},
the minimum neighbor distance (MiND) framework~\cite{RozzaLCCC12},
and the local intrinsic dimensionality (LID)
framework~\cite{dami2018estimation,LID1sisap17}.

% Local applications
Distance-based estimators
of local ID infer the data dimensionality solely
from the distribution of distances from
a reference point to its nearest neighbors;
they generally do not require any assumptions
as to whether the data can be modeled as a manifold.
%In addition to applications in manifold learning,
This convenience allows distance-based measures of local ID to
be used in the context of similarity search,
where they are used to assess the complexity of a search query
~\cite{KargerR02},
to control the early termination of search~\cite{CasanovaEHKNSZ17,xiguo,conf/sisap/HouleMOS14}, or the interleaving of feature sparsification with the construction
of a neighborhood graph~\cite{LID2sisap17,HouleOW17}.
Distance-based measures have also found applications in
deep neural network classification~\cite{MaWanHouZhoetal18};
the characterization and detection
of adversarial attacks in learning~\cite{AmsalegBBEFHRN21,MaLiWanErfetal18,WeerasingheAAEL21,WeerasingheAEL21}; explanability in deep learning models~\cite{Jia0RLH19}; subspace alignment in generative deep learning models~\cite{BaruaMEHB19,LiQZMR19};
and outlier detection~\cite{HouleSZ18,VriesCH12,WangELH21}.
%in the analysis of a projection-based heuristic~\cite{VriesCH12},
%and in the estimation of
%local density~.
The efficiency and effectiveness of the algorithmic applications
of local intrinsic dimensional estimation
(such as~\cite{CasanovaEHKNSZ17,xiguo,conf/sisap/HouleMOS14,HouleOW17}) depends
heavily on the quality of the estimators employed,
and on their ability to provide a trustworthy estimate
despite the limited number of available samples.
Distance-based local estimators are well-suited for
many of the applications in question,
since nearest neighbor distances are often precomputed
and are thus readily available for use in estimation.

% Flaws in the sota: slow convergence, locality
Local estimators of ID can potentially have significant impact
when used in subspace outlier detection, subspace clustering, or other
applications in which the intrinsic dimensionality is assumed
to vary from location to location.
However, in practical settings, the localities assumed in data analysis
are often too `tight' to provide the number of
neighborhood samples needed by current estimators of ID.
State-of-the-art outlier detection algorithms, for example, typically
use neighborhoods of 20 or fewer data points as localities within which
to assess the outlierness of test examples~\cite{CamposZSCMSAH16},
whereas estimators of local intrinsic dimension generally require sample sizes
in excess of 100 in order to converge~\cite{dami2018estimation}.
Simply choosing a number of samples sufficient for the convergence
of the estimator would very often lead to a violation of the
locality constraint,
and to ID estimates that are consequently less reliable,
as the samples would consist of points from several different natural
data groups with different local intrinsic dimensional characteristics.

% Global estimators used locally
% Clipping
Global estimators are sometimes adapted for local estimation of ID
simply by applying them to the subset of the data lying
within some region surrounding a point of interest.
Global methods such as PCA generally make use of most (if not all)
of the (quadratic) pairwise relationships within the data,
giving them an apparent advantage over expansion-based local estimators,
which use only the (linear) number of distances from the reference point
to each sample point.
However, embedding-based and projective-based ID estimators can be
very sensitive to noise; moreover, they can be greatly misled when the
chosen region is larger than the targeted locality, or when the data
distribution does not conform to a linear manifold~\cite{dami2018estimation}.
Also, as we shall argue in Section~\ref{S:skewed},
`clipping' of the data set to a region
can also give rise to boundary effects that have the potential
for extreme bias when estimating ID, whenever the region shape is not
properly accounted for in the ID model or estimation strategy.
With these issues in mind,
locally-restricted application of global ID estimation
should not automatically be regarded as valid for local ID estimation.
On the other hand, expansion-based estimators of local ID
can be seen to avoid clipping bias, since they model
(explicitly or implicitly)
the restriction of the distribution of distances to
a fixed-radius neighborhood centered at
the reference point~\cite{dami2018estimation}.

%Global estimators assume (implicitly or explicitly) that the data set is a
%representative sample from some underlying data distribution, possibly
%but not necessarily conforming to a given manifold.
%Unlike subsampling from the full data set, which would produce a
%smaller sample consistent with the original distribution,
%clipping the data with respect to a local region would produce a
%smaller sample that would not necessarily be consistent with the
%original distribution local region.
%
%Indeed, global distance-based correlation dimension
%(CD)~\cite{takens1985numerical}, if restricted to a neighborhood, would
%use all pairwise distances within the neighborhood to achieve its estimate.
%Although for a given neighborhood size this local use of CD would
%be expected to converge much faster than true local ID estimators,
%the result would be biased due to the clipping effect.

%
As the sizes of the natural groupings (clusters) of the data set
are generally not known in advance,
in order to ensure that the majority of the points are drawn from the same
local distribution, it is highly desirable to use
local estimators that can cope with
the smallest possible sample sizes~\cite{dami2018estimation,RozzaLCCC12}.
%\todo{Moreover, estimating ID often requires more neighbors
%than the aforementioned applications typically need.
%Hence, the use of ID estimates in these applications
%usually requires the computation of larger neighborhoods,
%resulting in a higher computational cost.}
%Thus, the development of local ID estimators with faster convergence
%properties is an important requirement for the effectiveness and
%the efficiency of subspace-based applications that make use of
%highly restricted localities.
One possible strategy for improving the convergence properties of
estimation without violating locality is to draw more measurements
from smaller data samples. For the case of distance-based
local estimation from a neighborhood sample of size $k$,
this would require the use of distances between pairs of neighbors
(potentially quadratic in $k$), and not merely the distances
from the reference point to its neighbors (linear in $k$).
This presents a challenge in that any additional measurements must
be used in a way that locally conforms to the underlying data distribution
without introducing bias due to clipping.

% pairwise distance strategy
In this paper, for distributions of data in real Hilbert spaces,
we develop an effective estimator of local intrinsic
dimension suitable for use in tight localities --- that is,
within neighborhoods of small size that are often employed in
such applications as outlier detection and nearest-neighbor classification.
Given a sample of $k$ points drawn from some target locality (generated
by restricting the data set to a spherical region of
radius $r$ centered at $\mathbf{q}$),
our estimator can be regarded as an aggregation of $2k$ expansion-based
LID estimation processes, each taking either a distinct
sample point $\mathbf{v}$ or its symmetric reflection relative to $\mathbf{q}$ (that is,
the point $2\mathbf{q}-\mathbf{v}$) as the origin of its expansion.
To ensure that these processes use only such information that is
available within the locality, the expansion processes are skewed,
in that their centers are allowed to shift gradually towards $\mathbf{q}$
as the radius of expansion approaches $r$.
Under the modeling assumption that the underlying local
intrinsic dimensionality is uniform throughout the region,
an estimator resulting from the
aggregation of $2k$ skewed expansion processes will be shown to use
all $O(k^2)$ pairwise distances within the sample,
without introducing clipping bias.

The main contributions of this paper include:
\vspace{-5pt}
\begin{itemize}
\item
an explanation of the clipping effect in skewed expansion-based
local ID estimation, and an illustration of its impact on bias;
\vspace{-5pt}
\item
a full description of the proposed LID-based estimator for tight localities,
as well as a theoretical justification under the assumption of continuity
of local ID;
\vspace{-5pt}
\item
an experimental investigation showing that our proposed tight local
estimation technique can achieve notably smaller variance, while
maintaining comparable levels of bias, at much smaller sample sizes
than state-of-the-art estimators;
\vspace{-5pt}
\item
experimental evidence that tight local estimation is more robust than standard expansion-based estimators when applied within the neighborhood of an outlier point.
\end{itemize}
A preliminary version of this paper appeared in~\cite{AmsalegCHKRT19}.
In this expanded version, we provide a generalized and complete
derivation of our estimator, including a theoretical justification of convergence, as well as an extended experimental analysis.
This derivation also corrects a presentational error in the statement
of the estimator in~\cite{AmsalegCHKRT19};
this error did not affect the experimental results,
as a correct implementation of the estimator was used throughout.

% Summary
The remainder of the paper is structured as follows.
In the next section,
we review the LID model of local
intrinsic dimensionality~\cite{LID1sisap17,LID2sisap17}
and its estimators~\cite{dami2018estimation}.
In Section~\ref{S:skewed}, we discuss the issue of the reuse
of neighborhood samples for the estimation of expansion-based
local intrinsic dimensionality. We also show how clipping bias
can result when the expansion originates at a point which is not
the center of the neighborhood.
In Section~\ref{S:tight}
we introduce our proposed tight LID-based estimator together with
its justification under the assumption of continuity of local ID.
In Section~\ref{S:framework},
we provide the details of our experimental framework,
and in Section~\ref{S:res}
we present an experimental comparison of our proposed estimator
with existing local and global ID estimators.
We conclude the discussion in Section~\ref{S:conclusion}.

%%%%%%%%%%%%%%%%%%%%%%%%%
\section{LID and Extreme Value Theory}
\label{S:LID}

In the theory of intrinsic dimensionality, classical
expansion models (such as the expansion dimension and generalized expansion dimension~\cite{KargerR02,ged}) measure the rate of growth in the number of data objects encountered as the distance from the reference sample increases. As an intuitive example, in Euclidean space, the volume of an $m$-dimensional ball grows proportionally to $r^m$, when its size is scaled by a factor of $r$. From this rate of volume growth with distance, the expansion dimension $m$ can be deduced as:
\begin{equation}
\frac{V_2}{V_1} = \left( \frac{r_2}{r_1} \right)^m \Rightarrow m = \frac{\ln(V_2/V_1)}{\ln(r_2/r_1)}.
\label{eq:expansion_example}
\end{equation}

\noindent By treating probability mass as a proxy for volume, classical expansion models provide a \textit{local view} of the dimensional structure of the data, as their estimation is restricted to a neighborhood around the sample of interest. Transferring the concept of expansion dimension to the statistical setting of continuous distance distributions leads to the formal definition of LID \cite{LID1sisap17}.

\begin{definition}[Local Intrinsic Dimensionality] \quad \\[-3pt]
Given a data sample $\mathbf{x} \in X$, let $R>0$ be a random variable denoting the distance from $\mathbf{x}$ to other data samples. If the c.d.f.\ $F(r)$ of $R$ is positive and continuously differentiable at distance $r>0$,
the LID of $\mathbf{x}$ at distance $r$ is given by:
\begin{equation} \label{eq:LID_r}
% \begin{split}
    \mathrm{ID}_F(r)
% &
\triangleq \lim_{\epsilon\to 0} \frac{\ln\big(F((1+\epsilon)\cdot r)\big/F(r)\big)}{\ln(1+\epsilon)} = \frac{r\cdot F'(r)}{F(r)},
% \end{split}
\end{equation}
whenever the limit exists.
\label{def:lid}
\end{definition}

\vspace{-3pt}
\noindent $F(r)$ is analogous to the volume $V$ in Equation~\eqref{eq:expansion_example}; however, we note that the underlying distance measure need not be Euclidean. The last equality of Equation~\eqref{eq:LID_r} follows by applying L'H\^{o}pital's rule to the limits~\cite{LID1sisap17}.
The local intrinsic dimension at $\mathbf{x}$ is in turn defined as the limit when the radius $r$ tends to zero:
\begin{equation} \label{eq:LID}
    \text{ID}^{*}_F \triangleq \lim_{r \to 0}  \text{ID}_F(r).
\end{equation}

$\text{ID}^{*}_F$ describes the relative rate at which the function $F(r)$ increases as the distance $r$ increases from $0$, and can be estimated using the distances of $\mathbf{x}$ to its $k$ nearest neighbors within the sample~\cite{dami2018estimation}.

%In the ideal case where the data in the vicinity of $\mathbf{x}$ is distributed uniformly within a submanifold, $\text{ID}^{*}_F$ equals the dimension of the submanifold; however, in general these distributions are not ideal, the manifold model of data does not perfectly apply, and $\text{ID}^{*}_F$ is not an integer. Nevertheless, the local intrinsic dimensionality does give a rough indication of the dimension of the submanifold containing $\mathbf{x}$ that would best fit the data distribution in the vicinity of $\mathbf{x}$. We refer readers to \cite{LID1sisap17,LID2sisap17} for more details concerning the LID model.

\textbf{Estimation of LID:}
In accordance with the statistic theory of extreme values, the smallest $k$ nearest neighbor distances could be regarded as extreme events associated with the lower tail of the underlying distance distribution. Under very reasonable assumptions, the tails of continuous probability distributions converge to the Generalized Pareto Distribution (GPD), a form of power-law distribution~\cite{book/Coles01}.
From this, \cite{dami2018estimation} developed several estimators of LID to heuristically approximate the true underlying distance distribution by a transformed GPD; among these, the Maximum Likelihood Estimator (MLE) --- which coincides with the Hill estimator~\cite{Hill75} for the scale parameter of untransformed GPDs, and the estimator of intrinsic dimensionality derived by Levina and Bickel~\cite{mleid} based on Poisson point processes --- exhibited a useful trade-off between statistical efficiency and complexity. Given a reference sample $\mathbf{x} \sim \mathcal{P}$, where $\mathcal{P}$ represents the global data distribution, the MLE estimator of the LID at $\mathbf{x}$ is:
\begin{equation} \label{eq:estimator}
\widehat{\mathrm{ID}}_{\mathrm{MLE}}(\mathbf{x})
=
- \Bigg( \frac{1}{k}\sum_{i=1}^{k}\ln \frac{r_i(\mathbf{x})}{r_k(\mathbf{x})}\Bigg)^{-1}.
\end{equation}

\noindent Here, $r_i(\mathbf{x})$ denotes the distance between $\mathbf{x}$ and its $i$-th nearest neighbor within a sample of points drawn from $\mathcal{P}$, where $r_k(\mathbf{x})$ is the maximum of the neighbor distances. In practice, the sample set is drawn uniformly at random from the available training data (omitting $\mathbf{x}$ itself), which itself is presumed to have been randomly drawn from $\mathcal{P}$.
%%% Now that the paper has been accepted, we no longer need the
%%% following sentence. :)  -MEH
%We emphasize that the LID defined in Equation~\eqref{eq:LID} is a \textit{theoretical} quantity, and that $\widehat{\mathrm{ID}}_{\mathrm{MLE}}$ is its \textit{estimator}.

%As described in the previous subsection, the GPD
%asymptotically models the lower tail of the distance distributions
%within a threshold distance $w$. The GPD takes explicit account
%of the effect of restricting the domain to the lower tail of
%the global distribution from which it derives, and thus
%any estimator of the scale parameter of the GPD would not
%need to consider sample points that do not lie within the neighborhood.
%This implies that valid LID estimators (including the Hill estimator) do not
%suffer from bias due to clipping when applied to the neighborhoods of
%reference points with a global distribution.

\section{LID Estimation and the Clipping Effect}
\label{S:skewed}

In practice, the LID model is typically applied to the c.d.f.\ $F$ induced by some global distribution of data with respect to a reference location $\mathbf{q}$. In the ideal case where the data in the vicinity of $\mathbf{q}$ is distributed uniformly within a subspace (or manifold), $\IDstar_F$ equals the dimension of the subspace. In general, however, these distributions are not ideal, the subspace model of data does not perfectly apply, and $\IDstar_F$ is not necessarily an integer. Instead, by characterizing the growth rate of the distribution of distances from $\mathbf{q}$, it naturally takes into account the effect of variation within the subspace, and error relative to the subspace, all in one value. Nevertheless, the local intrinsic dimensionality does give some indication of the dimension of the subspace containing $\mathbf{q}$ that would best fit the data distribution in the vicinity of $\mathbf{q}$, provided that the distribution of distances to $\mathbf{q}$ is smooth. We refer readers to \cite{LID1sisap17,LID2sisap17} for more details concerning the LID model.

From the global perspective, it is not necessarily the case that
$\IDstar$ exists for every possible reference location in the domain.
However, if the distribution is in some sense smooth in the vicinity
of $\mathbf{q}$, it is reasonable to assume that for some point $\mathbf{v}$ sufficiently
close to $\mathbf{q}$, the underlying value of $\IDstar_{F_{\mathbf{v}}}$
could be a close approximation of $\IDstar_F$, where $F_{\mathbf{v}}$ is the
c.d.f.\ for the distance distribution induced relative to $\mathbf{v}$.
We therefore adopt the following definition of
the continuity of LID.
%the continuity of LID,
%recently proposed for the analysis of adversarial perturbation
%on nearest-neighbor classification~\cite{AmsalegBBEHNR17}:

\vspace{-2pt}
\begin{definition}
\label{D:continuousID}
The local intrinsic dimensionality will be said to be
{\em (uniformly) continuous}
at $\mathbf{q}\in\mathcal{S}$ if the following conditions hold:
\begin{enumerate}
\item
There exists a distance $\rho>0$ for which all points
$\mathbf{v}\in\mathcal{S}$ with $\|\mathbf{v}-\mathbf{q}\|\leq\rho$
admit a distance distribution whose c.d.f.\ $F_{\mathbf{v}}$ is continuously differentiable and positive within
some open interval with lower bound $0$.
\item
For each $\mathbf{v}$ satisfying Condition 1,
$\IDstar_{F_{\mathbf{v}}}$ exists.
\item
$\lim_{s\to 0} \IDstar_{F_{\psi(s)}}$
converges uniformly to $\IDstar_{F_{\mathbf{q}}}$,
where $\psi(s)=s\mathbf{v}+(1-s)\mathbf{q}$
interpolates $\mathbf{q}$ and $\mathbf{v}$.
\end{enumerate}
\end{definition}
\vspace{-2pt}

For the data model underlying our proposed estimator, we will
assume that the local intrinsic dimensionality is continuous in
the vicinity of the test point $\mathbf{q}$. Under the assumption of
continuity, the estimator will use estimates of $\IDstar_{F_{\mathbf{v}}}$
for points $\mathbf{v}$ close
to $\mathbf{q}$ to help stabilize the estimate of $\IDstar_F$.

\begin{figure}
\centering%
\includegraphics[width=.35\linewidth,trim={11cm 15.5cm 15 115},clip]{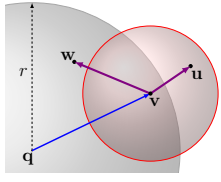}
\caption{When estimating the local ID at $\mathbf{v}$ to support
the estimation of the local ID at $\mathbf{q}$,
using the sample $\mathbf{u}$ violates the locality restriction
for $\mathbf{q}$, but ignoring it in favor of $\mathbf{w}$ can lead to
clipping bias.}
\label{F:clipping}
\end{figure}

However, straightforwardly estimating and aggregating values of
$\IDstar_{F_{\mathbf{v}}}$ over all neighbors $\mathbf{v}$ of $\mathbf{q}$
can give rise either to clipping bias, or a violation of locality, or both.
To see this, consider the situation shown in Figure~\ref{F:clipping},
in which we have a sample $V=\{v_i|\,1\leq i\leq k\}$
of points drawn from the
restriction of the global distribution to a neighborhood
$B(\mathbf{q},r)$ of radius $r$.
For a given neighbor $\mathbf{v}\in V$, many if not most of its own neighbors
may lie well outside the vicinity of $\mathbf{q}$.
If the estimation makes use of a neighbor $\mathbf{u}$ external to
$B(\mathbf{q},r)$, then locality is violated, which can have drastic
consequences for any task that makes use of the estimates.

On the other hand, if the locality of $\mathbf{q}$ is not to be violated,
then straightforward use of the sample set $V$ to estimate the
characteristics of the distribution of distances from $\mathbf{v}$
would suffer from clipping bias --- close neighbors of $\mathbf{v}$
(such as $\mathbf{u}$ in Figure~\ref{F:clipping})
will have effectively been replaced by other members of~$V$
(e.g.\ $\mathbf{w}$) that are farther from $\mathbf{v}$.
As will be shown in Section~\ref{S:res}, for LID this typically
has the effect of introducing a strong negative bias
on the estimated values.

%%%%%%%%%%%%%%%%%%%%%%%%%
\section{Tight LID Estimation}
\label{S:tight}

For our estimator, which we will refer to as
$\widehat{\mathrm{ID}}_\mathrm{TLE}$,
we limit the sample points to those points of
the data set that lie within a tight neighborhood of the test point
$\mathbf{q}$.
In order to avoid clipping bias, we adjust the distributions
of distances computed from a nearby point $\mathbf{x}$
by taking advantage of
the assumption of uniform continuity of local ID, as
laid out in Definition~\ref{D:continuousID}.

\subsection{LID Estimation from Moving Centers.}
\label{SS:moving}

\begin{figure}
\centering%
\includegraphics[width=.45\columnwidth]{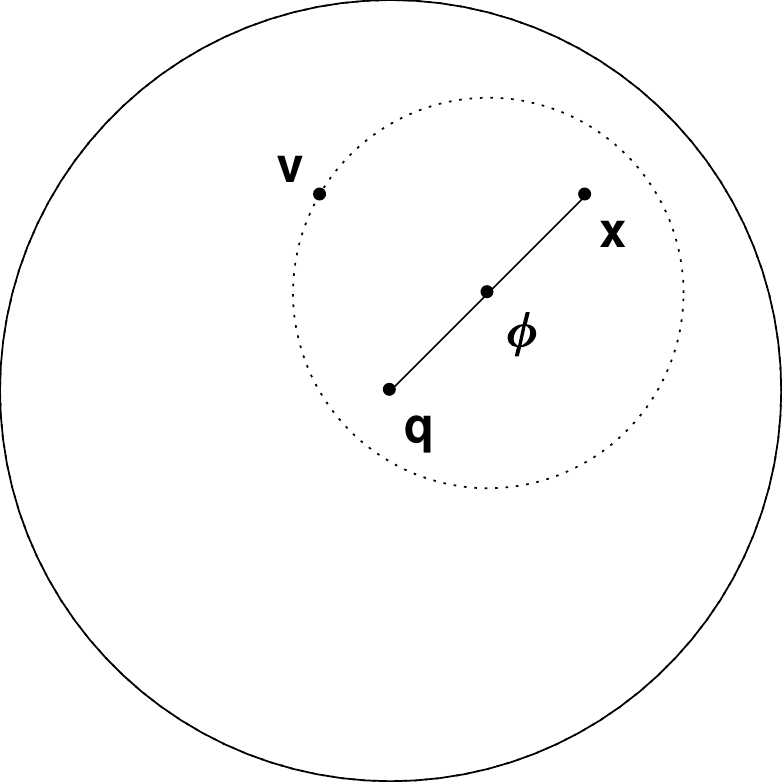}
\caption{Example positions of $\mathbf{q}$, $\mathbf{x}$, $\phi$,
and another sample point $\mathbf{x}$ encountered by the expanding
circle centered at~$\phi$.}
\label{F:inside-with-v}
\end{figure}

Let $r$ be the radius of the neighborhood $V,$ and let $\mathbf{x}$
be a point within distance $r$ of $\mathbf{q}$.
The distribution of distances based at $\mathbf{x}$
is generated through a smooth interpolative process
involving an expanding circle whose
center is smoothly transformed from $\mathbf{x}$ to $\mathbf{q}$
as its radius is increased from $0$ to $r$.
The radii of these circles, together with the probability measure
associated with their interiors, determine a distribution of distance
values.
More formally,
if $r$ is the radius of the neighborhood $V$,
the point $\mathbf{x}$
can be associated with a distribution
whose
c.d.f.\ $F_{\mathbf{q},\mathbf{x},r}$ is defined as
\begin{eqnarray*}
\phi_{\mathbf{q},\mathbf{x},r}(t)
& \triangleq &
(t/r)\cdot\mathbf{q}+(1-t/r)\cdot\mathbf{x}
%\mbox{,~~and}
\\
F_{\mathbf{q},\mathbf{x},r}(t)
& \triangleq &
F_{\phi_{\mathbf{q},\mathbf{x},r}(t)}(t),
\end{eqnarray*}
where the interpolated point $\phi_{\mathbf{q},\mathbf{x},r}(t)$ is defined
over the range $t\in[0,r]$, and $F_{\phi_{\mathbf{q},\mathbf{x},r}(t)}$ is
the c.d.f.\ of the distribution of distances from
$\phi_{\mathbf{q},\mathbf{x},r}(t)$. For any $t\in[0,r]$, the value
$F_{\phi_{\mathbf{q},\mathbf{x},r}(t)}$ is the probability of a sample point
lying inside the unique circle with center
$\phi_{\mathbf{q},\mathbf{x},r}(t)$ and radius $t$.
Figure~\ref{F:inside-with-v} illustrates this setting.

\begin{theorem}
\label{T:interpolation}
If the local intrinsic dimensionality $\IDstar_{F_{\mathbf{q}}}$
is uniformly continuous, then
$\IDstar_{F_{\mathbf{q},\mathbf{x},r}} = \IDstar_{F_{\mathbf{q}}}$.
\end{theorem}

\begin{qedproof}
Under the assumption of continuity,
there exists $\rho>0$ such that for any $0\leq s\leq\rho$,
the Moore-Osgood theorem implies that:
\begin{eqnarray*}
\IDstar_{F_{\mathbf{q}}}
& = &
\lim_{t\to r^{-}}
\IDstar_{F_{\phi_{\mathbf{q},\mathbf{x},r}(t)}}
\\
& = &
\lim_{t\to r^{-}}
\lim_{s\to 0^{+}}
\ID_{F_{\phi_{\mathbf{q},\mathbf{x},r}(t)}}(s)
\\
& = &
\lim_{s\to 0^{+}}
\lim_{t\to r^{-}}
\ID_{F_{\phi_{\mathbf{q},\mathbf{x},r}(t)}}(s)
\, .
%\\
\end{eqnarray*}
The inner limit can be replaced by a subsequence
of values of $\ID_{F_{\phi_{\mathbf{q},\mathbf{x},r}(t)}}(s)$,
for choices of $t$ that tend to $r$ as $s$ tends to zero.
Here, we simply choose $t=r-s$.
Noting that $\phi_{\mathbf{q},\mathbf{x},r}(r)=\mathbf{q}$,
we obtain
\begin{eqnarray*}
\IDstar_{F_{\mathbf{q}}}
& = &
\lim_{s\to 0^{+}}
\ID_{F_{\phi_{\mathbf{q},\mathbf{x},r}(r-s)}}(s)
%\\
%& = &
%\lim_{s\to 0^{+}}
%\ID_{F_{\mathbf{q},\mathbf{x},r}}(0)
%\\
%& = &
\:\: = \:\:
\IDstar_{F_{\mathbf{q},\mathbf{x},r}}
\, .
\end{eqnarray*}
\end{qedproof}

Under the assumption of continuity, the local ID at $q$ can therefore
be estimated from the distribution $F_{\mathbf{q},\mathbf{x},r}$ for
any location $\mathbf{x}$ falling within a sufficiently small neighborhood
of $\mathbf{q}$.  For the purpose of estimation,
the distance value associated with a sample point $\mathbf{v}\in V$
is determined by the radius of the expanding circle at the time
its boundary encounters $\mathbf{v}$, and not the actual distance
from $\mathbf{x}$ to $\mathbf{v}$.
This distance --- which we will denote by
$d_{\mathbf{q},r}(\mathbf{x},\mathbf{v})$ ---
is given by the value $t$ such that
\[
d_{\mathbf{q},r}(\mathbf{x},\mathbf{v})
\:
\triangleq
\:
t
\:
=
\:
\| \phi_{\mathbf{q},\mathbf{x},r}(t) - \mathbf{v} \|.
\]

We observe that when $\mathbf{v}=\mathbf{x}$, this expression reduces
to $t=t\|\mathbf{q}-\mathbf{x}\|/r$, in which case either $t=0$, or
$\|\mathbf{q}-\mathbf{x}\|=r$ and $t$ is indeterminate.
For this reason, we deem
$d_{\mathbf{q},r}(\mathbf{x},\mathbf{v})$
to be zero whenever $\mathbf{v}=\mathbf{x}$.
Similarly, when $\mathbf{q}=\mathbf{x}$, the expression reduces
to $t=\|\mathbf{q}-\mathbf{v}\|$, the distance that would be achieved
if the expansion center were stationary at $\mathbf{q}$.
When $\mathbf{v}=\mathbf{q}$, the expression yields
$t= (1-t/r) \|\mathbf{q}-\mathbf{x}\| $, in which case
\[
t
=
\frac{
r \|\mathbf{q}-\mathbf{x}\|
}{
r + \|\mathbf{q}-\mathbf{x}\|
}
\, .
\]
The remaining cases are accounted for by the following theorem.

\begin{theorem}
\label{T:adjusteddist}
Assume that $\mathbf{q}$, $\mathbf{x}$ and $\mathbf{v}$ are points where
$\mathbf{x}\neq\mathbf{q}$ and $\mathbf{x}\neq\mathbf{v}$.
Furthermore, assume that $0<\|\mathbf{q}-\mathbf{x}\|\leq r$
and $0\leq\|\mathbf{q}-\mathbf{v}\|\leq r$ for some positive radius $r>0$.
If $\|\mathbf{q}-\mathbf{x}\| = r$, then
\[
d_{\mathbf{q},r}(\mathbf{x},\mathbf{v})
=
\frac
{
r
\,
(
\mathbf{v} - \mathbf{x}
)
\cdot
(
\mathbf{v} - \mathbf{x}
)
}
{
2
\,
(
\mathbf{q} - \mathbf{x}
)
\cdot
(
\mathbf{v} - \mathbf{x}
)
}
\, .
\]
Otherwise, if $\|\mathbf{q}-\mathbf{x}\| < r$, then
\[
d_{\mathbf{q},r}(\mathbf{x},\mathbf{v})
=
\sqrt{
\left(
\mathbf{u}
\cdot
(
\mathbf{q} - \mathbf{x}
)
\right)^2
+
r
\mathbf{u}
\cdot
(
\mathbf{v} - \mathbf{x}
)
}
-
\mathbf{u}
\cdot
(
\mathbf{q} - \mathbf{x}
)
\, ,
\]
where
\[
\mathbf{u}
\triangleq
\frac{
r
(
\mathbf{v} - \mathbf{x}
)
}{
%\left(
r^2
-
\|
\mathbf{q} - \mathbf{x}
\|^2
%\right)
}
\, .
\]
\end{theorem}

\begin{qedproof}
From the definition, we can consider $t$ such that
\[
t
=
\| \phi_{\mathbf{q},\mathbf{x},r}(t) - \mathbf{v} \|
\, .
\]
Then
\begin{eqnarray*}
t
& = &
\|
(
\phi_{\mathbf{q},\mathbf{x},r}(t)
-
\mathbf{x}
)
+
(
\mathbf{x} - \mathbf{v}
)
\|
\\
& = &
\left\|
\frac{t}{r}
(
\mathbf{q} - \mathbf{x}
)
-
(
\mathbf{v} - \mathbf{x}
)
\right\|
\, .
\end{eqnarray*}
Squaring, we obtain
\begin{eqnarray*}
t^2
& = &
\frac{t^2}{r^2}
\|
\mathbf{q} - \mathbf{x}
\|^2
-
2\frac{t}{r}
(
\mathbf{q} - \mathbf{x}
)
\cdot
(
\mathbf{v} - \mathbf{x}
)
+
\|
\mathbf{v} - \mathbf{x}
\|^2
\, .
\end{eqnarray*}
Multiplying by $r^2$ and unifying the terms with factor $t^2$,
and then letting $z = r^2-\|\mathbf{q} - \mathbf{x}\|^2$, gives
\begin{eqnarray*}
0
& = &
z
t^2
+
2rt
(
\mathbf{q} - \mathbf{x}
)
\cdot
(
\mathbf{v} - \mathbf{x}
)
-
r^2
\|
\mathbf{v} - \mathbf{x}
\|^2
\, .
\end{eqnarray*}
If $r=\|\mathbf{q} - \mathbf{x}\|$, then $z=0$, and
\begin{eqnarray*}
t
& = &
\frac
{
r
\,
(
\mathbf{v} - \mathbf{x}
)
\cdot
(
\mathbf{v} - \mathbf{x}
)
}
{
2
\,
(
\mathbf{q} - \mathbf{x}
)
\cdot
(
\mathbf{v} - \mathbf{x}
)
}
\, .
\end{eqnarray*}
Otherwise, if $r>\|\mathbf{q} - \mathbf{x}\|$, then $z>0$. Dividing
through by $z$, and substituting $\mathbf{u}$ for $(\mathbf{v}-\mathbf{x})r/z$,
yields
\begin{eqnarray*}
0
& = &
t^2
+
2t
\mathbf{u}
\cdot
(
\mathbf{q} - \mathbf{x}
)
-
r
\mathbf{u}
\cdot
(
\mathbf{v} - \mathbf{x}
)
\, .
\end{eqnarray*}
Solving for the non-negative root and then simplifying, we obtain
\begin{eqnarray*}
t
& = &
\sqrt{
\left(
\mathbf{u}
\cdot
(
\mathbf{q} - \mathbf{x}
)
\right)^2
+
r
\mathbf{u}
\cdot
(
\mathbf{v} - \mathbf{x}
)
}
-
\mathbf{u}
\cdot
(
\mathbf{q} - \mathbf{x}
)
\, .
\end{eqnarray*}
\end{qedproof}

\subsection{MLE Estimation from Multiple Centers.}
\label{SS:multiple}

The MLE estimator for a single moving center is obtained
by using Equation~\ref{eq:estimator}
with adjusted distances of the form
$d_{\mathbf{q},r}(\mathbf{x},\mathbf{v})$,
for all samples $\mathbf{v}\in V\setminus\{\mathbf{x}\}$.

\begin{figure}
\centering%
\includegraphics[width=.429\columnwidth]{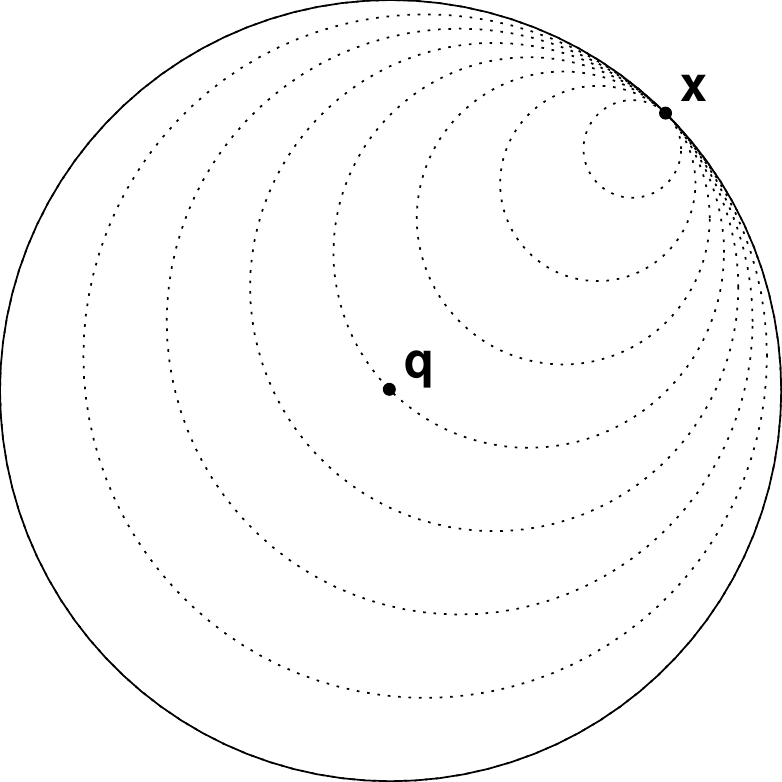}
\includegraphics[width=.429\columnwidth]{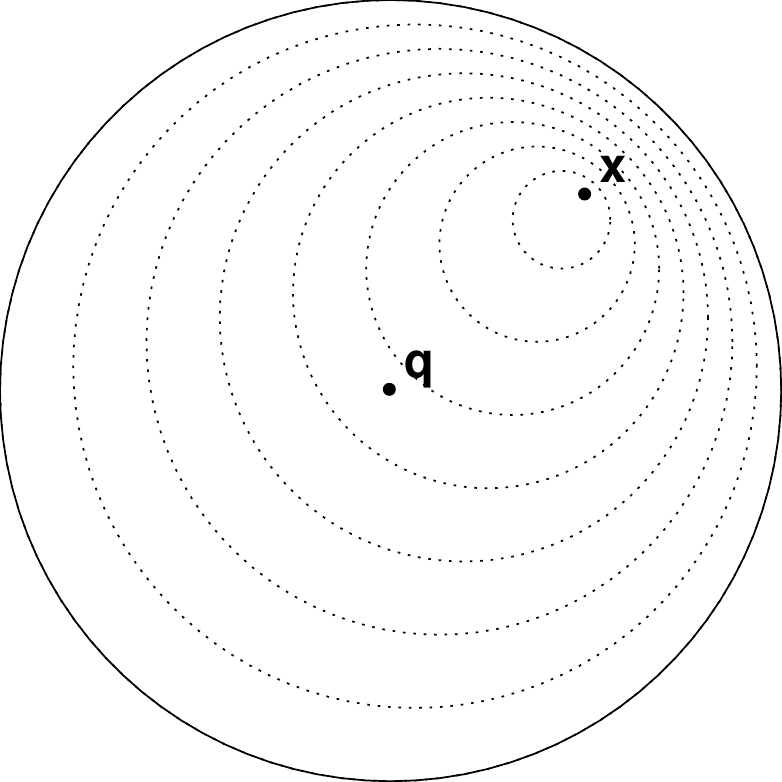}
\caption{Choosing point $\mathbf{x}$ from which to initiate moving centers
from the neighborhood samples $V$. Left: $\mathbf{x}$ on the
border of the neighborhood; Right: $\mathbf{x}$ inside the
neighborhood.}
\label{F:border+inside}
\end{figure}

\begin{figure}
\centering%
\includegraphics[width=.429\columnwidth]{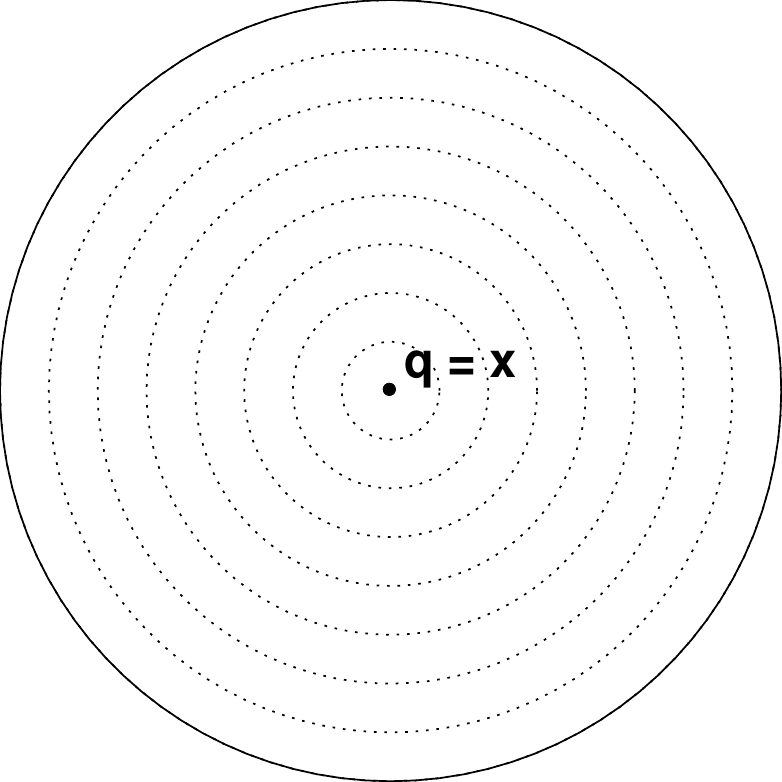}
\caption{Choosing point $\mathbf{x}$ from which to initiate moving centers:
$\mathbf{x} = \mathbf{q}$.}
\label{F:central}
\end{figure}

\begin{figure}
\centering%
\includegraphics[width=.43\columnwidth]{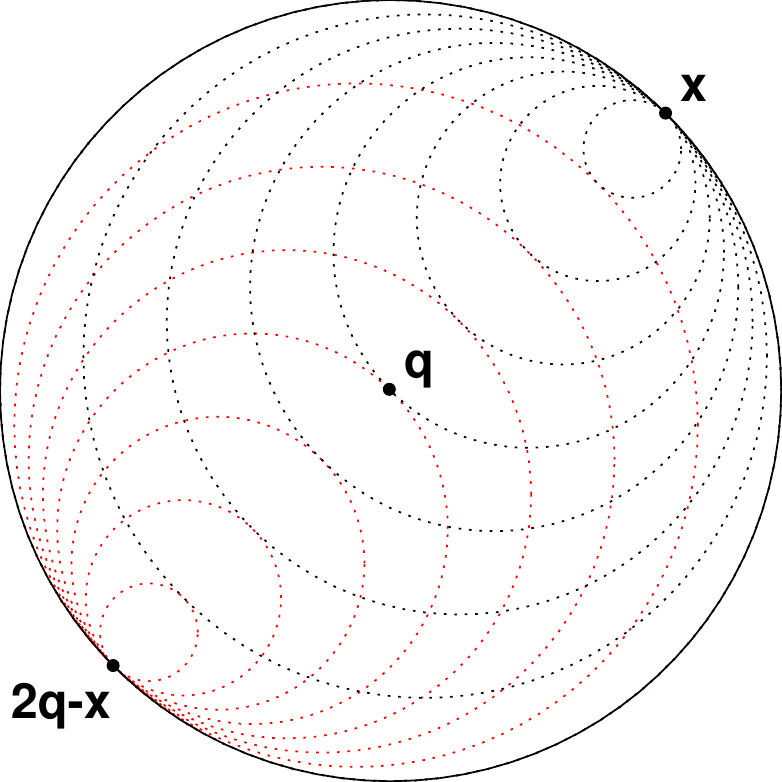}
\includegraphics[width=.43\columnwidth]{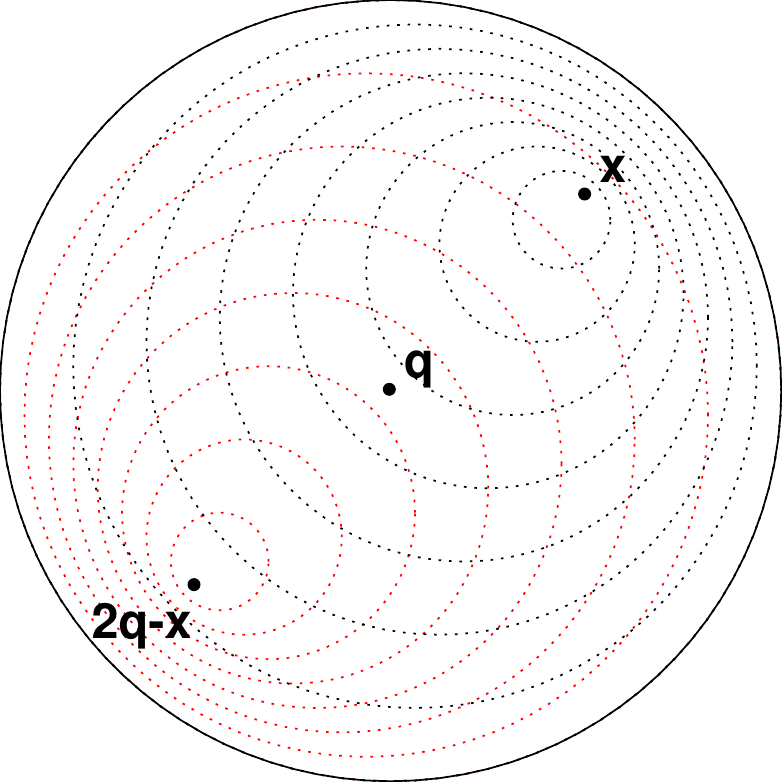}
\caption{Choosing point $\mathbf{x}$ from which to initiate moving centers
from the neighborhood samples $V$, and from their reflections through $\mathbf{q}$. Left: $\mathbf{x}$
on the border of the neighborhood; Right: $\mathbf{x}$ inside the
neighborhood.}
\label{F:border+inside-with-mirror}
\end{figure}

There are many possible ways of choosing locations $\mathbf{x}$ from which to
initiate a moving center for LID estimation. Here, we make use of candidates
of the following three forms:
%\begin{enumerate}
\begin{itemize}
\item
the neighborhood samples $\mathbf{v}\in V$ (yielding \textit{non-central} measurements for LID estimation,
as in Figure~\ref{F:border+inside});
\item
the neighborhood center $\mathbf{q}$ itself
(\textit{central} measurements, Figure~\ref{F:central}); 
\item
for each sample $\mathbf{v}\in V$,
its symmetric reflection $2\mathbf{q}-\mathbf{v}$
taken through the neighborhood center
$\mathbf{q}$ (\textit{reflected} measurements,
Figure~\ref{F:border+inside-with-mirror}).
%\end{enumerate}
\end{itemize}
The use of reflection through $\mathbf{q}$ is
motivated by a desire to balance out whatever non-uniformity may exist
in the neighborhood samples, to obtain a more stable estimate.

The preliminary version of this paper focused on tight LID estimation through
the the combined use of all three types of measurements~\cite{AmsalegCHKRT19}.
Here, in addition to the original estimation strategy, we identify three new
forms and their estimators,
depending upon whether central or reflected measurements are employed
together with the non-central measurements. Variants in which central measurements are included will be denoted by the subscript `c', and those in which reflected measurements are disallowed will be denoted using the superscript `n'.
In the following expressions, $V_{*}\triangleq V\cup\{\mathbf{q}\}$,
and $r$ is the distance from $\mathbf{q}$
to its farthest neighbor in $V$.
\begin{itemize}
\item
$\TLEminus$ (the default estimator proposed in this paper) --- reflected measurements are used, but central measurements are not:
\begin{eqnarray*}
\lefteqn{\IDTLEminus(\mathbf{q}) \: =} \\
%& = &
&   &
-
\Bigg(
\frac{1}{2|V|\cdot(|V|-1)}
\cdot
\\
& &
%\mbox{~~~~}
\sum_{\stackrel{(\mathbf{x},\mathbf{v})\in V}{\mathbf{x}\neq\mathbf{v}}}
\bigg[
\ln \frac{d_{\mathbf{q},r}(\mathbf{x},\mathbf{v})}{r}
{+}
\ln \frac{d_{\mathbf{q},r}(\mathbf{2\mathbf{q}{-}\mathbf{x}},\mathbf{v})}{r}
\bigg]
\Bigg)^{-1} ,
%\label{eq:tightestimator}
\end{eqnarray*}
\item
$\TLEplus$ (the variant originally proposed in~\cite{AmsalegCHKRT19}) --- central measurements are included together with reflected measurements:
\begin{eqnarray*}
\lefteqn{\IDTLEplus(\mathbf{q}) \: =} \\
%& = &
&  &
-
\Bigg(
\frac{1}{2|V_{*}|\cdot(|V_{*}|-1)}
\cdot
\\
& &
%\mbox{~~~~}
\sum_{\stackrel{(\mathbf{x},\mathbf{v})\in V_{*}}{\mathbf{x}\neq\mathbf{v}}}
\bigg[
\ln \frac{d_{\mathbf{q},r}(\mathbf{x},\mathbf{v})}{r}
{+}
\ln \frac{d_{\mathbf{q},r}(\mathbf{2\mathbf{q}{-}\mathbf{x}},\mathbf{v})}{r}
\bigg]
\Bigg)^{-1} ,
%\label{eq:tightestimator}
\end{eqnarray*}
\item
$\TLEminusNR$ --- no reflected or central measurements are used.
\begin{eqnarray*}
\lefteqn{\IDTLEminusNR(\mathbf{q}) \: =} \\
%& = &
&   &
-
\Bigg(
\frac{1}{|V|\cdot(|V|-1)}
\sum_{\stackrel{(\mathbf{x},\mathbf{v})\in V}{\mathbf{x}\neq\mathbf{v}}}
\ln \frac{d_{\mathbf{q},r}(\mathbf{x},\mathbf{v})}{r}
\Bigg)^{-1} \, ,
%\label{eq:tightestimator}
\end{eqnarray*}
\item
$\TLEplusNR$ --- no reflected measurements are used, but central measurements are included:
\begin{eqnarray*}
\lefteqn{\IDTLEplusNR(\mathbf{q}) \: =} \\
%& = &
&   &
-
\Bigg(
\frac{1}{|V_{*}|\cdot(|V_{*}|-1)}
\sum_{\stackrel{(\mathbf{x},\mathbf{v})\in V_{*}}{\mathbf{x}\neq\mathbf{v}}}
\ln \frac{d_{\mathbf{q},r}(\mathbf{x},\mathbf{v})}{r}
\Bigg)^{-1} \, ,
%\label{eq:tightestimator}
\end{eqnarray*}
\end{itemize}

%Collectively, we will refer to these four estimation forms as `TLE'.
All four estimator variants can be derived using
MLE techniques in a manner analogous to that in
which $\widehat{\mathrm{ID}}_{\mathrm{MLE}}$
was derived in~\cite{dami2018estimation};
%for this reason, and for considerations of space, we omit the details here.
for considerations of space, we provide the details only for $\IDTLEminus$.

\subsection{Derivation of Asymptotic MLE Estimator.}
\label{SS:mle}

%Maximization of the likelihood function is one of the most widely used 
%parameter estimation techniques in statistics.
%The Maximum Likelihood Estimator (MLE) 
%has no optimality guarantees for finite samples,
%but has the advantage of being
%asymptotically consistent, optimal, and efficient 
%(in that it achieves the Cramer-Rao bound).

The derivation of $\IDMLE$, the MLE estimator for LID presented in~\cite{dami2018estimation}, holds approximately under the asymptotic assumption that samples are drawn from a neighborhood whose radius is allowed to tend to zero. Here, we employ the same assumptions to show that their derivation also applies to TLE estimation. 

The MLE estimator for $\IDstar_{F_{\mathbf{q}}}$ makes use of the distribution of central distance measurements from $\mathbf{q}$, and the associated c.d.f.\ $F_{\mathbf{q}}$. The estimator was derived by maximizing the log-likelihood function of probability densities taken at the sample distances,
\[
\mathcal{L}(\theta |\, V)
\: = \:
\sum_{\mathbf{x}\in V}
\ln
F'_{\mathbf{q}}(r_{\mathbf{x}}|\, \theta)
\, ,
\]
where $r_{\mathbf{x}}\triangleq \|\mathbf{x}-\mathbf{q}\|$
and $\theta$ is a parameter representing the quantity to be estimated, namely $\IDstar_{F_{\mathbf{q}}}$.
The probability density function $F'_{\mathbf{q}}$ was approximated by differentiating the c.d.f. of the asymptotic limit of the tail distribution, which is known to follow a Generalized Pareto Distribution with shape parameter equal to the negative reciprocal of the LID value~\cite{dami2018estimation}:
\begin{equation*}
F'_{\mathbf{q}}(t)
\: \approx \:
\frac{F_{\mathbf{q}}(r)}{r}
\cdot
\theta
\left(
\frac{t}{r}
\right)^{\theta-1}
\, .
\end{equation*}
Maximizing the log-likelihood function and solving for $\theta$ leads to the formula shown as Equation~\ref{eq:estimator}.

For the case of non-central distance measurements, including those reflected through $\mathbf{q}$, for each sample point $\mathbf{x}\in V$ the c.d.f. function $F_{\mathbf{q}}$ can be replaced by $F_{\mathbf{q},\mathbf{x},r}$ (for the non-reflected distance measurements) and $F_{\mathbf{q},2\mathbf{q}-\mathbf{x},r}$ (for the reflected distance measurements). The log-likelihood function then becomes
\begin{eqnarray*}
\mathcal{L}(\theta |\, V)
& = &
\sum_{\stackrel{(\mathbf{x},\mathbf{v})\in V}{\mathbf{x}\neq\mathbf{v}}}
\left[
\ln
F'_{\mathbf{q},\mathbf{x},r}
\left(
d_{\mathbf{q},r}(\mathbf{x},\mathbf{v})
|\, 
\theta
\right)
\right.
\\
& &
\left.
\mathrm{~~~~~~}
+
\ln
F'_{\mathbf{q},2\mathbf{q}-\mathbf{x},r}
\left(
d_{\mathbf{q},r}(2\mathbf{q}-\mathbf{x},\mathbf{v})
|\, 
\theta
\right)
\right]
\, ,
\end{eqnarray*}
with the probability densities approximated in the same manner as for $F'_{\mathbf{q}}$:
\begin{eqnarray*}
F'_{\mathbf{q},\mathbf{x},r}(t)
& \approx &
\frac{F_{\mathbf{q},\mathbf{x},r}(r)}{r}
\cdot
\theta
\left(
\frac{t}{r}
\right)^{\theta-1}
\\
F'_{\mathbf{q},2\mathbf{q}-\mathbf{x},r}(t)
& \approx &
\frac{F_{\mathbf{q},2\mathbf{q}-\mathbf{x},r}(r)}{r}
\cdot
\theta
\left(
\frac{t}{r}
\right)^{\theta-1}
\, .
\end{eqnarray*}

Here, the use of the same LID parameter $\theta$ throughout is justified under the assumption that the the uniform continuity of LID values in the vicinity of $\mathbf{q}$ leads (via Theorem~\ref{T:interpolation}) to the conclusion
that
\[
\IDstar_{F_{\mathbf{q},\mathbf{x},r}}
\: = \:
\IDstar_{F_{\mathbf{q},2\mathbf{q}-\mathbf{x},r}}
\: = \:
\IDstar_{F_{\mathbf{q}}}
\, .
\]
Maximizing this new log-likelihood function (through partial differentiation with respect to $\theta$) can easily be seen to lead to the estimator of $\IDTLEminus$ stated in Section~\ref{SS:multiple}.

%\todo{MEH: 1. Say something about instabilities. 2. Put in MLE details. 3. Integrate figures + EPS version of clipping.pdf. 4. Expand background discussion of ID. 5. Adjust the intro, acknowledge the SDM version. 6. Add TLE+ and TLE- to the MLE description. 7. Argue better for balancing. 8. Discuss how balanced is better. 9. Discuss how TLE+ and TLE- do not differ much when $q$ is an inlier, but may diverge when $q$ is an outlier (out of the scope of this paper).}

%\todo{MR: 1. Finalize figures - DONE. 2. Add new experimental details and discussions - DONE (integrated with supplement, added Figures~\ref{fig:synth1-d-gaussian-lier} (the inliers), \ref{fig:synth1-k-d50}, \ref{fig:synthm-boxplots-k100}. 3. Add TLE- experiments, and balanced vs unbalanced experiments for both TLE+ and TLE- - DONE}

%\todo{LA, others?: check, suggest new content, polish, etc.?}

\section{Experimental Framework}
\label{S:framework}

%\todo{MR: 1. Add some intro - DONE. 2. Change support page to GitHub - DONE. 3. Explain differences from conference version - DONE.}

In this section we describe our experimental setup: the competing estimation methods, the two families of synthetic data, and the real data sets employed. All source code and data necessary to replicate our experiments are available through GitHub:
\href{https://github.com/radacha/tle}{\url{https://github.com/radacha/tle}}. 

As an extension of the preliminary version of our paper~\cite{AmsalegCHKRT19}, we have included the three new TLE estimator variants in our experimentation. 
Note that the TLE estimator used in~\cite{AmsalegCHKRT19} is denoted by TLE$_\mathrm{c}$ in this version.
All implementations of TLE estimators have been made more stable and reliable by handling additional boundary cases, as outlined in Section~\ref{SS:moving}.

%%%%%%%%%%%%%%%%%
\subsection{Competing Estimation Methods.}

To show the advantages and limitations of TLE, we first compare our proposed default estimator $\IDTLEminus$ with its variants, justifying our choice of the version that excludes central measurements and includes reflected measurements. We then compare $\IDTLEminus$ with other recognized local estimation methods of intrinsic dimensionality: MLE and method of moments (MoM) estimators of LID~\cite{dami2018estimation}; a local application of Grassberger and Procaccia's correlation dimension~\cite{grassberger2004measuring}, restricted to $k$-neighborhoods (LCD); (generalized) expansion dimension (ED, GED)~\cite{ged}; and a local application of PCA, again restricted to $k$-neighborhoods (LPCA). We also consider a local application of $\textrm{MiND}_{\mathrm{ml1}}$ and~$\textrm{MiND}_{\mathrm{mli}}$~\cite{RozzaLCCC12}, two parameterless methods designed for producing global ID estimates of query sets. All local methods are parameterized by the neighborhood size~$k$. The parameter choices for all methods are summarized in Table~\ref{t:parameters}.

%We denote by l-PCA the estimator obtained by applying PCA on the respective
%neighborhoods of size $k=100$.

\begin{table}
\centering
\begin{tabular}{ l l }
  \hline
  Method	& Parameters \\ \hline\hline
  $\widehat{\mathrm{ID}}_\mathrm{TLE}$
  									& $k = 100$ \\ \hline
  $\widehat{\mathrm{ID}}_\mathrm{MLE}$~\cite{dami2018estimation}
  									& $k = 100$ \\ \hline
  $\widehat{\mathrm{ID}}_\mathrm{MoM}$~\cite{dami2018estimation}
  									& $k = 100$ \\ \hline
  $\mbox{LCD}$~\cite{grassberger2004measuring}			
  									& $k = 100$ \\ \hline
  $\mbox{LPCA}$~\cite{Jolliffe86}			
  									& $k = 100$, $\Theta = 0.025$	\\ \hline
  $\textrm{MiND}_{\mathrm{ml1}}$~\cite{RozzaLCCC12}
  									& None \\ \hline
  $\textrm{MiND}_{\mathrm{mli}}$~\cite{RozzaLCCC12}
  									& $k = 100$ \\ \hline
  $\mathrm{PCA}$~\cite{Jolliffe86}
  									& $\Theta = 0.025$ \\ \hline
\end{tabular}
\caption{Parameter choices for the methods used in the experiments.}
\label{t:parameters}
\end{table}

It should be noted that LPCA and methods from the $\textrm{MiND}$ family must
be provided with knowledge of the representational dimension, which may give
them an advantage in head-to-head comparison with other methods. Moreover,
when applied to synthetic data sets, LPCA and $\textrm{MiND}_\mathrm{mli}$
can often return the exact dimension, since they can return only
integer-valued estimates. While it may be claimed that the intrinsic
dimension should ideally be an integer, for real data this is not always the
case. For example, LID has been shown to be equivalent to a measure of the
indiscriminability of the distance measure, which is in general not an
integer~\cite{dami2018estimation}. Furthermore, non-integer values of ID can
indicate non-linear properties of an underlying manifold, such as convexity.

%%%%%%%%%%%%%%%%%
\subsection{Synthetic Data.}

Our study includes two families of synthetic data sets. For each manifold we
generated 20 sets of $10^4$ points, and in each experiment we report the
averages of observed means and standard deviations of ID measures over the 20
sets. The first family is i.i.d. Gaussian, uniform in the unit cube and
multidimensional torus, meant to evaluate behavior of local ID estimators
with increasing dimensionality and neighborhood size. The second family (m)
is a benchmark collection of various types of
manifolds~\cite{RozzaLCCC12,dami2018estimation}, summarized in
Table~\ref{t:manifolds}.

\begin{table}
\begin{tabular}{ c r r l }
  \hline
  \scriptsize{Manifold} 	& $d$ 	& $D$ 	& Description \\ \hline\hline
%  h-$d$		& $d$	& $d$	& Uniformly sampled \\
%  			&		&		& $d$-dimensional hypercube. \\ \hline
  m1		& 10	& 11	& Uniformly sampled sphere. \\ \hline
  m2		& 3		& 5		& Affine space. \\ \hline
  m3		& 4		& 6		& Concentrated figure \\
  			& 		& 		& confusable with a 3d one. \\ \hline
  m4		& 4		& 8		& Non-linear manifold. \\ \hline
  m5		& 2		& 3		& 2-d Helix \\ \hline
  m6		& 6		& 36	& Non-linear manifold. \\ \hline
  m7		& 2		& 3		& Swiss-Roll. \\ \hline
  m8		& 12	& 72	& Non-linear manifold. \\ \hline
  m9		& 20	& 20	& Affine space. \\ \hline
  m10a		& 10	& 11	& Uniformly sampled hypercube. \\ \hline
  m10b		& 17	& 18	& Uniformly sampled hypercube. \\ \hline
  m10c		& 24	& 25	& Uniformly sampled hypercube. \\ \hline
  m11		& 2		& 3		& M\"{o}bius band 10-times twisted. \\ \hline
  m12		& 20	& 20	& Isotropic multivariate Gaussian. \\ \hline
  m13		& 1		& 13	& Curve. \\ \hline
\end{tabular}
\caption{Artificial data sets used in the experiments.}
\label{t:manifolds}
\end{table}

\subsection{Real Data.}

The use of real-world data sets lacks the ground truth available for
synthetic data. Therefore, to evaluate TLE on such sets, we compare the bias
and variance characteristics directly against competing methods using the 8
real data sets listed in Table~\ref{t:manifolds2}.

\begin{table}
\centering\footnotesize
\begin{tabular}{ l r r }
  \hline
  Data set 	& Instances 	& Dim.		 	\\ \hline
  \emph{ALOI}~\cite{Boujemaa}
  			& 110250		& 641			\\
  \emph{ANN\_SIFT1M}~\cite{PQ11PAMI}
  			& $10^6$		& 128			\\
  \emph{BCI5}~\cite{bci5}
  			& 31216			& 96			\\
  \emph{CoverType}~\cite{blackard1999comparative}
  			& 581012 		& 54 			\\
  \emph{Gisette}~\cite{arcene-gisette}
  			& 7000			& 5000			\\
  \emph{Isolet}~\cite{isolet}
  			& 7797			& 617			\\
  \emph{MNIST}~\cite{journal/pieee/LeCunBBH98}
  			& 70000			& 784			\\
  \emph{MSD}~\cite{bertin2011million}
  			& 515345		& 90			\\ \hline
\end{tabular}
\caption{Real data sets used in the experiments.}
\label{t:manifolds2}
\end{table}

\begin{itemize}
\item The \emph{ALOI} (\emph{Amsterdam Library of Object Images}) data
    consists of $110250$ color photos of $1000$ different objects. Photos
    are taken from varying angles under various illumination conditions.
    Each image is described by a $641$-dimen\-sional vector of color and
    texture features~\cite{Boujemaa}.
\item The \emph{ANN\_SIFT1M} data set consists of $10^6$ $128$-dimensional
    SIFT descriptors randomly selected from the \emph{ANN\_SIFT} data set
    which contains $2.8 \cdot 10^{10}$ SIFT descriptors extracted from $3
    \cdot 10^{7}$ images. The data set was introduced in~\cite{PQ11PAMI}.
\item \emph{BCI5}~\cite{bci5} is a brain-computer interface data set in
    which the classes correspond to brain signal recordings taken while the
    subject contemplated one of three different actions (movement of the
    right hand, movement of the left hand, and the utterance of words
    beginning with the same letter).
\item \emph{CoverType}~\cite{blackard1999comparative} consists of 581012
    geographical locations (a surface of 30 by 30 meters) described by 54
    attributes. each location is majorly covered by one of seven tree
    species.
\item \emph{Gisette}~\cite{arcene-gisette} is a subset of the MNIST
    handwritten digit image data set~\cite{journal/pieee/LeCunBBH98},
    consisting of 50-by-50-pixel images of the highly confusable digits '4'
    and '9'. 2500 random features were artificially generated and added to
    the original 2500 features, so as to embed the data into a
    higher-dimensional feature space.
\item \emph{Isolet}~\cite{isolet} is a set of 7797 human voice recordings
    in which 150 subjects read each of the 26 letters of the alphabet
    twice. Each entry consists of 617 features representing utterances of
    the recording.
\item The \emph{MNIST} database~\cite{journal/pieee/LeCunBBH98} contains of
    $70000$ recordings of handwritten digits. The images have been
    normalized and discretized to a $28 \times 28$-pixel grid. The
    gray-scale values of the resulting $784$ pixels are used to form the
    feature vectors.
\item \emph{MSD}~\cite{bertin2011million} is a subset of the `Million Song
    Database' which is a set of radio recordings (from the years 1922 to
    2011) described by 12 timbre averages and 78 timbre covariances.
\end{itemize}

\section{Experimental Results}
\label{S:res}

\subsection{TLE Variants.}

\begin{figure}
\includegraphics[width=\columnwidth]{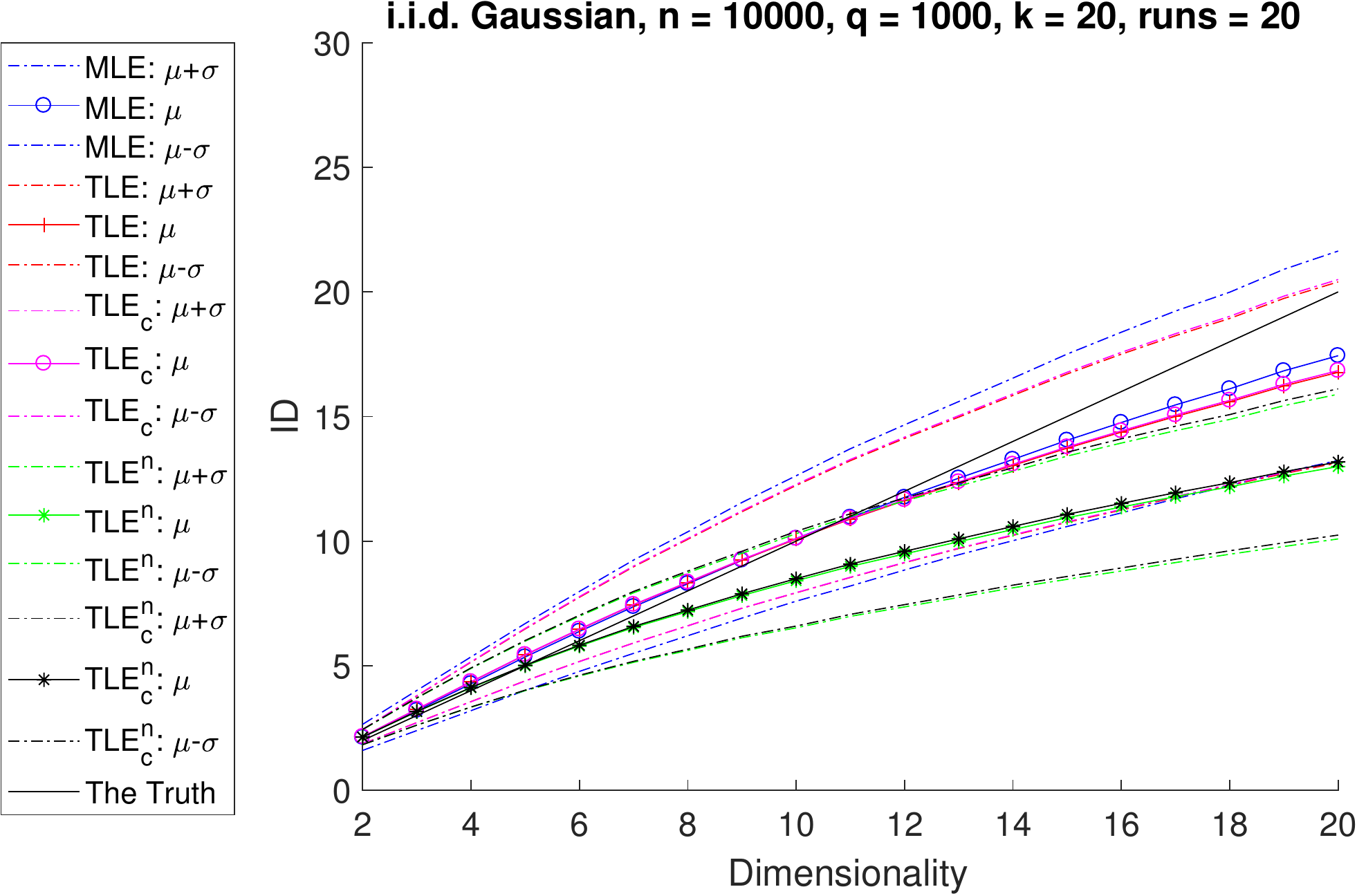}
\caption{Means and standard deviations of ID values estimated by MLE and TLE variants on i.i.d.\ Gaussian data,
for neighborhood size 20 and various dimensionalities.}\label{fig:synth1-d-gaussian-tlevariants}
\end{figure}

To illustrate the similarities and differences between the proposed variants of TLE and justify our choice of the variant excluding central and including reflected measurements, consider  Figure~\ref{fig:synth1-d-gaussian-tlevariants} which was obtained by plotting the mean values of ID estimates, together with standard deviations, obtained on 10000 i.i.d.\ Gaussian random data points with dimensionalities ranging from 2 to 20, and averaged over 20 runs. All methods were executed with neighborhood size $k = 20$. Also plotted is `The Truth', the embedding dimensionality of the data space, which is identical to the theoretical true local ID at each data point.

It can be seen that inclusion of central distance measurements has little or no effect on the produced TLE estimates. Inclusion of reflected (non-central) measurements, however, is essential for reducing the bias of the estimator. In general, compared to MLE (and to the other methods as well), TLE reduces the variance of the estimates, at the expense of an increase in bias which, for the variants that exclude reflected measurements, can be excessive. For these reasons, as the default TLE method we recommend for simplicity the variant that excludes central distance measurements, and includes non-central and reflected ones. 

\begin{figure}
\includegraphics[width=\columnwidth]{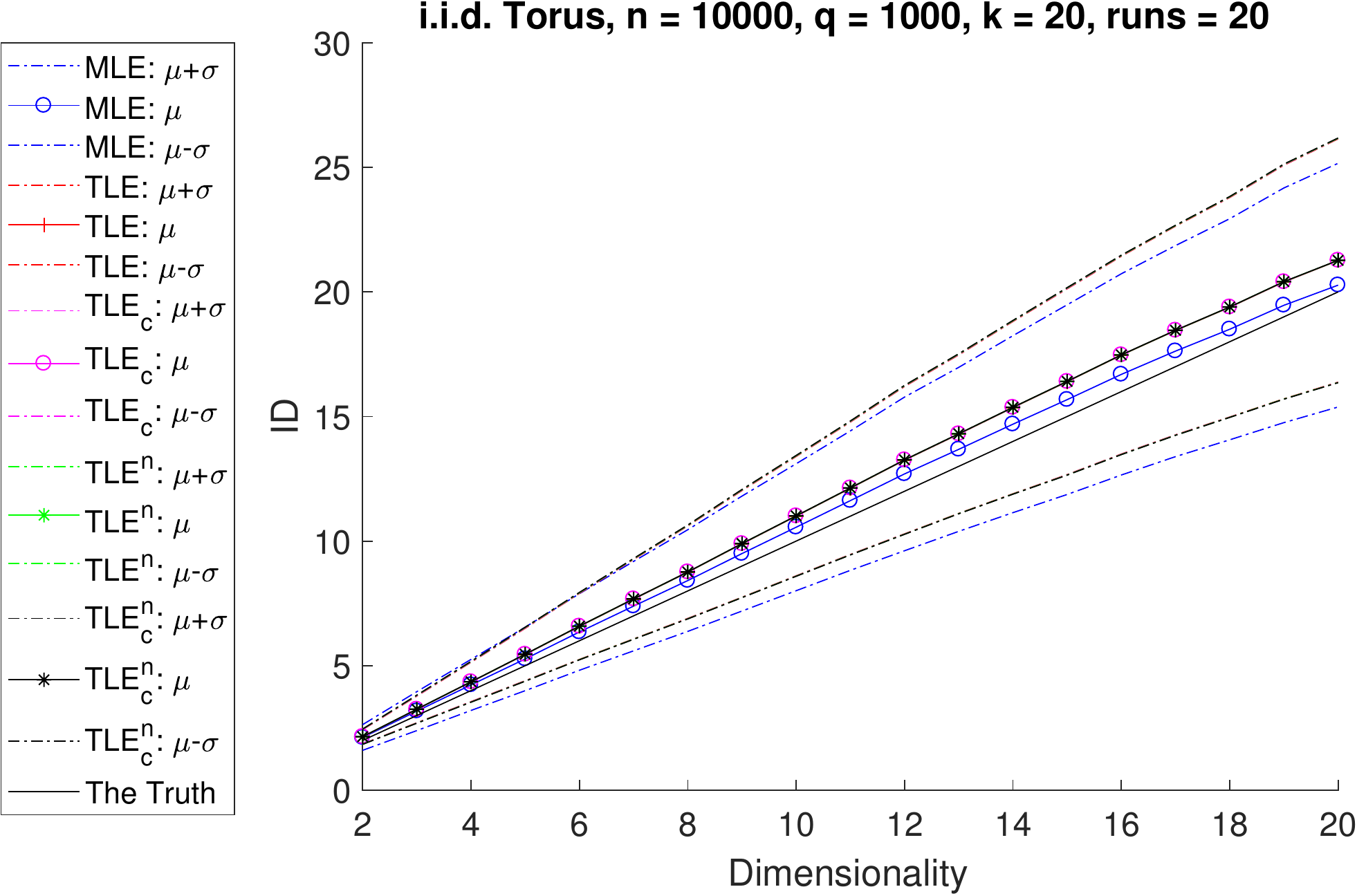}
\caption{Means and standard deviations of ID values estimated by MLE and TLE variants on i.i.d.\ torus data,
for neighborhood size 20 and various dimensionalities.}\label{fig:synth1-d-torus-tlevariants}
\end{figure}

We observed similar trends on the other synthetic and real data sets used in the experiments (omitted here), except for i.i.d.\ uniform random data in the multidimensional torus (shown in Figure~\ref{fig:synth1-d-torus-tlevariants}). What is interesting about this setting is that the distribution is both uniform and without boundary effects, and is therefore immune to clipping bias. Here, as expected, all variants of the TLE estimator exhibited identical behavior. However, when boundaries are present (as is the case with all our other synthetic and real data sets), the bias-variance trends described above can clearly be observed.

\subsection{Synthetic Data.}

\begin{figure}
\includegraphics[width=\columnwidth]{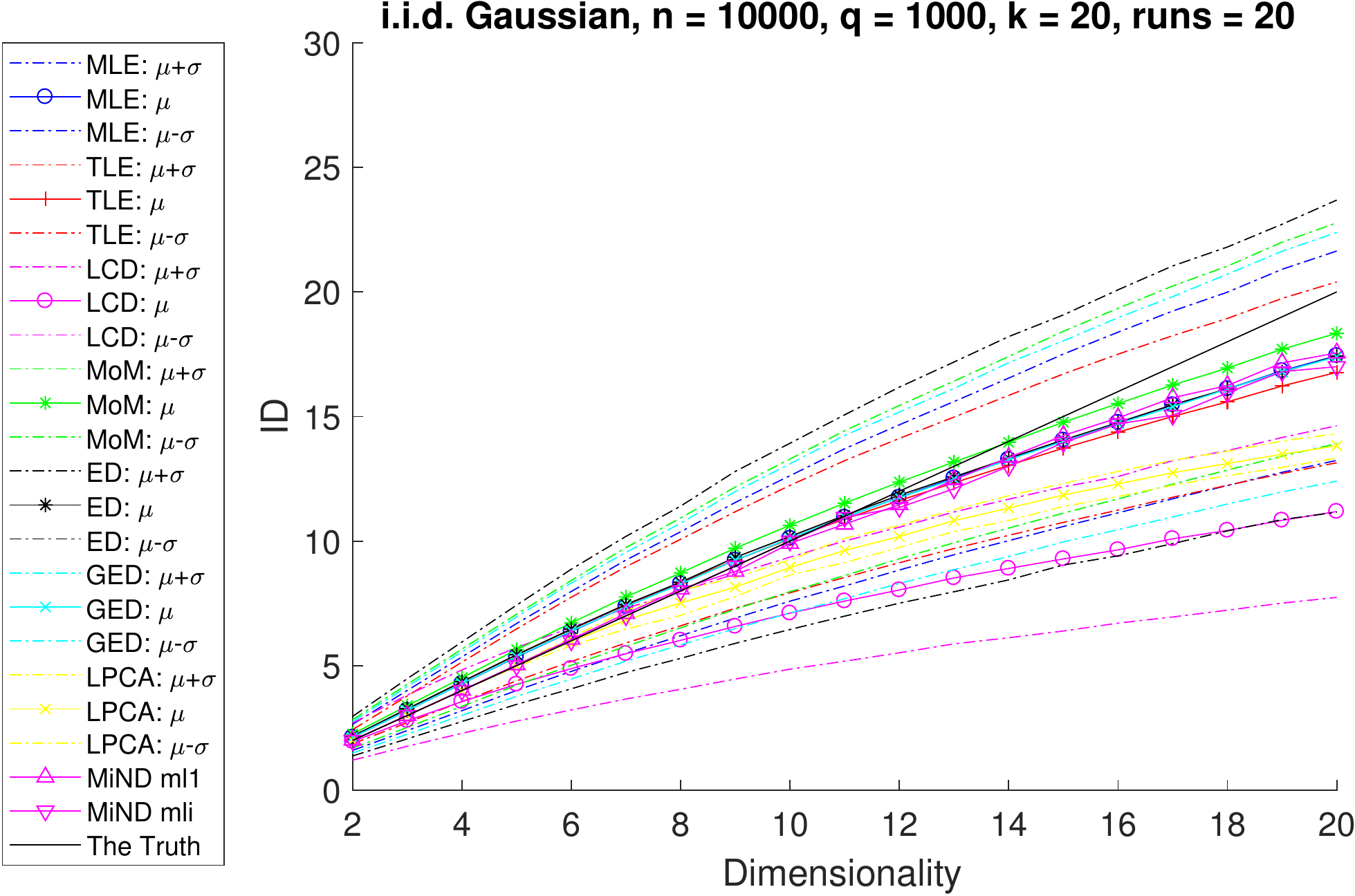}
\caption{Means and standard deviations of ID values on i.i.d.\ Gaussian data,
for neighborhood size 20 and various dimensionalities, as estimated for all data points.}\label{fig:synth1-d-gaussian}
\end{figure}

As a first comparison of TLE with competing ID estimators, let us consider Figure~\ref{fig:synth1-d-gaussian}, which was obtained by plotting the mean values of ID estimates, together with standard deviations (where appropriate), on the same i.i.d.\ Gaussian random data as in Figure~\ref{fig:synth1-d-gaussian-tlevariants}. The general shape of the ID estimate curves suggests that estimators progressively become more negatively biased as dimensionality increases. However, it can be seen that TLE consistently exhibits smaller variance, at the same time maintaining bias comparable to the state-of-the art methods, notably MLE and MoM. By not accounting for distances to the neighbors of neighbors that are outside the locality boundary, LCD can be expected to accumulate clipping bias; as our experimentation has shown, LCD indeed has considerably more negative bias than TLE. For small to medium dimensionalities (2 to 10), the slightly stronger bias of TLE actually produces more accurate estimates, whereas in higher dimensions ($>$10) the bias results in a greater deviation from the ground truth. Please note that standard deviations are not shown for the two MiND methods as they produce estimates for data sets as a whole, and not for individual data points.

%TODO: reformulate when reference is located:
%The general shape of the ID estimate
%curves, where estimators progressively become more biased downward as
%dimensionality increases, compared to the ground truth, is consistent with
%the theoretical result by [TODO: Mike, please insert reference].

\begin{figure*}
\includegraphics[width=.49\textwidth]{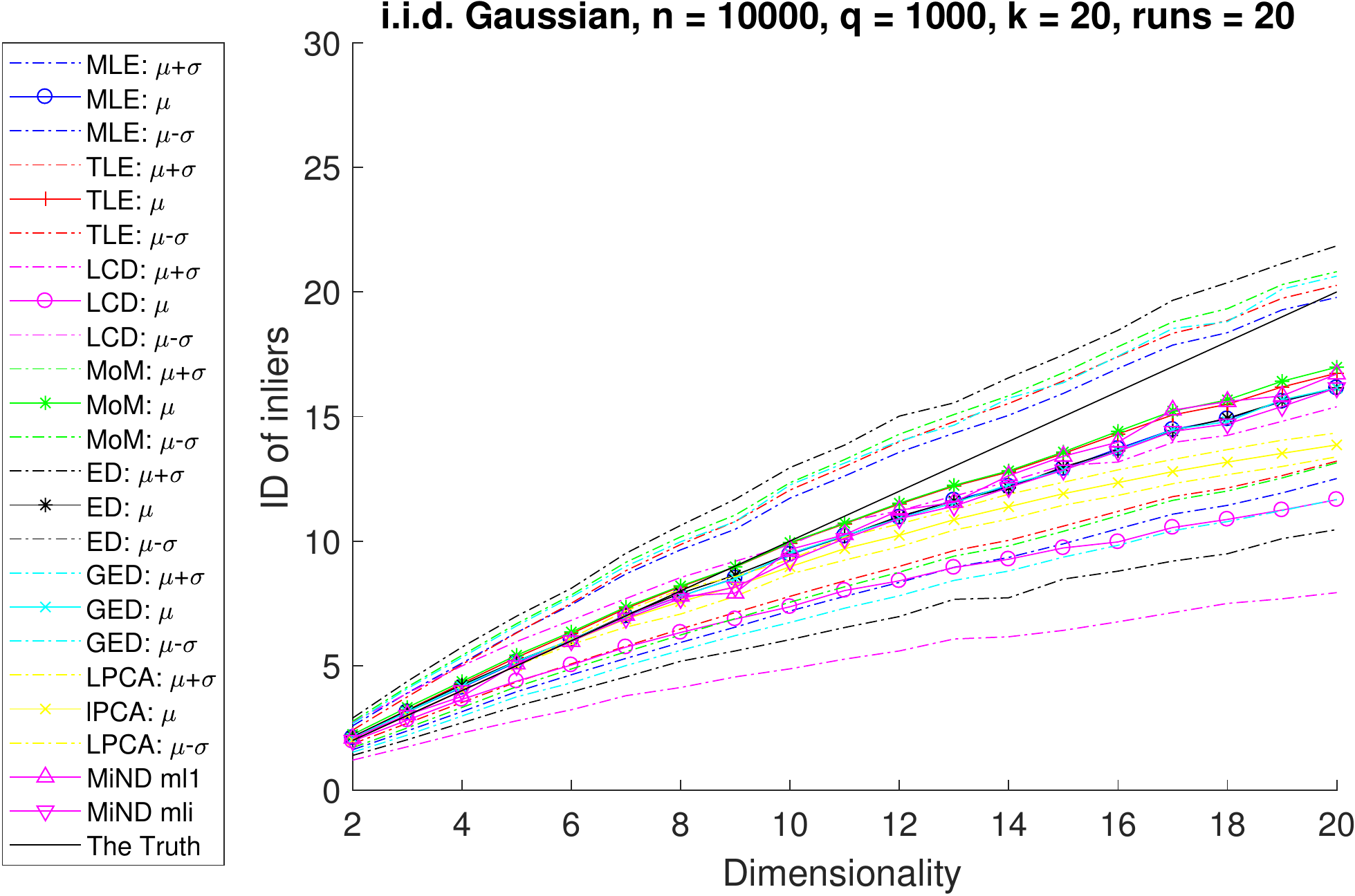}
\includegraphics[width=.49\textwidth]{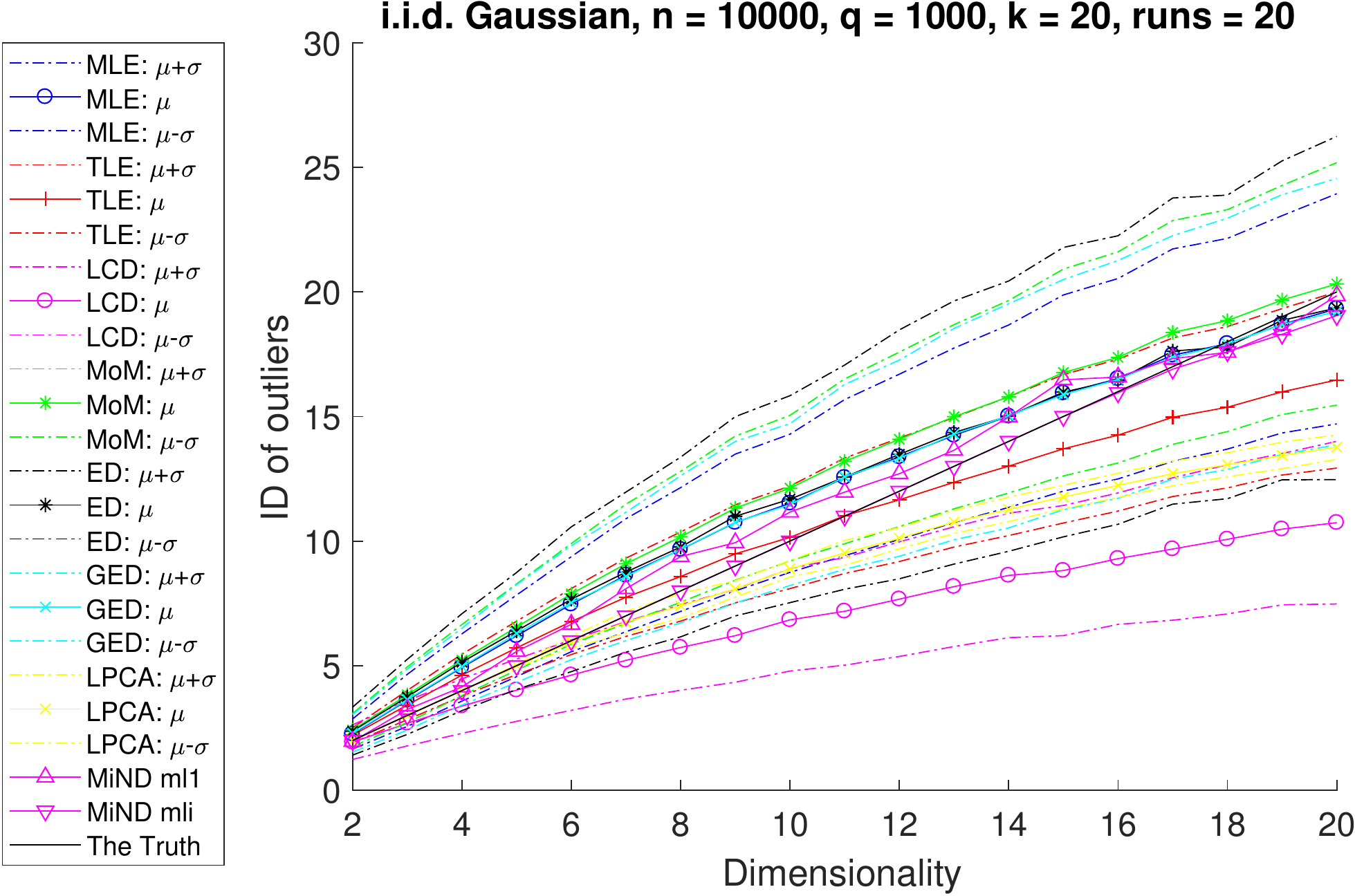}
\caption{Means and standard deviations of estimated ID values on i.i.d.\ Gaussian data,
for neighborhood size 20 and various dimensionalities. Left: estimates for 10\% of strongest inliers.
Right: estimates for 10\% of strongest outliers.}\label{fig:synth1-d-gaussian-lier}
\end{figure*}

To illustrate another scenario where TLE may offer an advantage,
Figure~\ref{fig:synth1-d-gaussian-lier} shows the same ID estimates plotted
only for the 10\% of strongest inliers (left) and outliers (right) in the
same i.i.d.\ Gaussian random data, where inlierness and outlierness
is determined by the distance to the data distribution center.
In the case of outliers, it can be seen
that the plot for TLE stays approximately the same as in the previous figure
where the entire data set was used, whereas other methods become notably more
positively biased, with variance still larger than that of TLE. In the case of
inliers, the plot for TLE does not significantly change, while most other
methods become (slightly) more negatively biased. In all, it seems that TLE
is more robust to point outlierness than other methods, which may be a
consequence of its use of distance relationships within the neighborhood rather than those to the outlier itself. We observed
similar trends in bias and variance on i.i.d.\ uniform and torus data. A deeper
analysis of this behavior is left as a point for future work.

\begin{figure*}
\includegraphics[width=.49\textwidth]{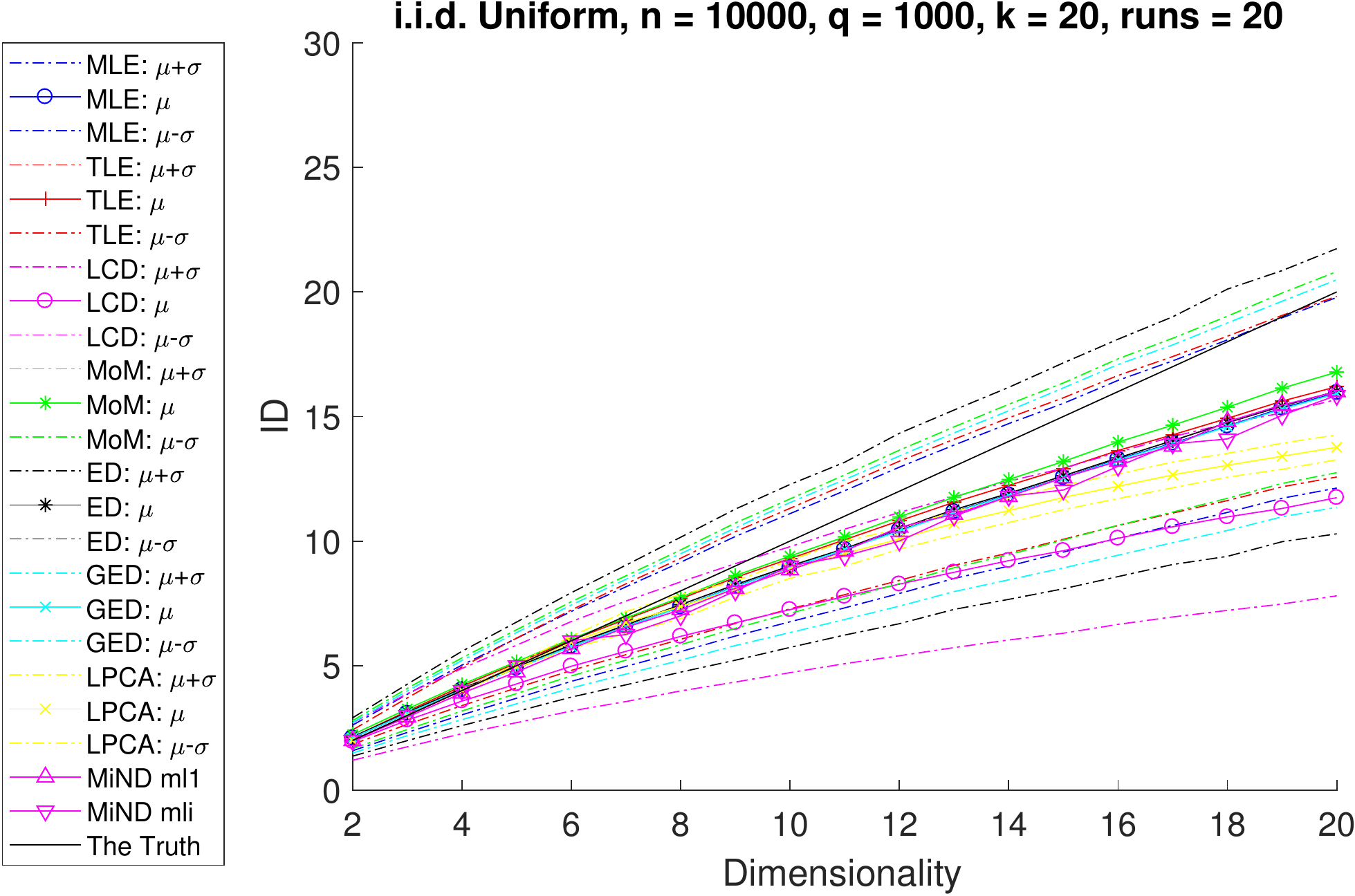}
\includegraphics[width=.49\textwidth]{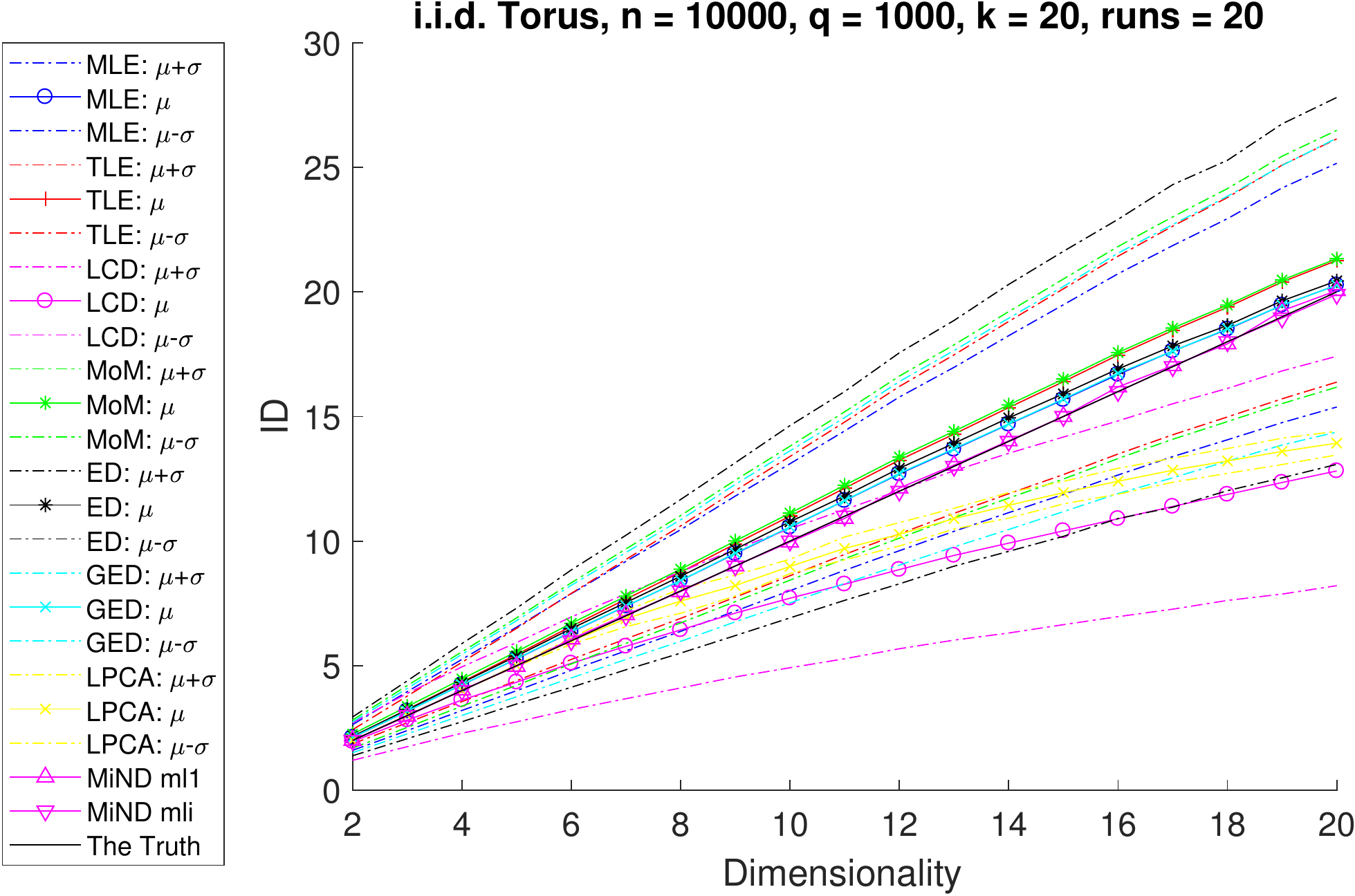}
\caption{Means and standard deviations of estimated ID values, for neighborhood size 20 and various dimensionalities.
Left: i.i.d.\ uniform data. Right: i.i.d.\ torus data.}\label{fig:synth1-d-uniform+torus}
\end{figure*}

In order to support the observations concerning bias and variance from
Figure~\ref{fig:synth1-d-gaussian} (left), in
Figure~\ref{fig:synth1-d-uniform+torus} we show the same plots using two
other synthetic data distributions: i.i.d.\ uniform in the unit cube (left), and
the multidimensional torus (right). For uniform data
(Figure~\ref{fig:synth1-d-uniform+torus}, left), the trends are similar as
with the Gaussian, with the notable difference that there is stronger
negative bias affecting all methods. It appears that boundary effects may be
a significant factor in this, since for the torus data
(Figure~\ref{fig:synth1-d-uniform+torus}, right) the overall bias is much more strongly positive 
(although with increasing dimensionality, all
methods will eventually fall below the ground truth). In all cases, the
general trend persists: TLE produces comparable bias and smaller variance as compared to the other estimation methods in our study.

\begin{figure*}[p]
\includegraphics[width=.49\textwidth]{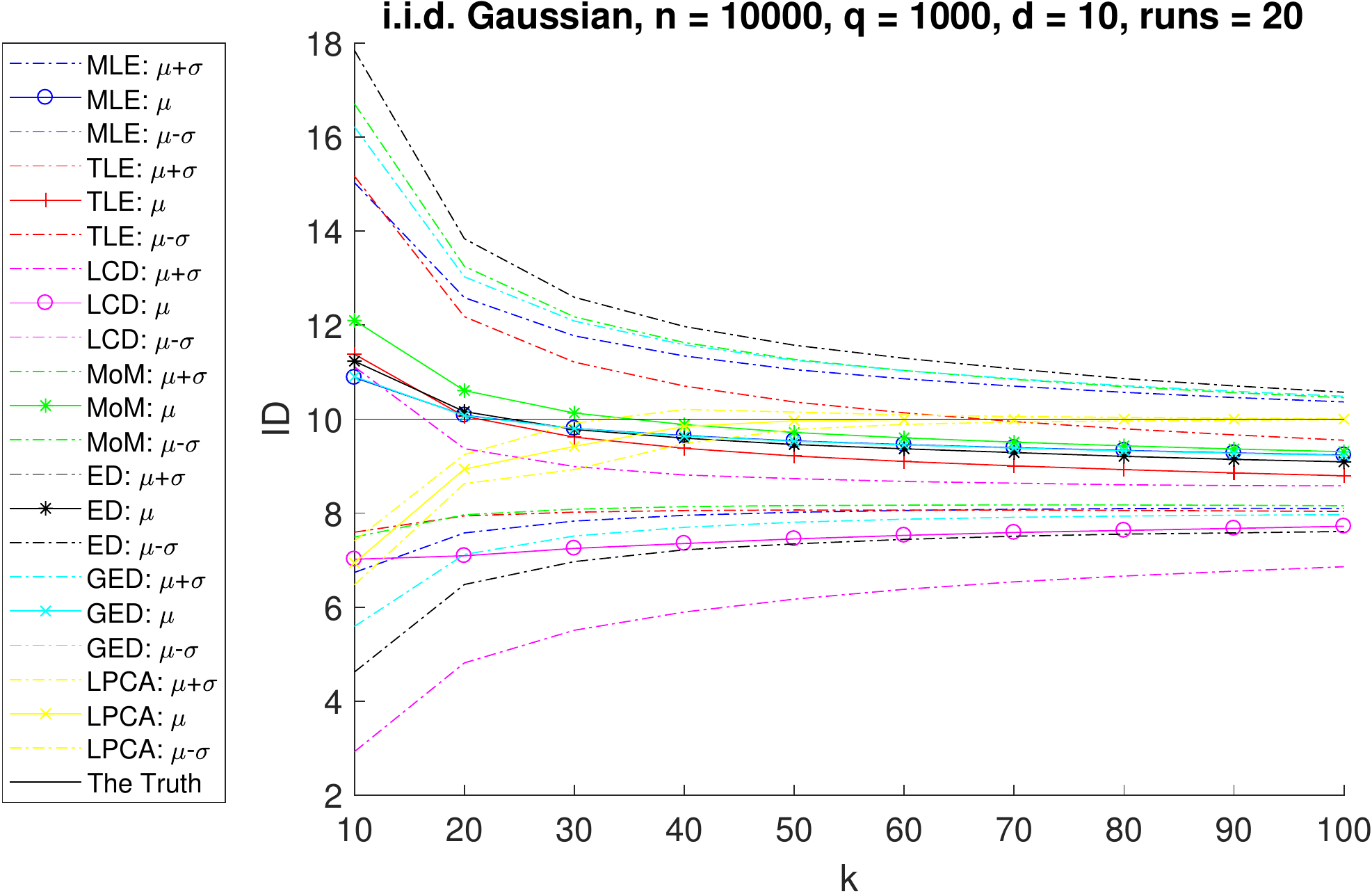}
\includegraphics[width=.49\textwidth]{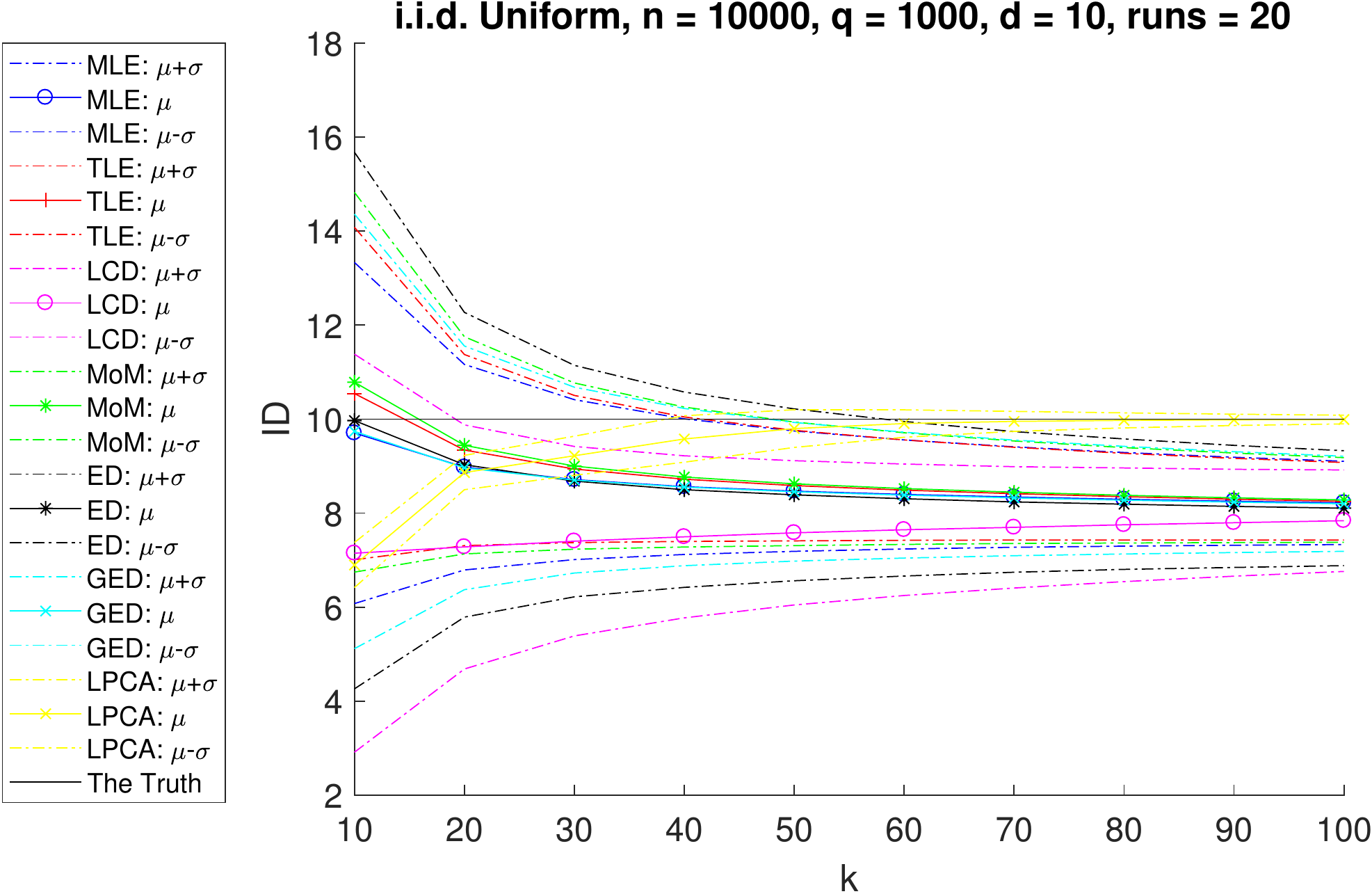}
\caption{Means and standard deviations of estimated ID values, for dimensionality 10 and various neighborhood sizes.
Left: i.i.d.\ Gaussian data. Right: i.i.d.\ uniform data.}\label{fig:synth1-k-d10}
\end{figure*}

\begin{figure*}[p]
\includegraphics[width=.49\textwidth]{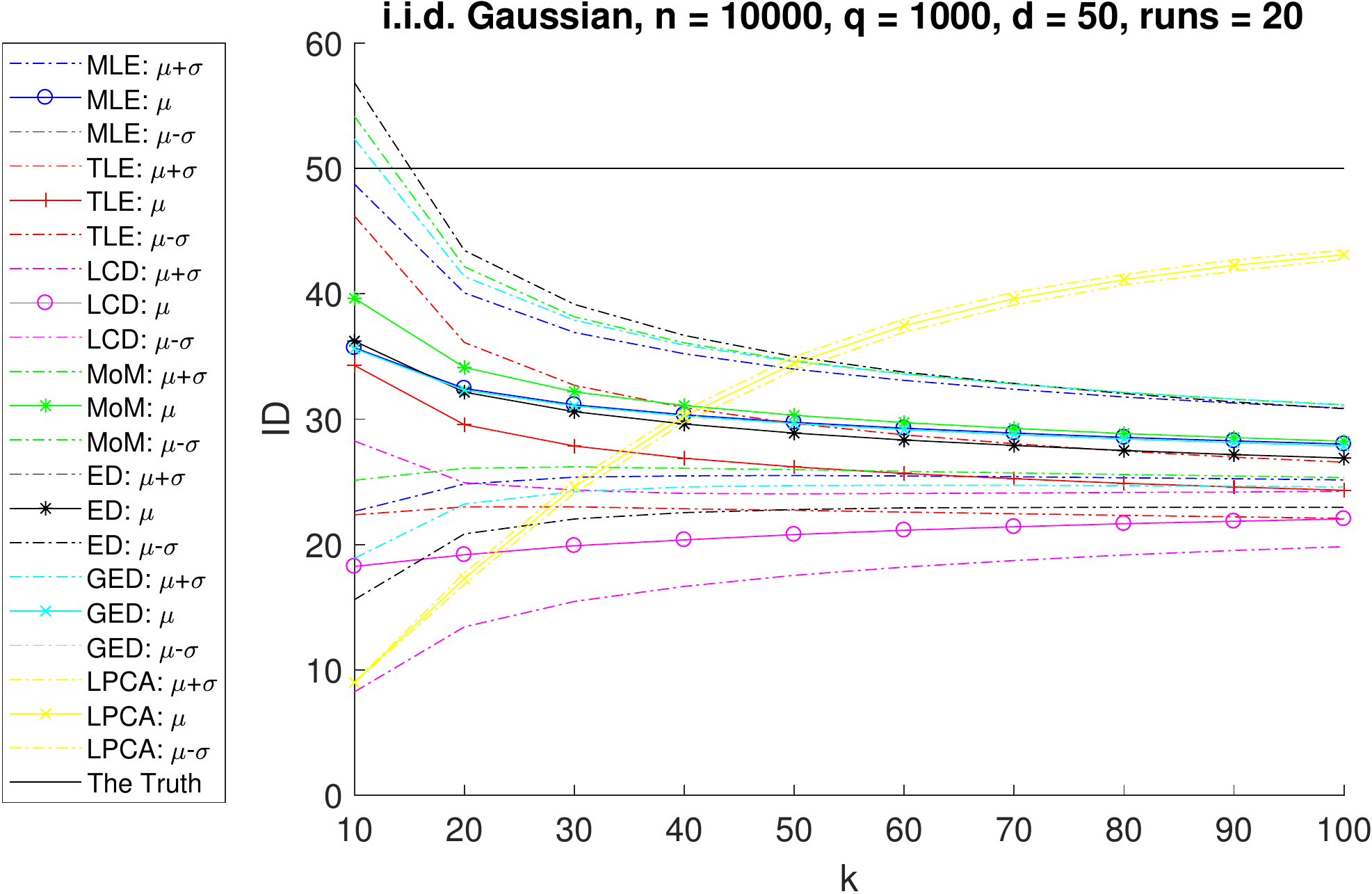}
\includegraphics[width=.49\textwidth]{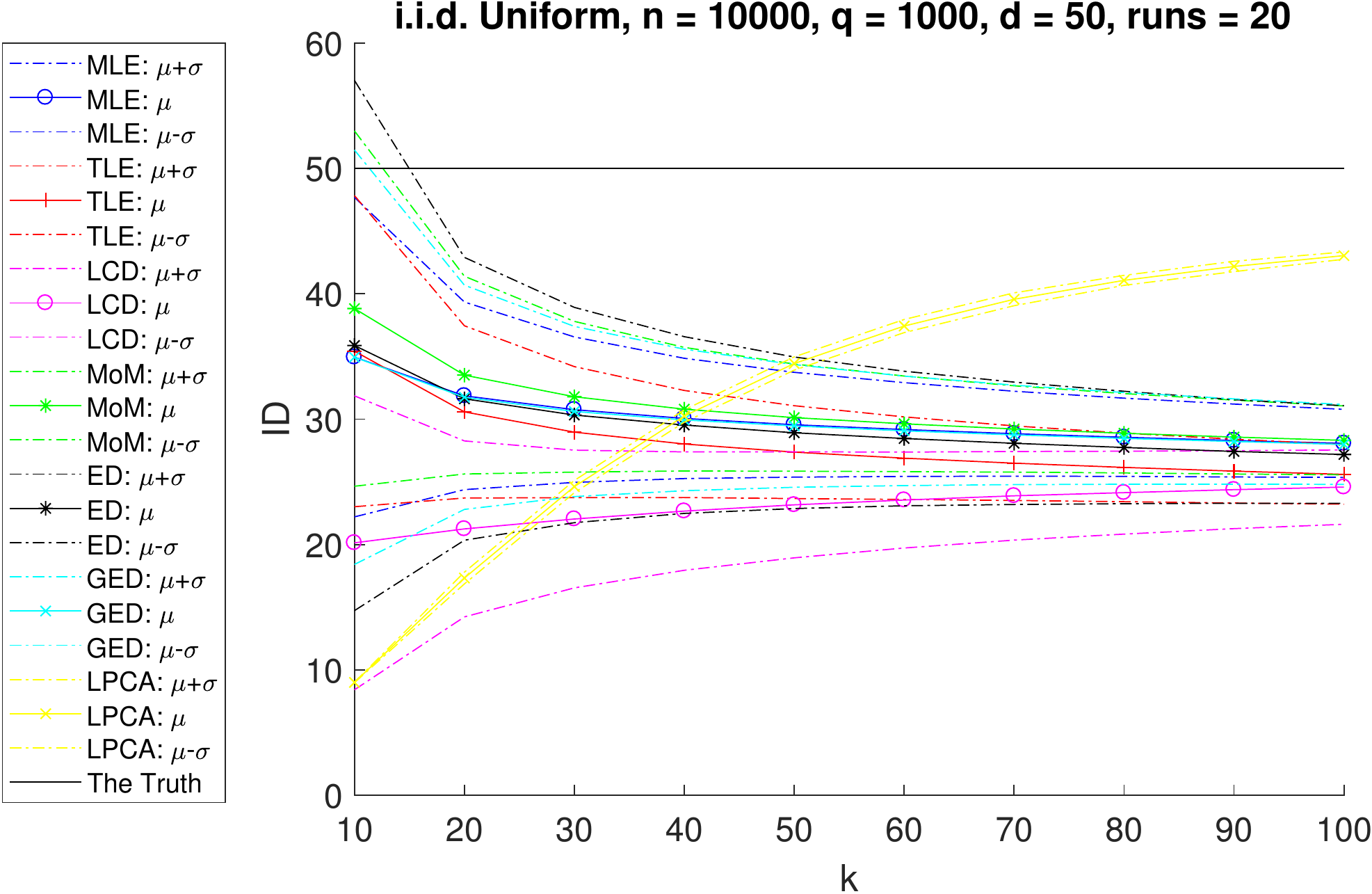}
\caption{Means and standard deviations of estimated ID values, for dimensionality 50 and various neighborhood sizes.
Left: i.i.d.\ Gaussian data. Right: i.i.d.\ uniform data.}\label{fig:synth1-k-d50}
\end{figure*}

In addition to ground truth dimensionality, another important factor to consider
when analyzing the performance of ID estimation methods is the neighborhood
size~$k$. We repeated our experimentation on on i.i.d.\ Gaussian and uniform random data, this time with varying $k$ and
dimensionality fixed at $d = 10$ and $d = 50$; the results are shown in Figures~\ref{fig:synth1-k-d10} and~\ref{fig:synth1-k-d50}. It is evident that for all methods, negative bias increases
with increasing $k$. The general trend of TLE having
comparable bias and smaller variance is also exhibited here, permitting TLE
to be used with smaller values of~$k$, thus at least partially avoiding this
source of bias (in addition to avoiding the potential computational
bottleneck due to TLE having quadratic time complexity in~$k$). Naturally,
the bias of all methods worsens as dimensionality increases (exhibited by the
significantly lower plots in Figure~\ref{fig:synth1-k-d50} for $d = 50$ than
Figure~\ref{fig:synth1-k-d10} for $d = 10$), but the described relative
trends remain.

\begin{figure*}
\begin{centering}
\includegraphics[width=.30\textwidth]{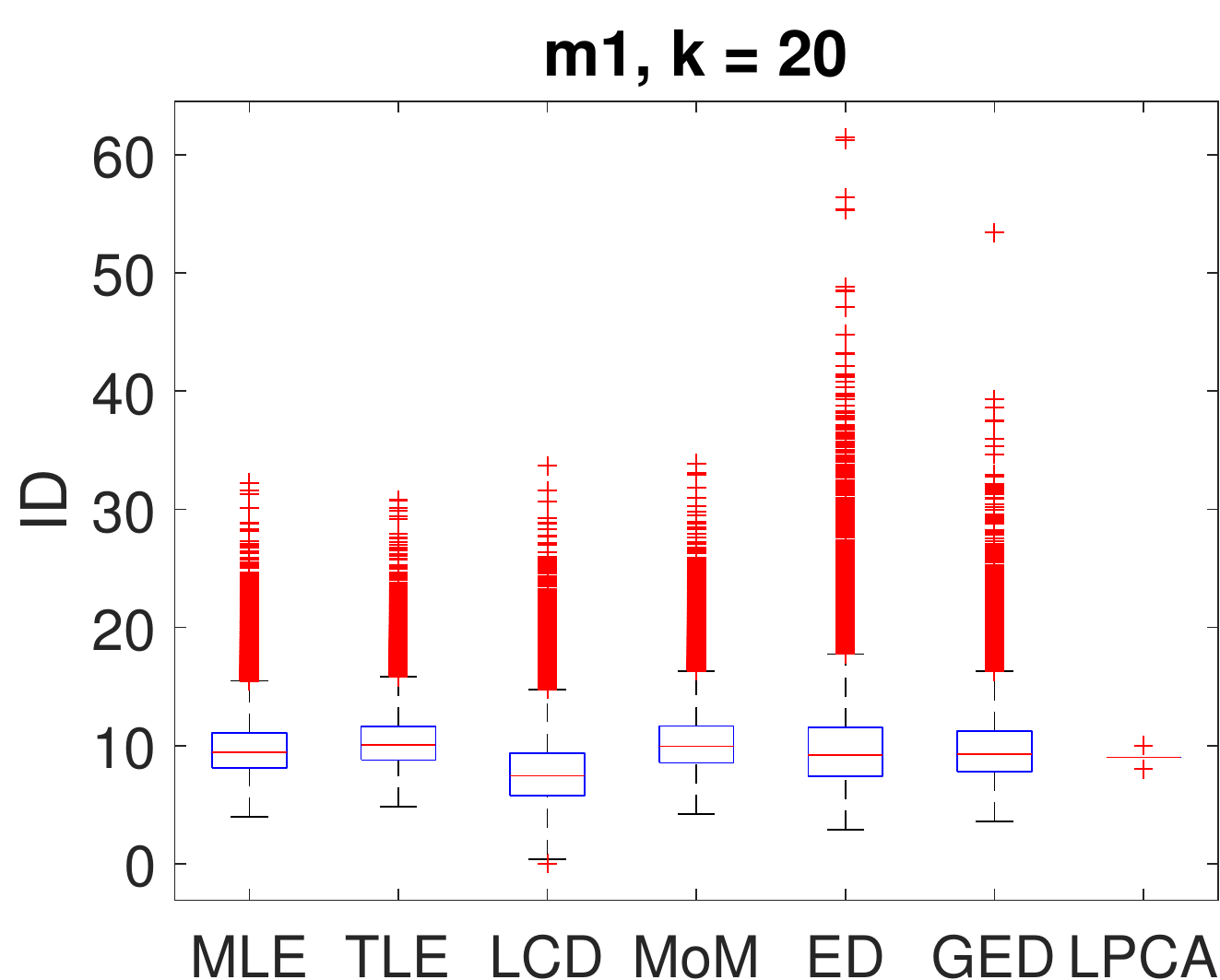}
\includegraphics[width=.30\textwidth]{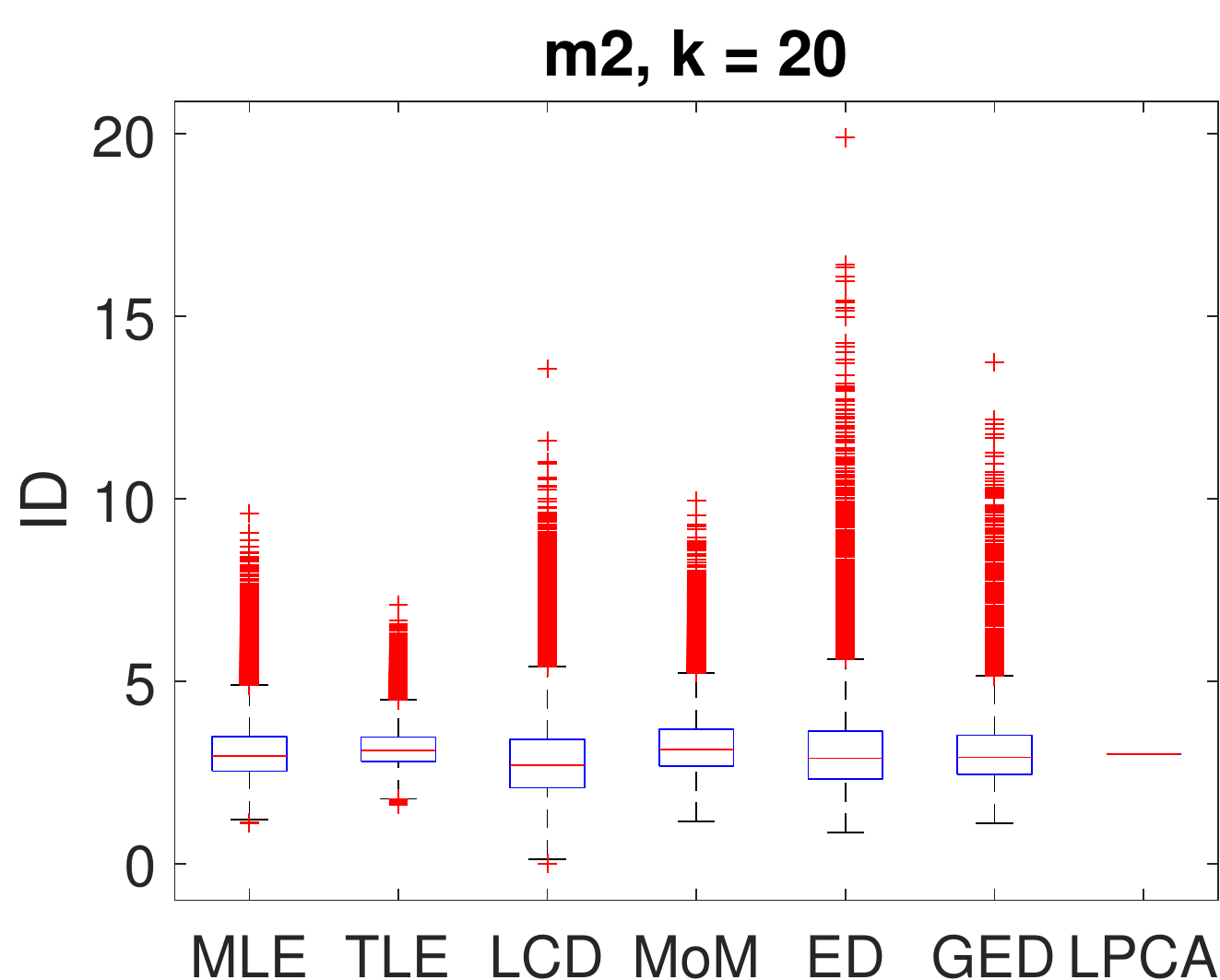}
\includegraphics[width=.30\textwidth]{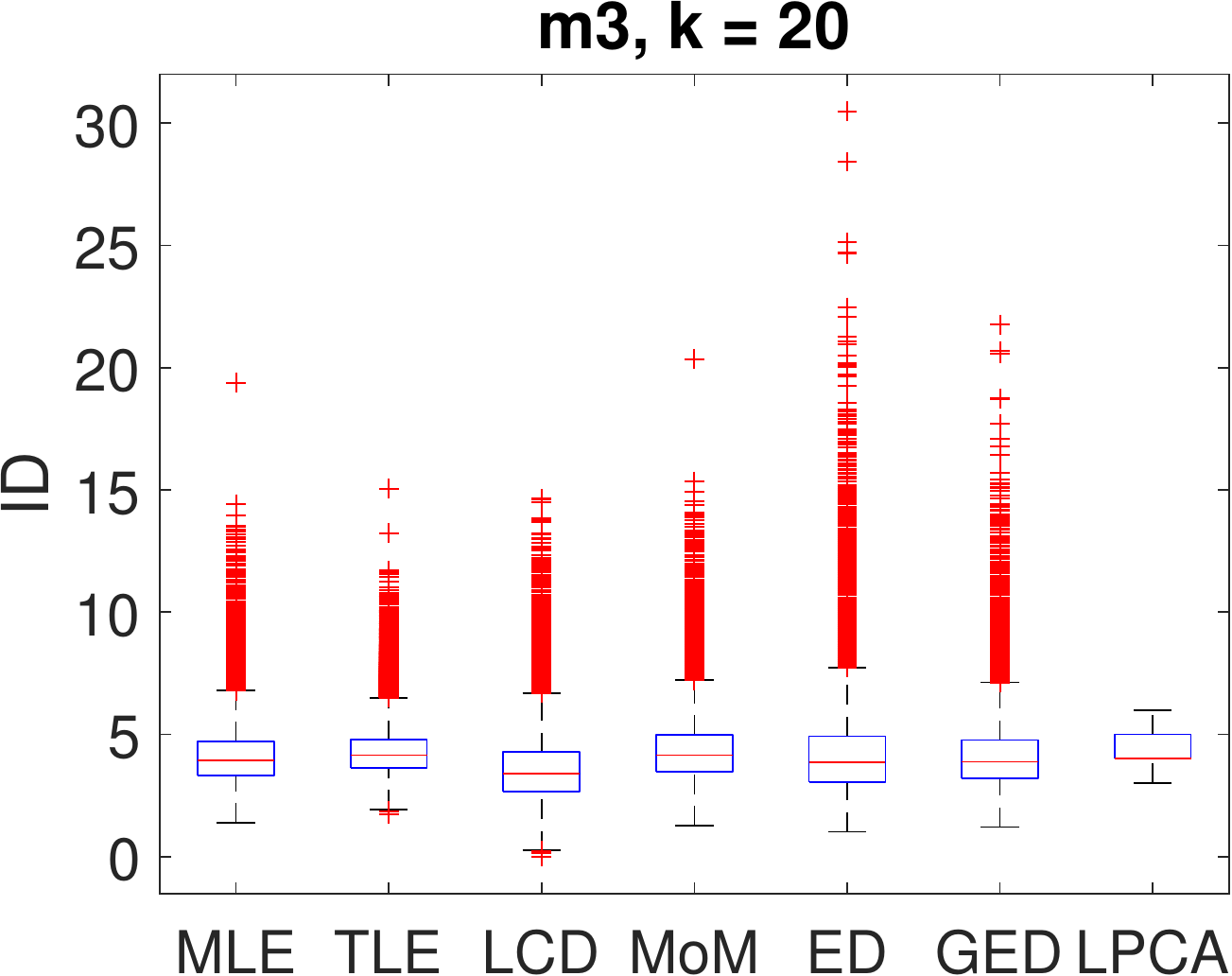}\\
\includegraphics[width=.30\textwidth]{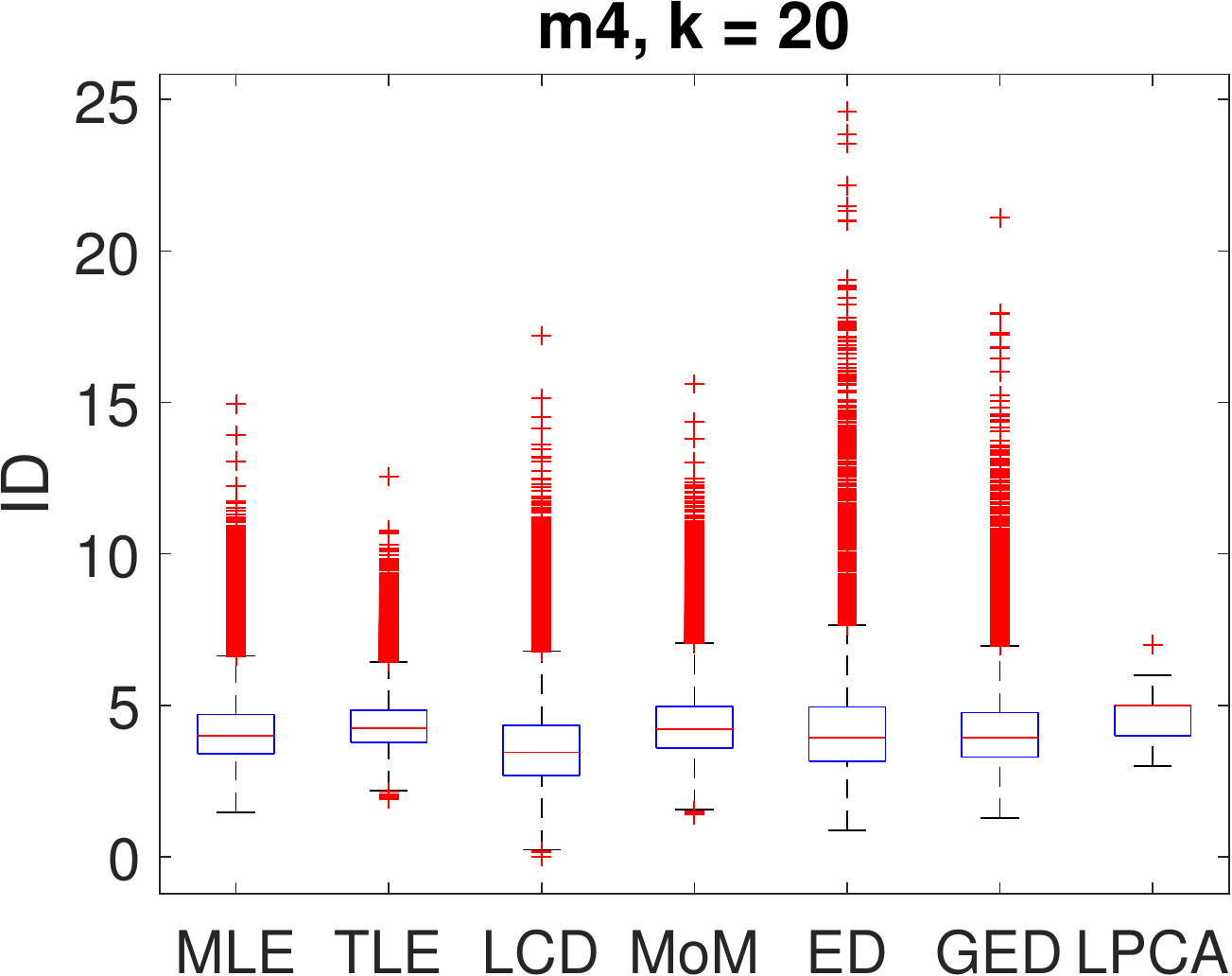}
\includegraphics[width=.30\textwidth]{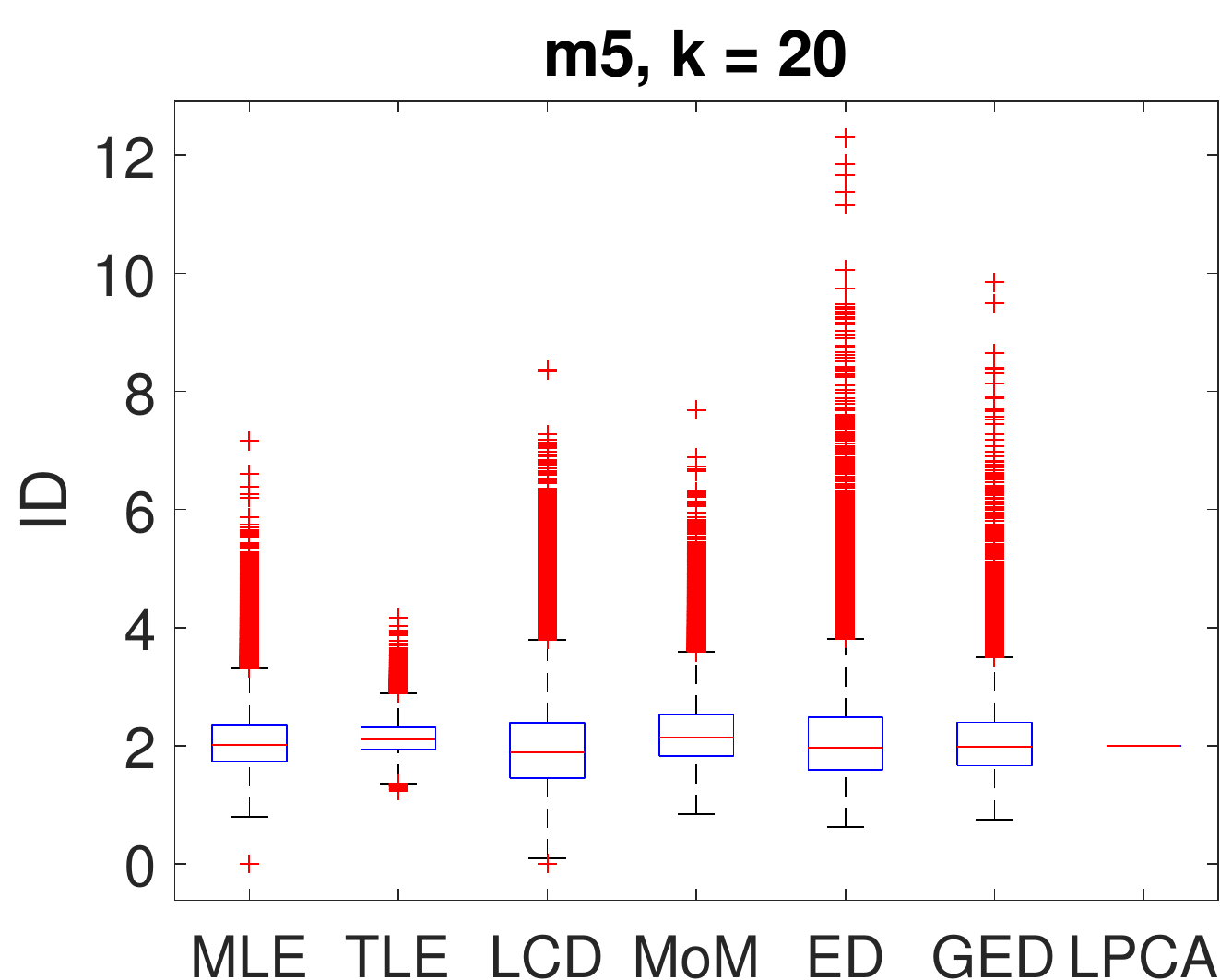}
\includegraphics[width=.30\textwidth]{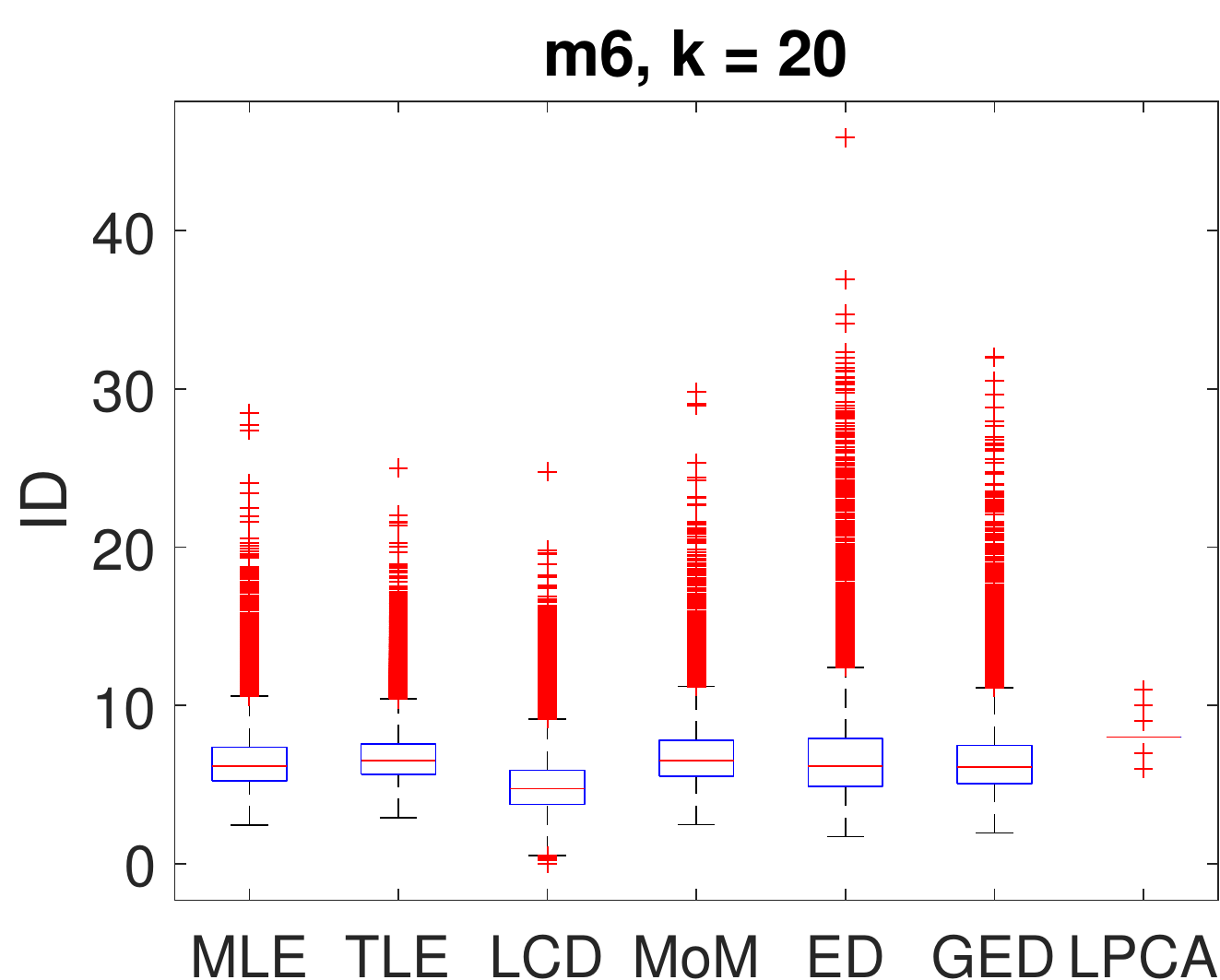}\\
\includegraphics[width=.30\textwidth]{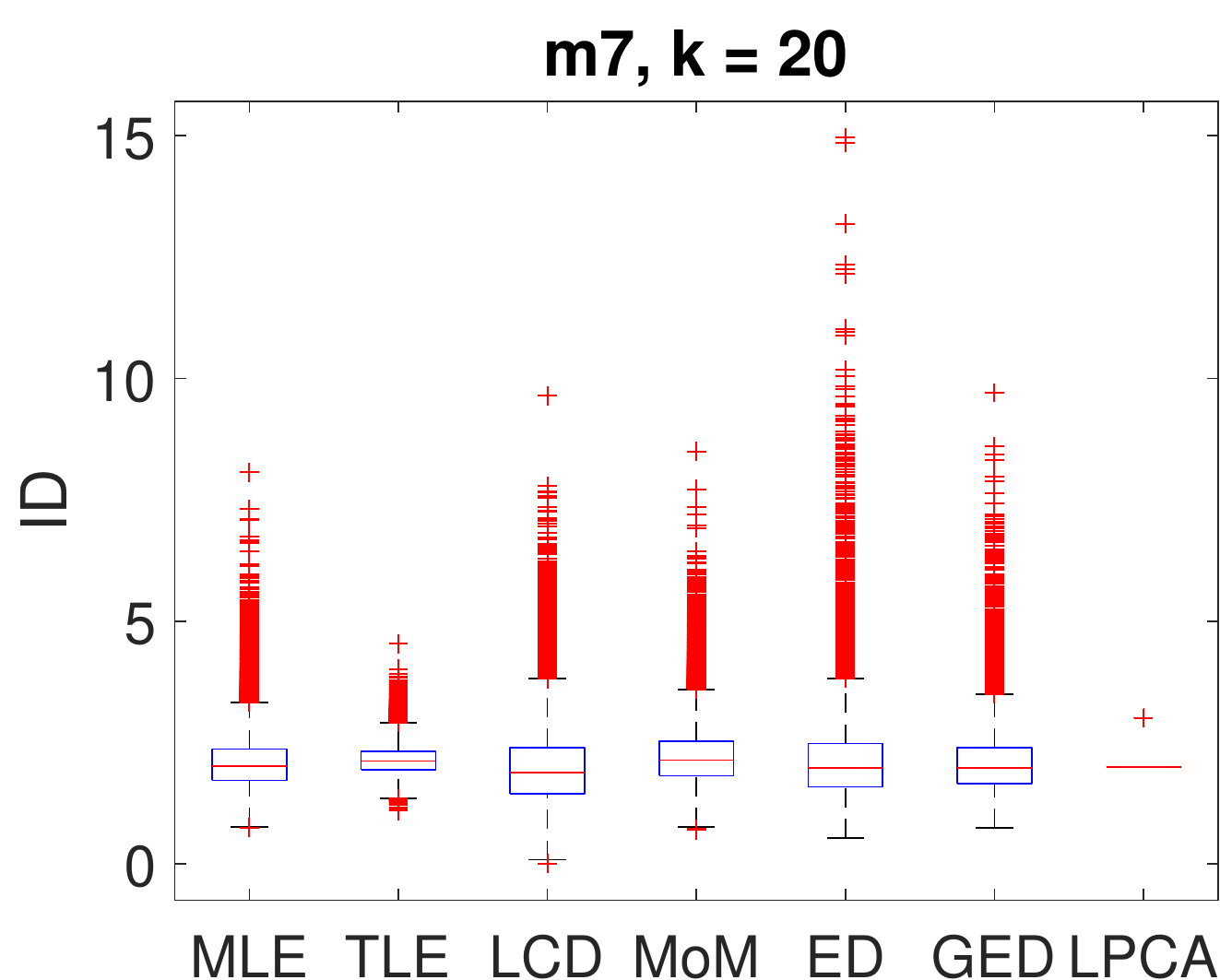}
\includegraphics[width=.30\textwidth]{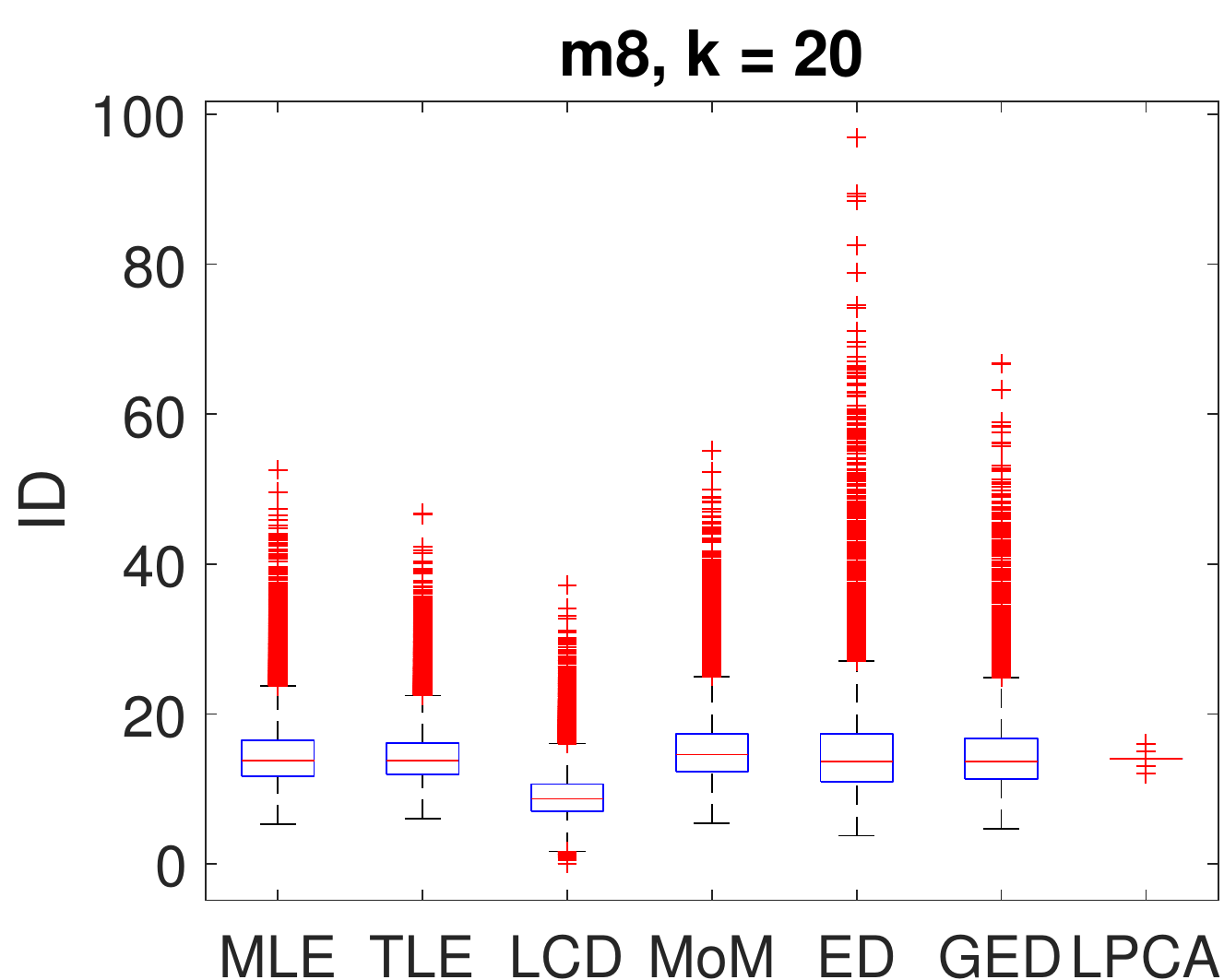}
\includegraphics[width=.30\textwidth]{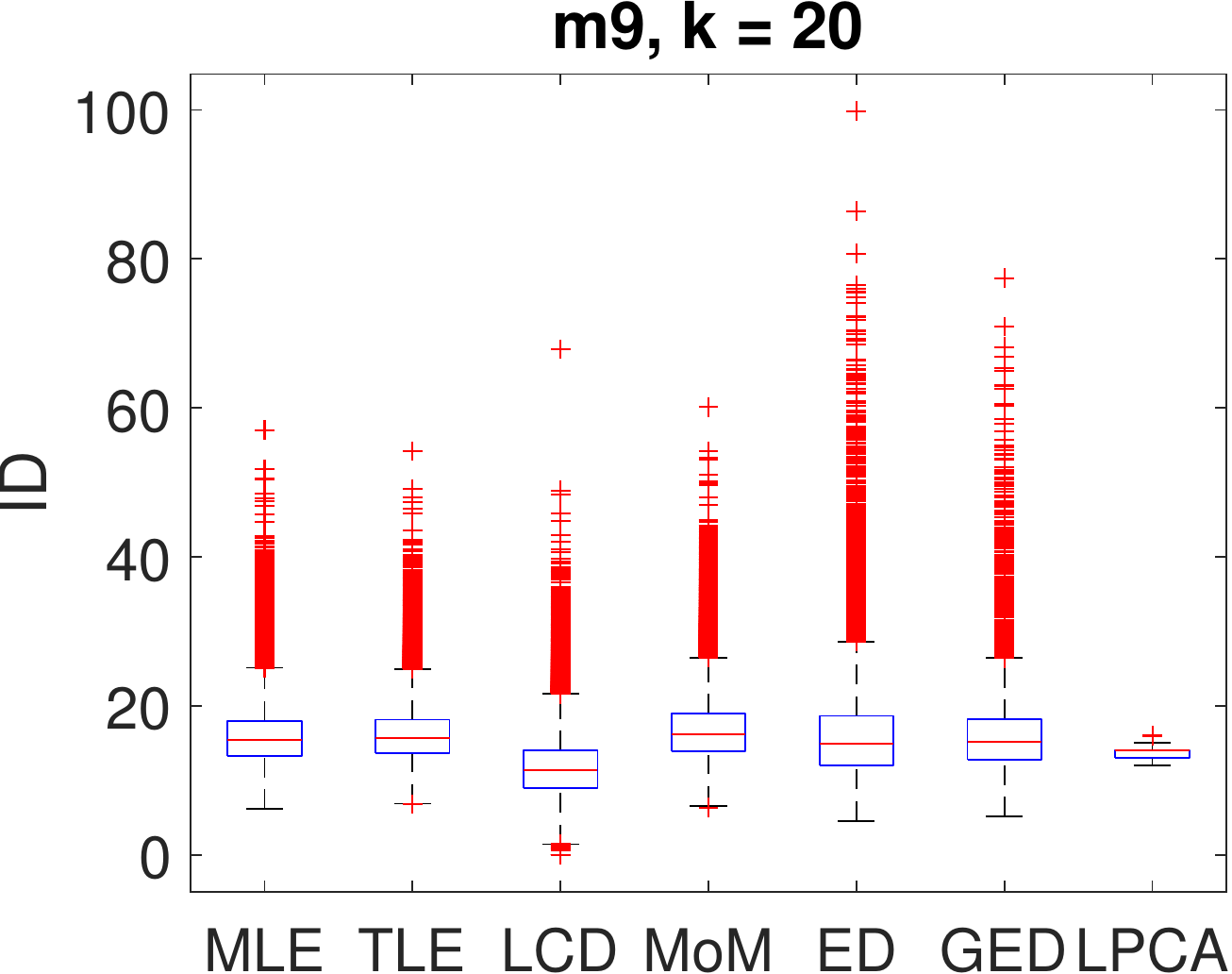}\\
\includegraphics[width=.30\textwidth]{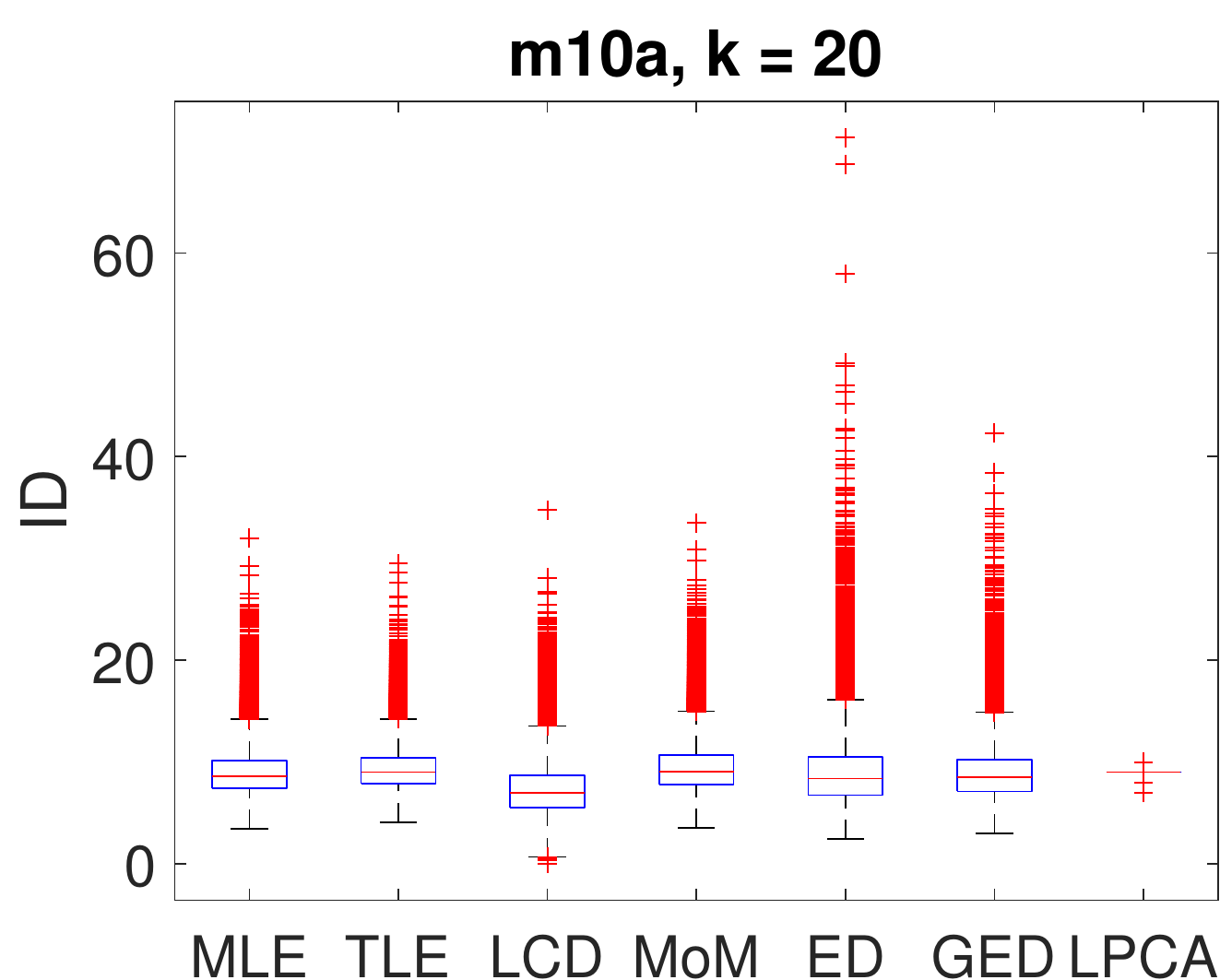}
\includegraphics[width=.30\textwidth]{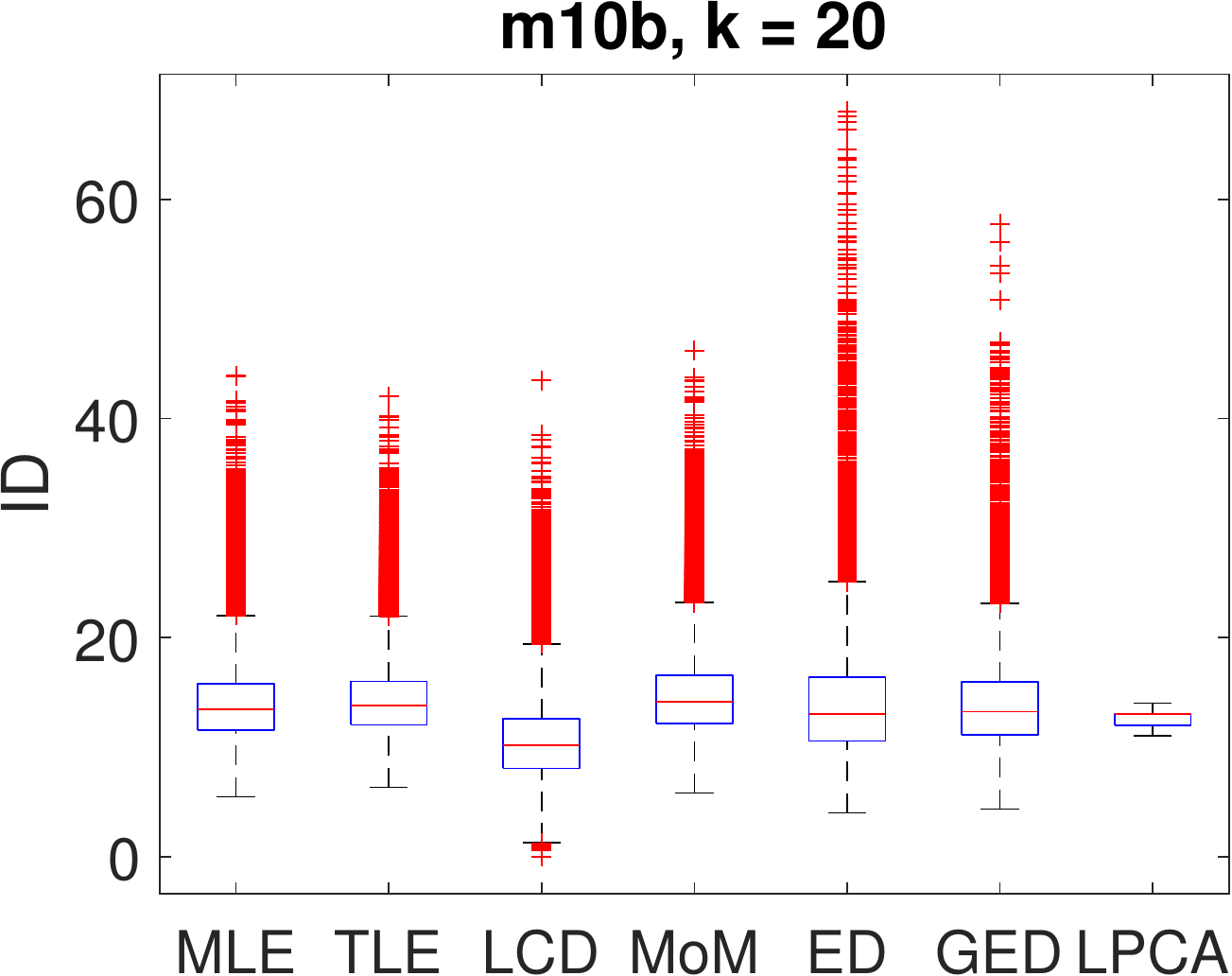}
\includegraphics[width=.30\textwidth]{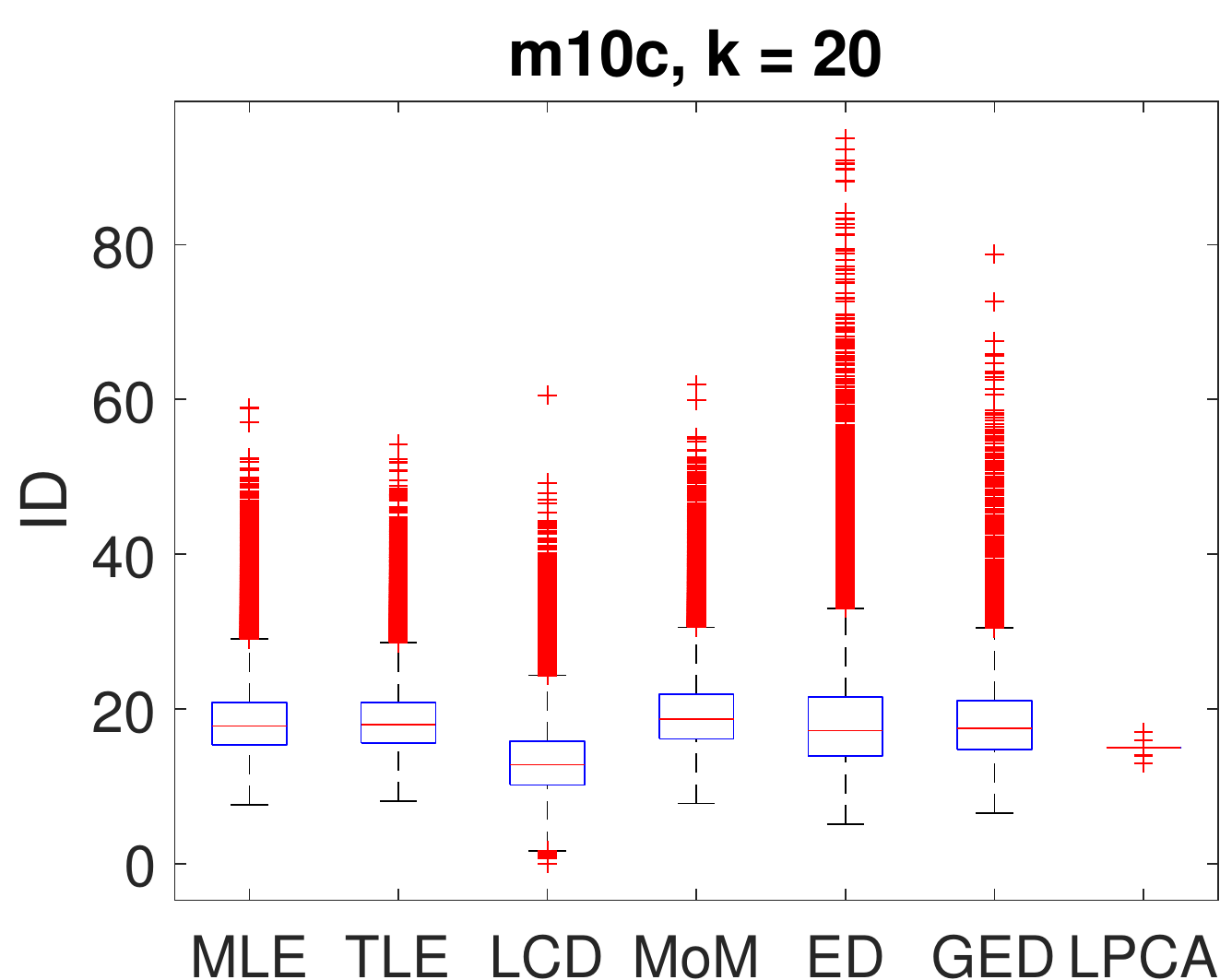}\\
\includegraphics[width=.30\textwidth]{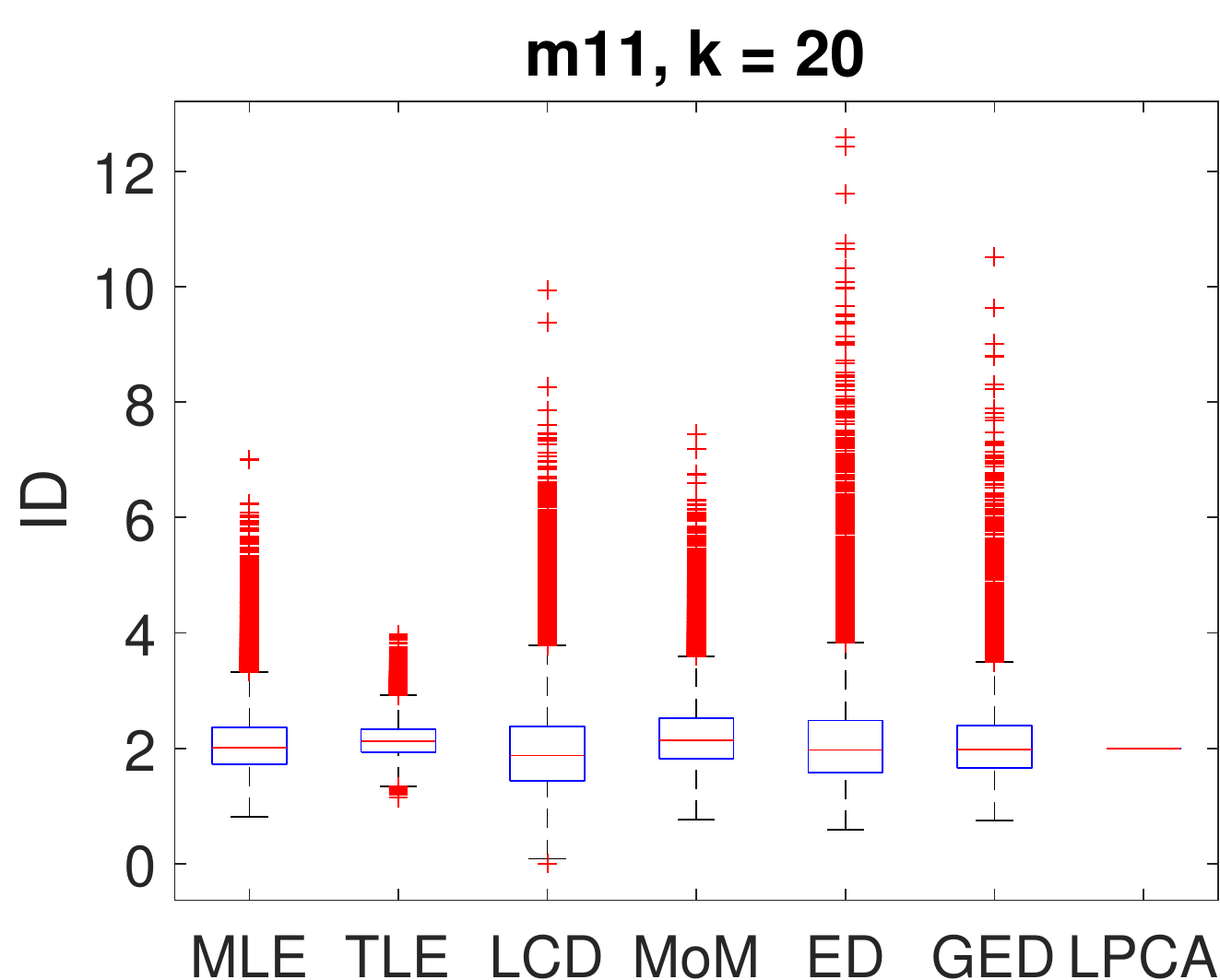}
\includegraphics[width=.30\textwidth]{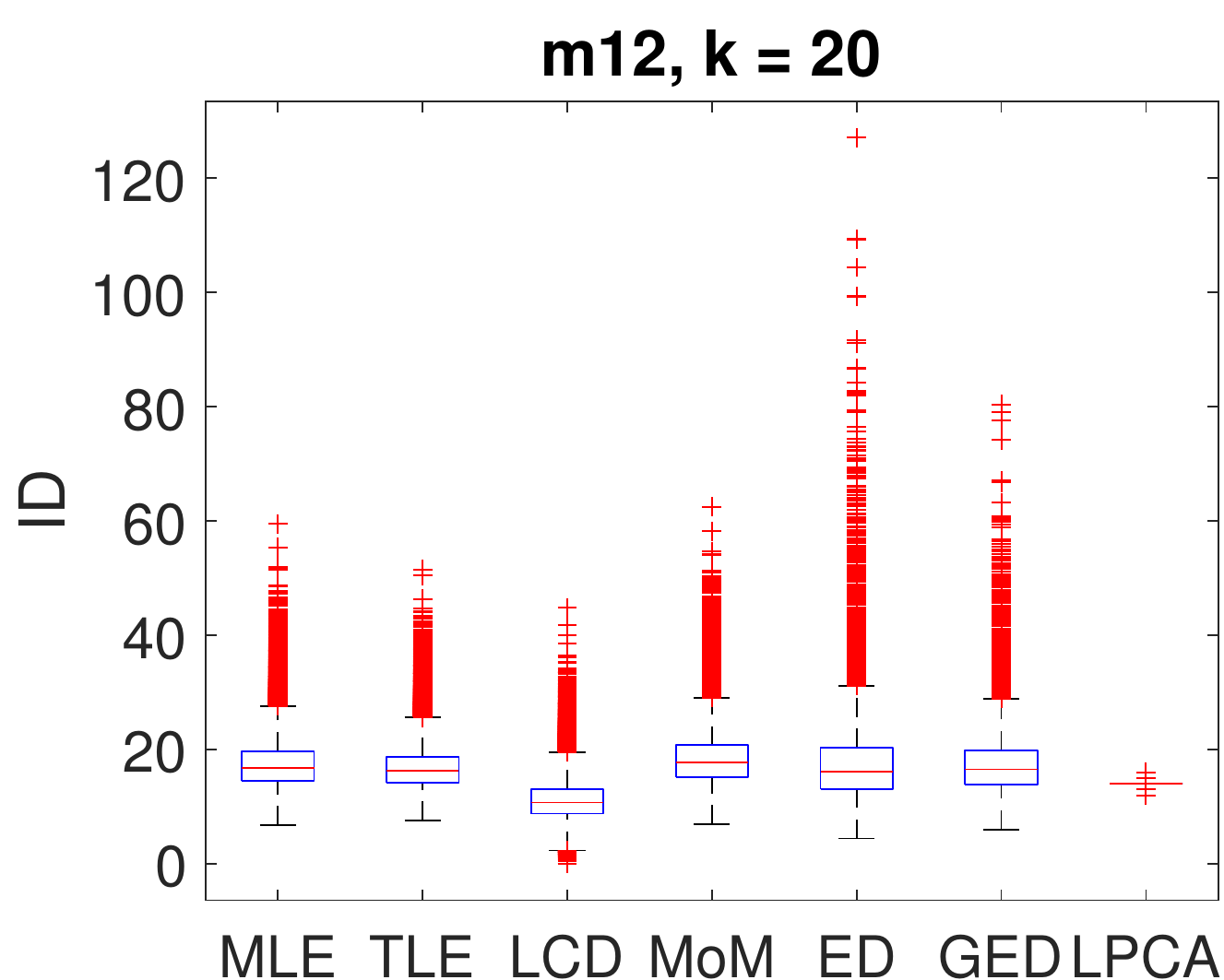}
\includegraphics[width=.30\textwidth]{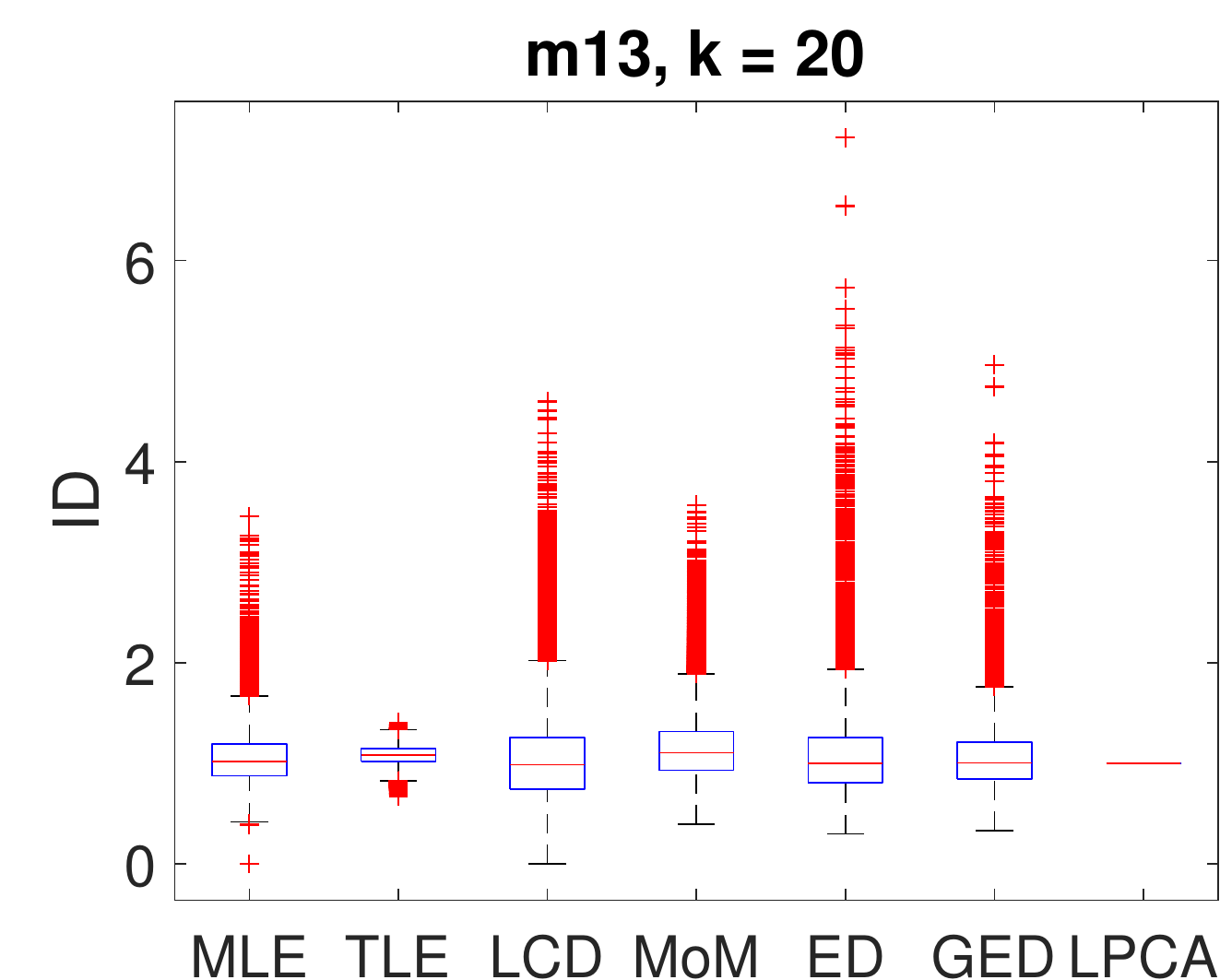}\\
\end{centering}
\caption{Box plots of estimated ID values, for neighborhood size 20, on various synthetic data.}\label{fig:synthm-boxplots-k20}
\end{figure*}

\begin{figure*}
\begin{centering}
\includegraphics[width=.30\textwidth]{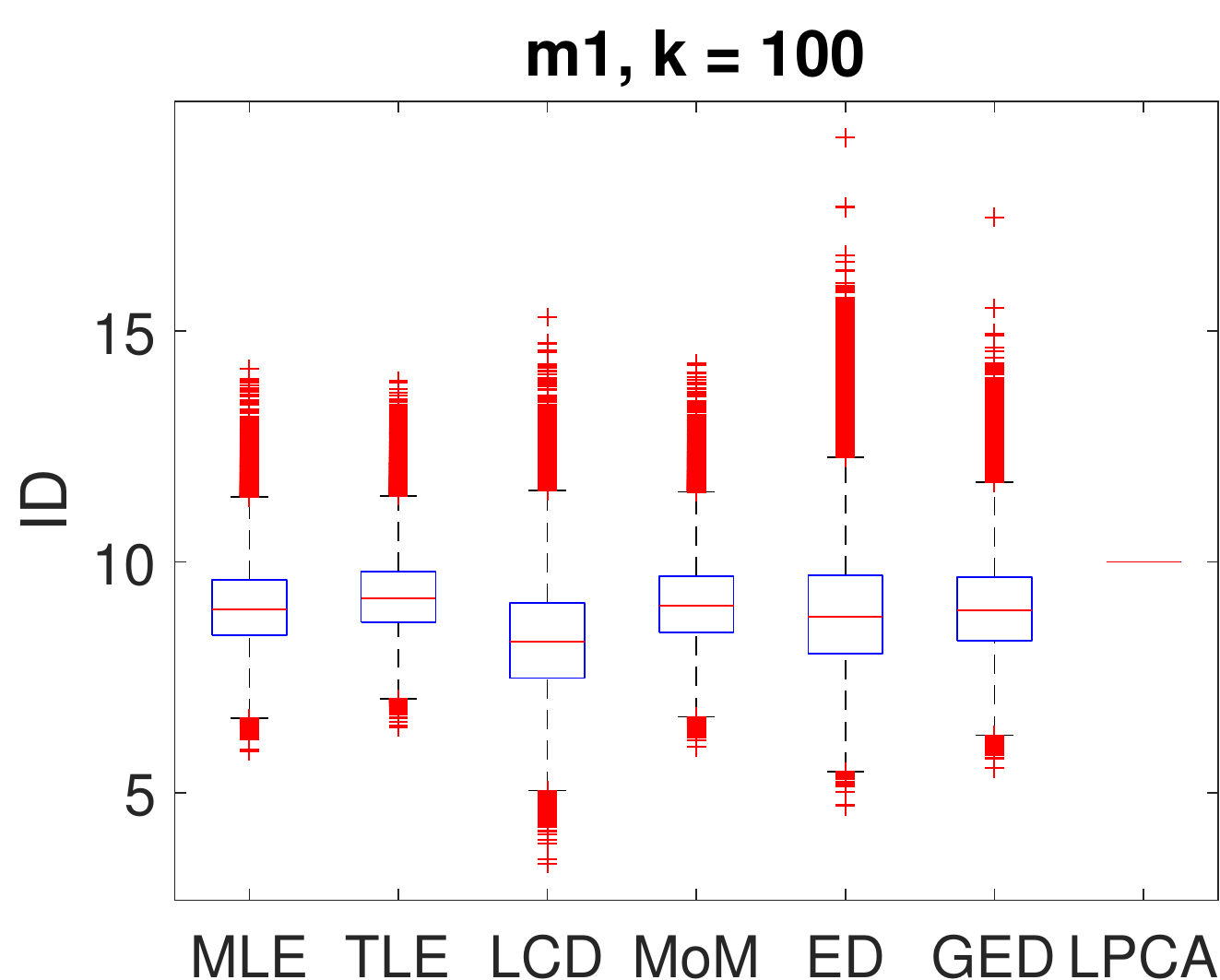}
\includegraphics[width=.30\textwidth]{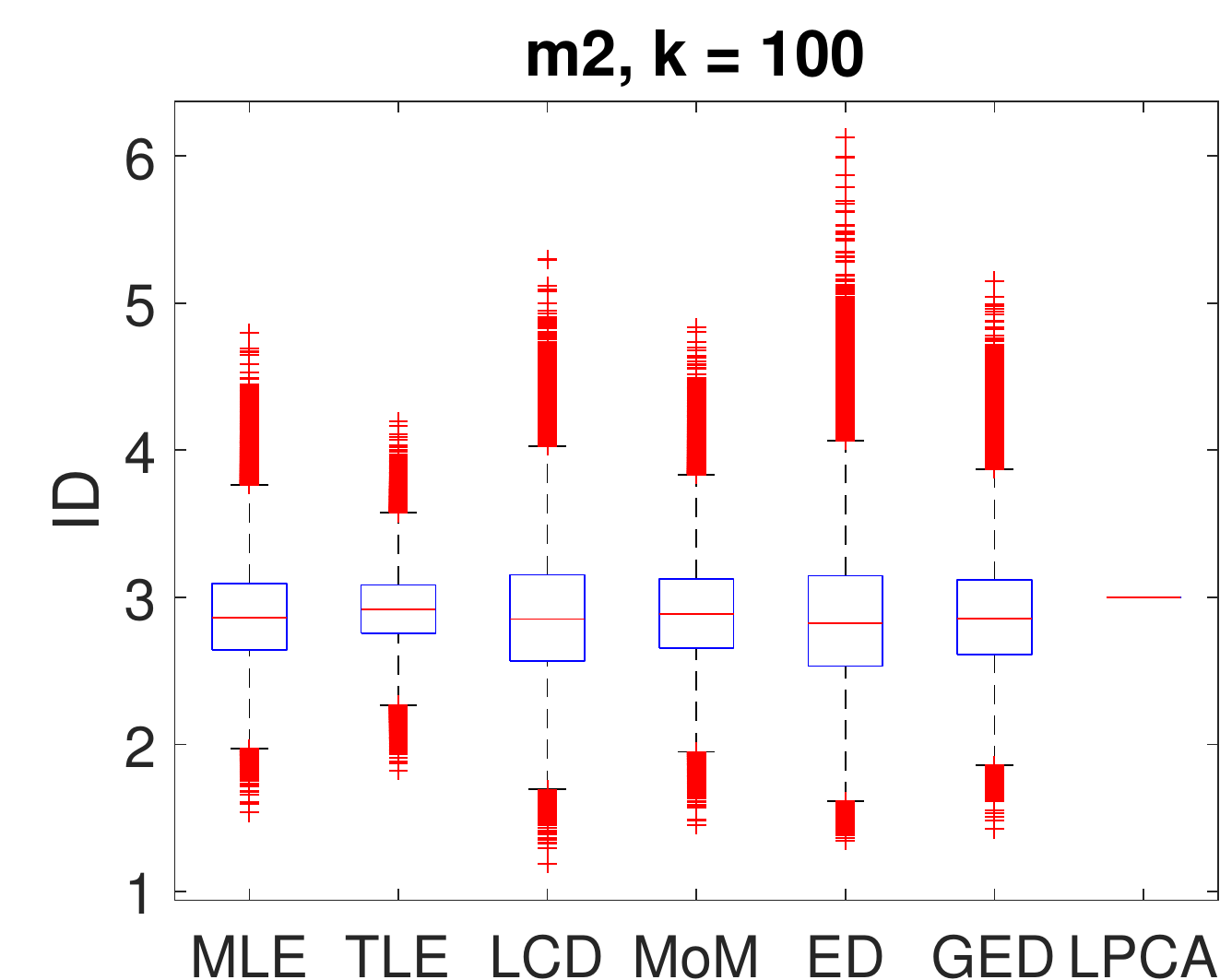}
\includegraphics[width=.30\textwidth]{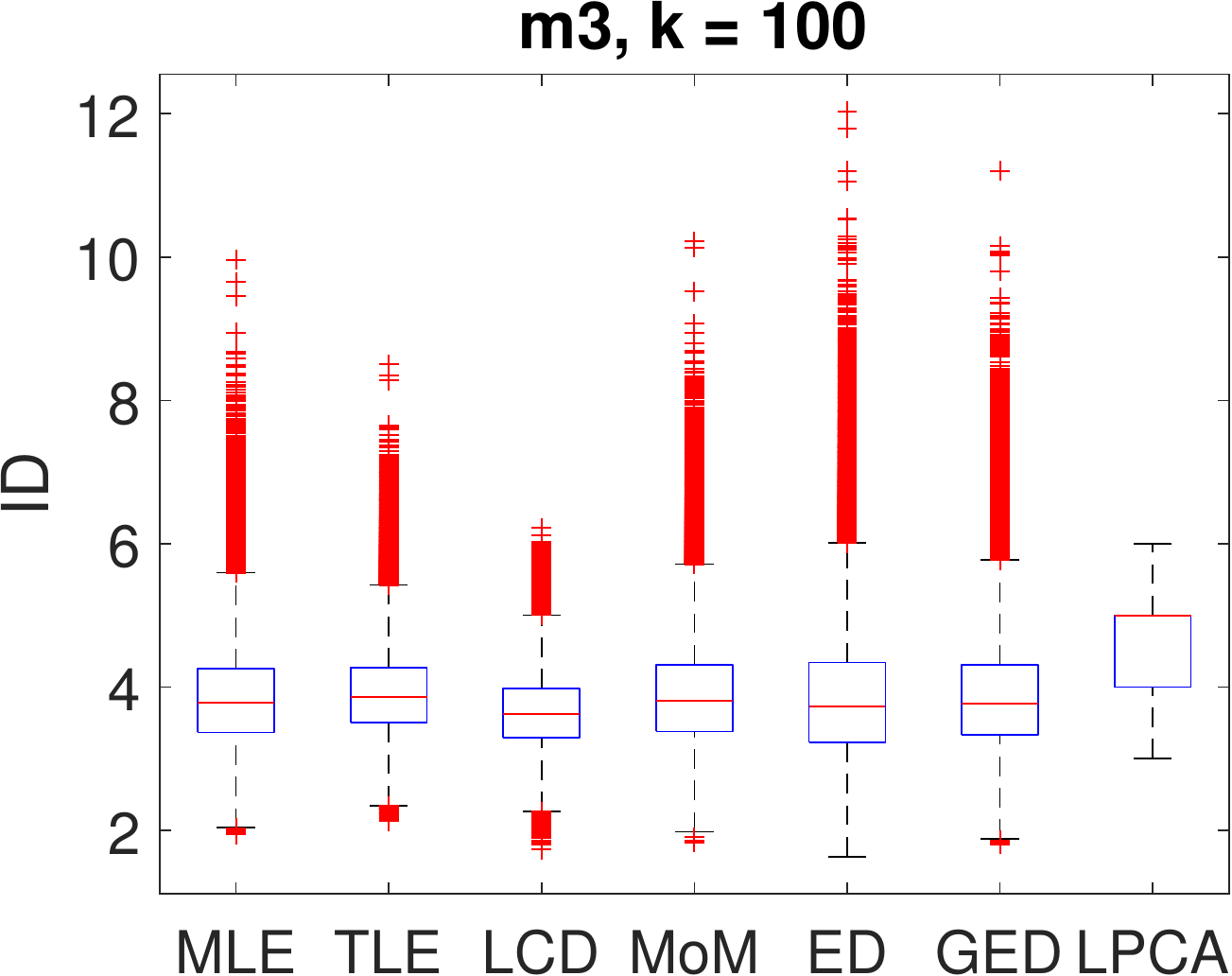}\\
\includegraphics[width=.30\textwidth]{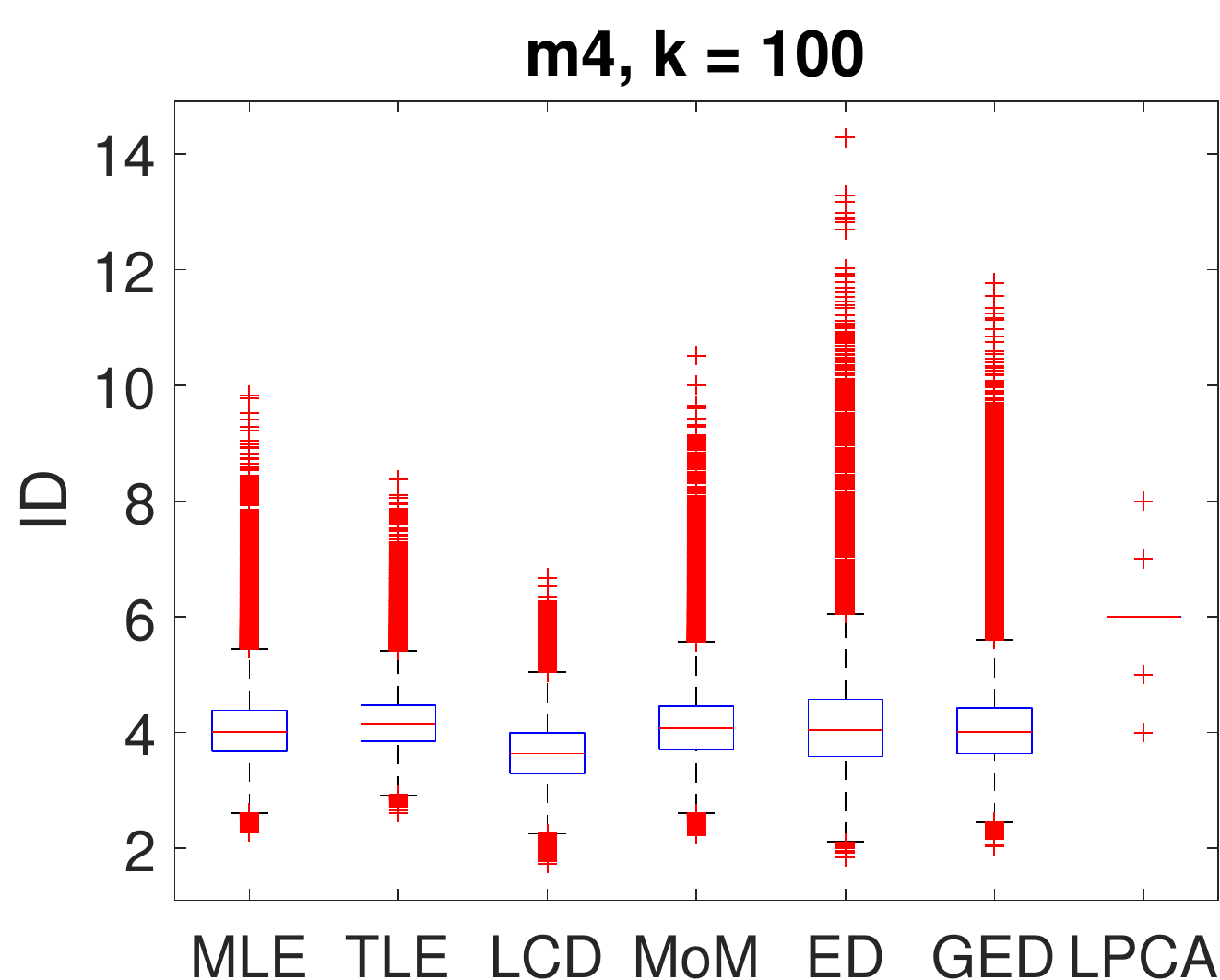}
\includegraphics[width=.30\textwidth]{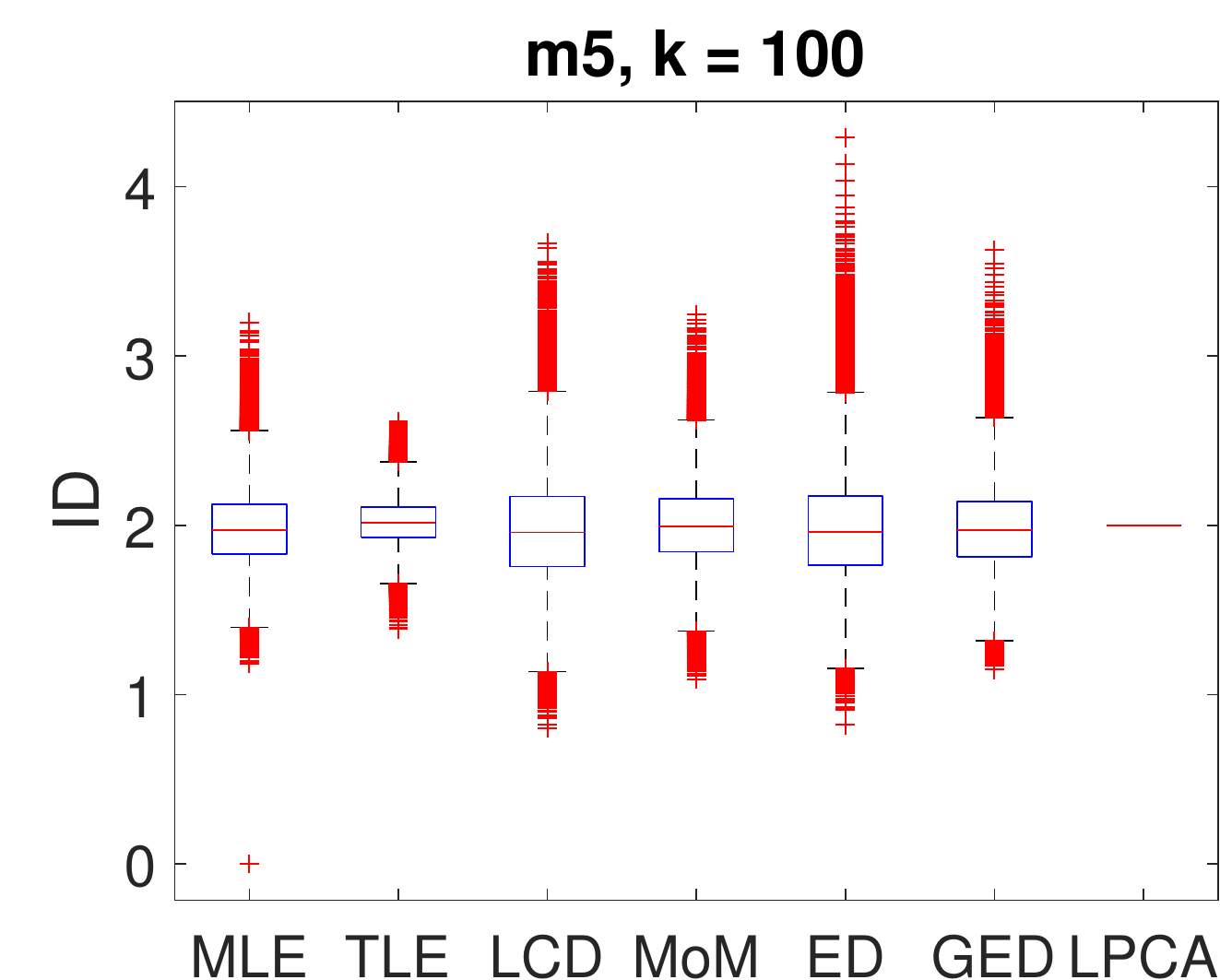}
\includegraphics[width=.30\textwidth]{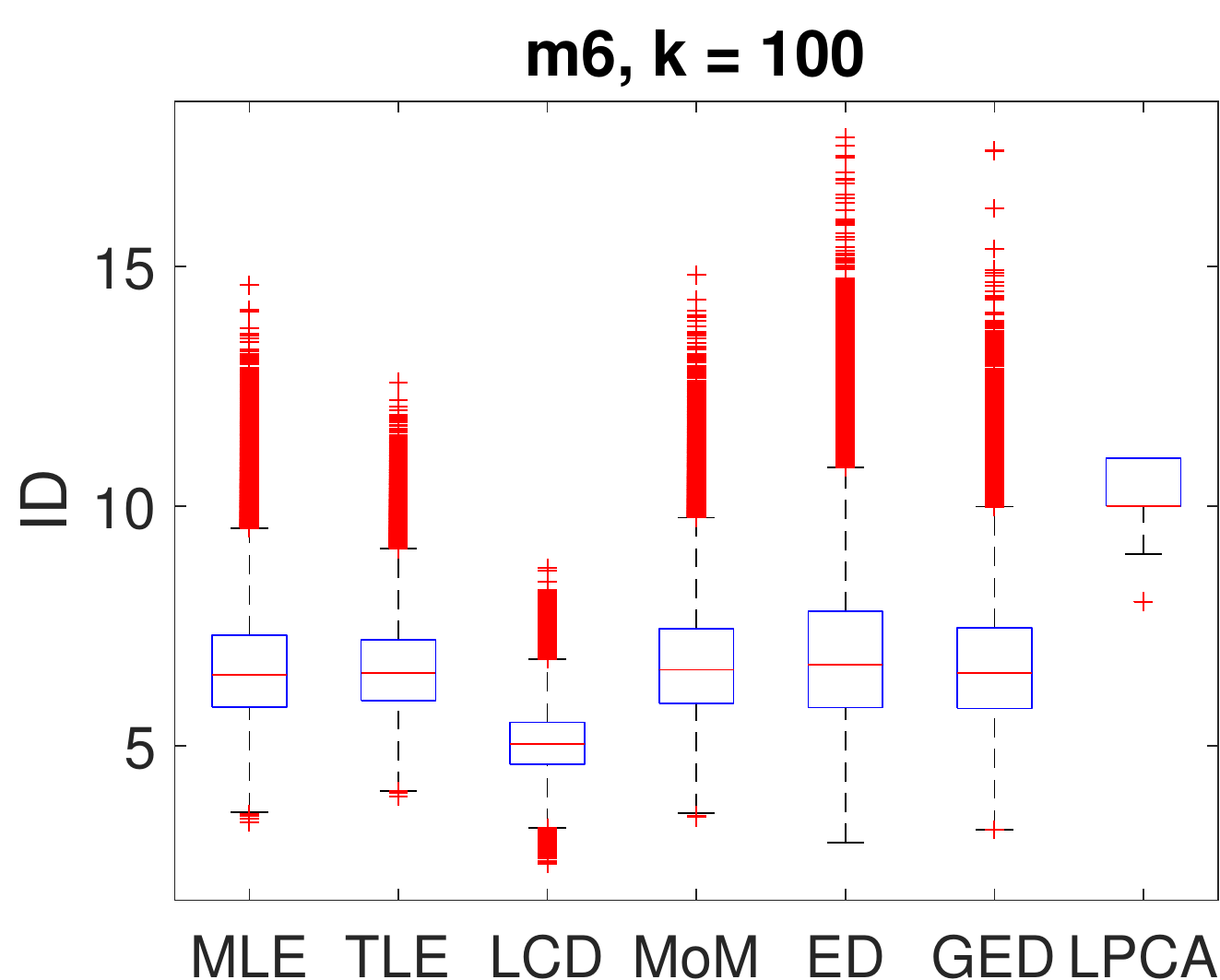}\\
\includegraphics[width=.30\textwidth]{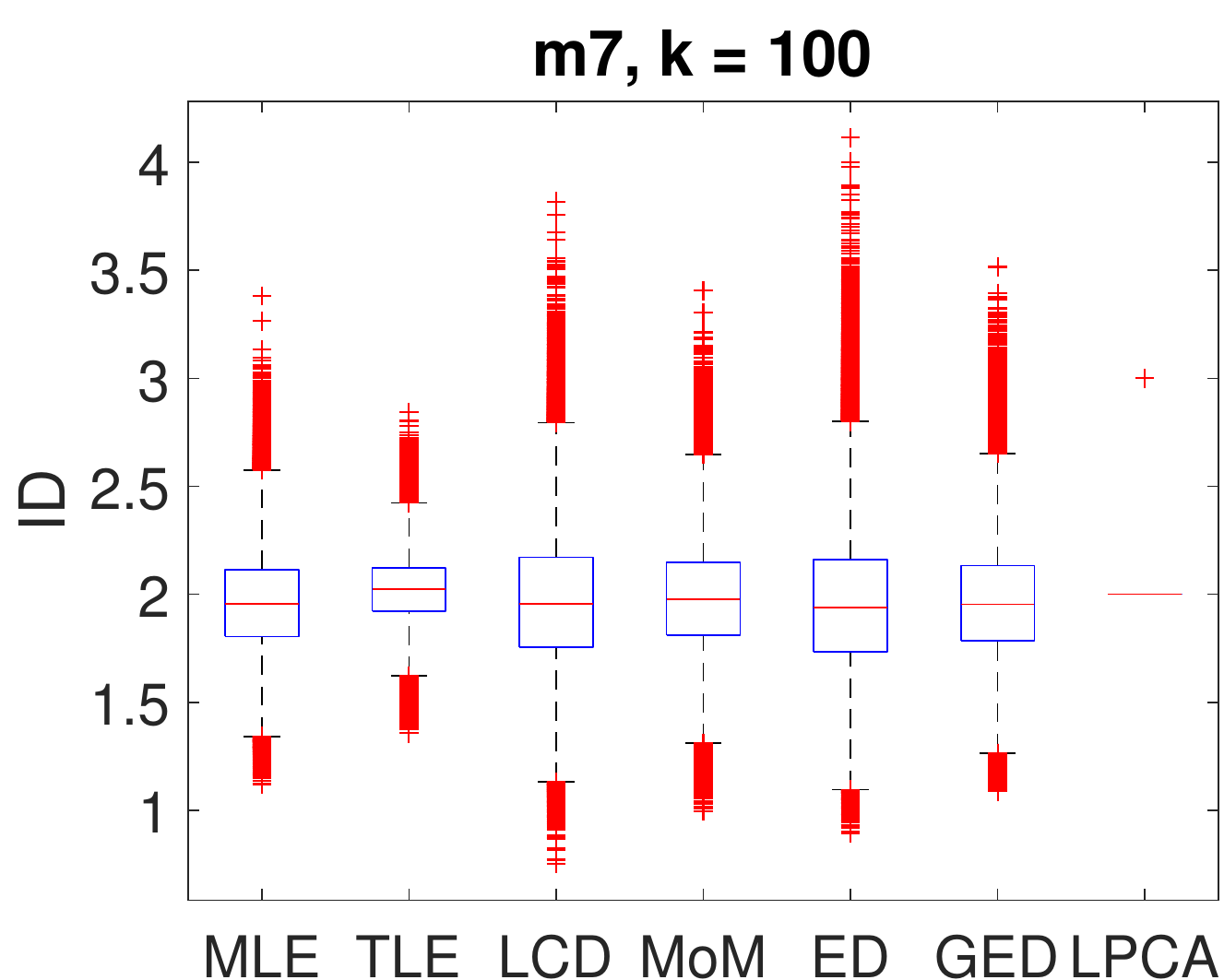}
\includegraphics[width=.30\textwidth]{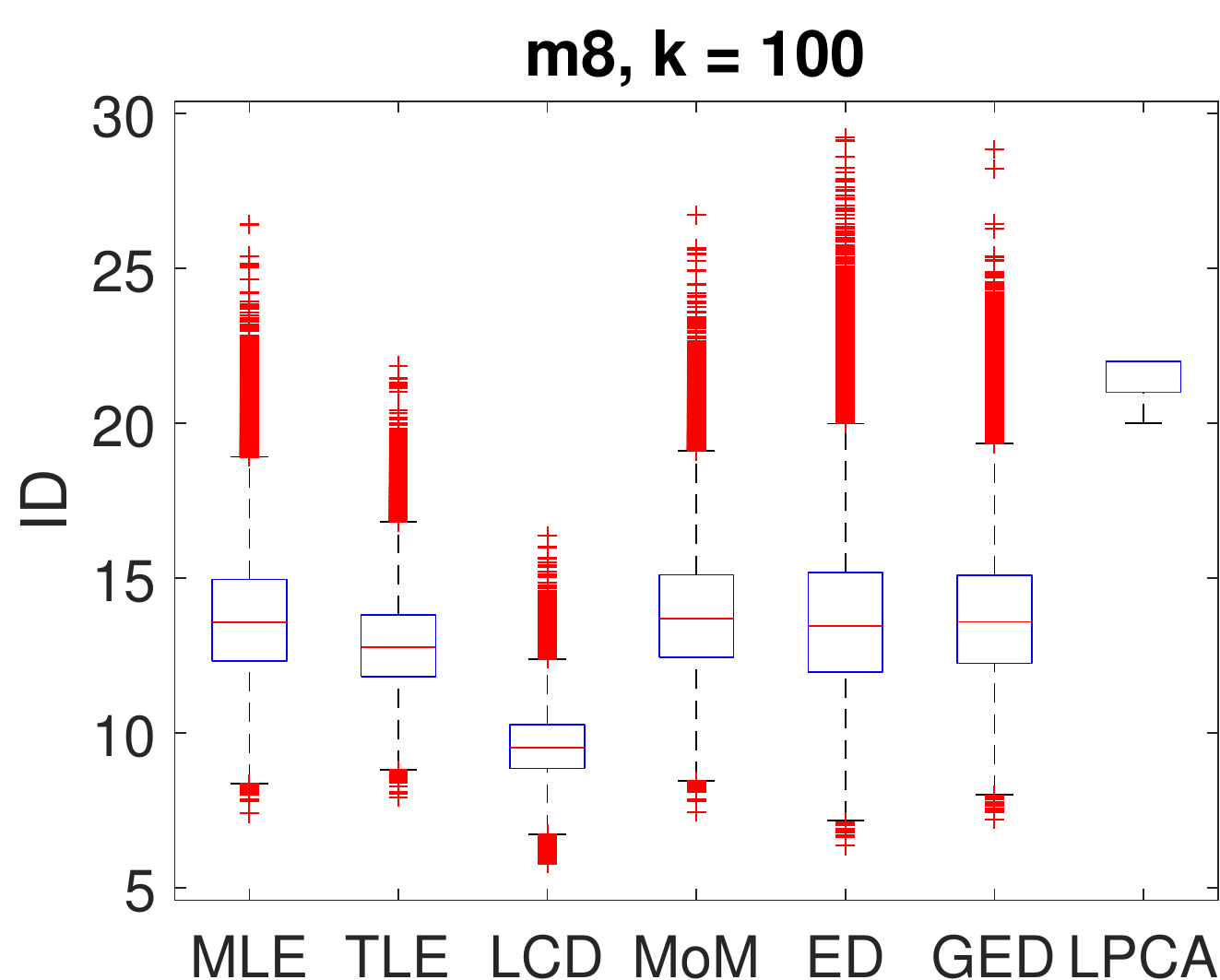}
\includegraphics[width=.30\textwidth]{fig/synth-m/boxplots/boxplot-m9-k20.pdf}\\
\includegraphics[width=.30\textwidth]{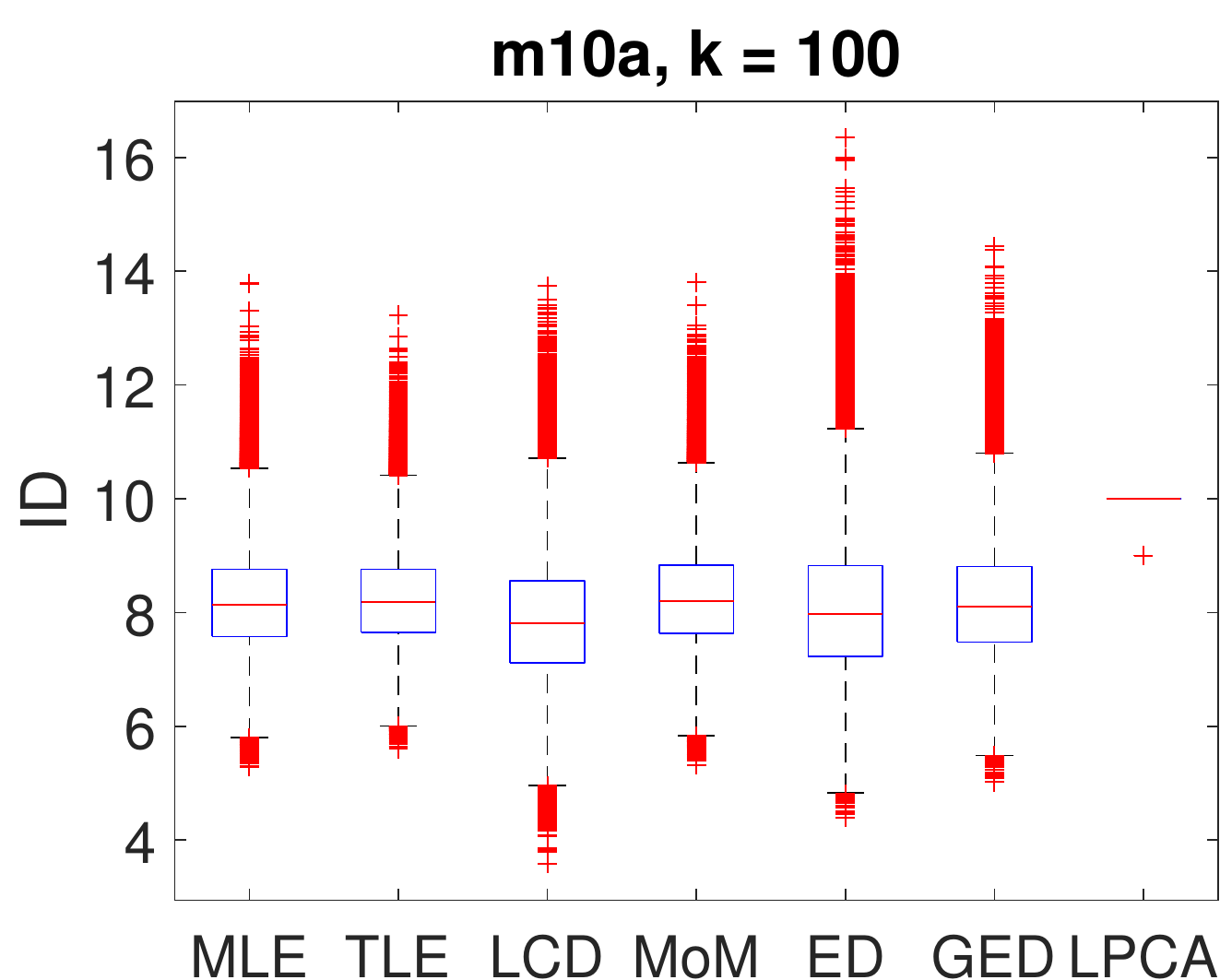}
\includegraphics[width=.30\textwidth]{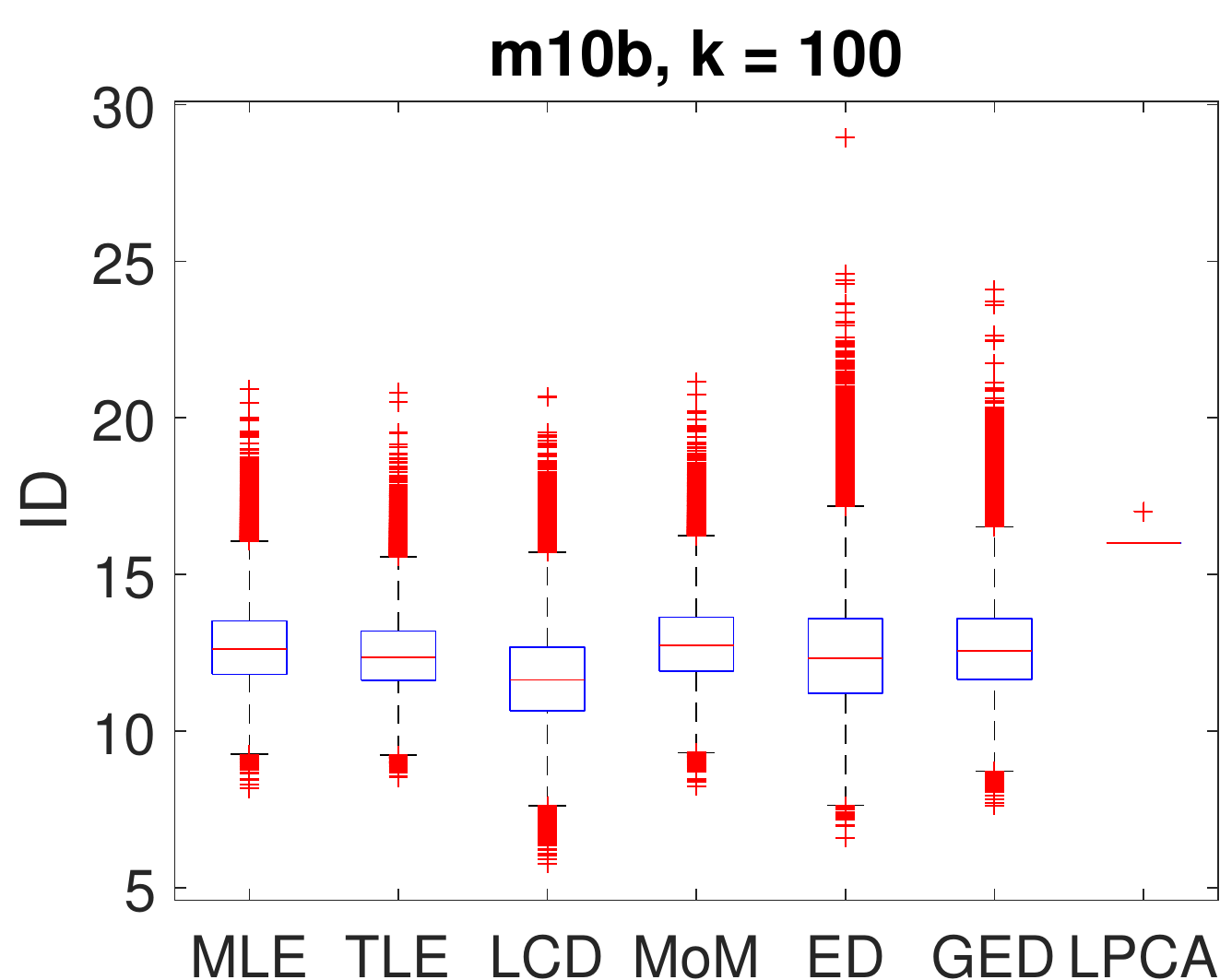}
\includegraphics[width=.30\textwidth]{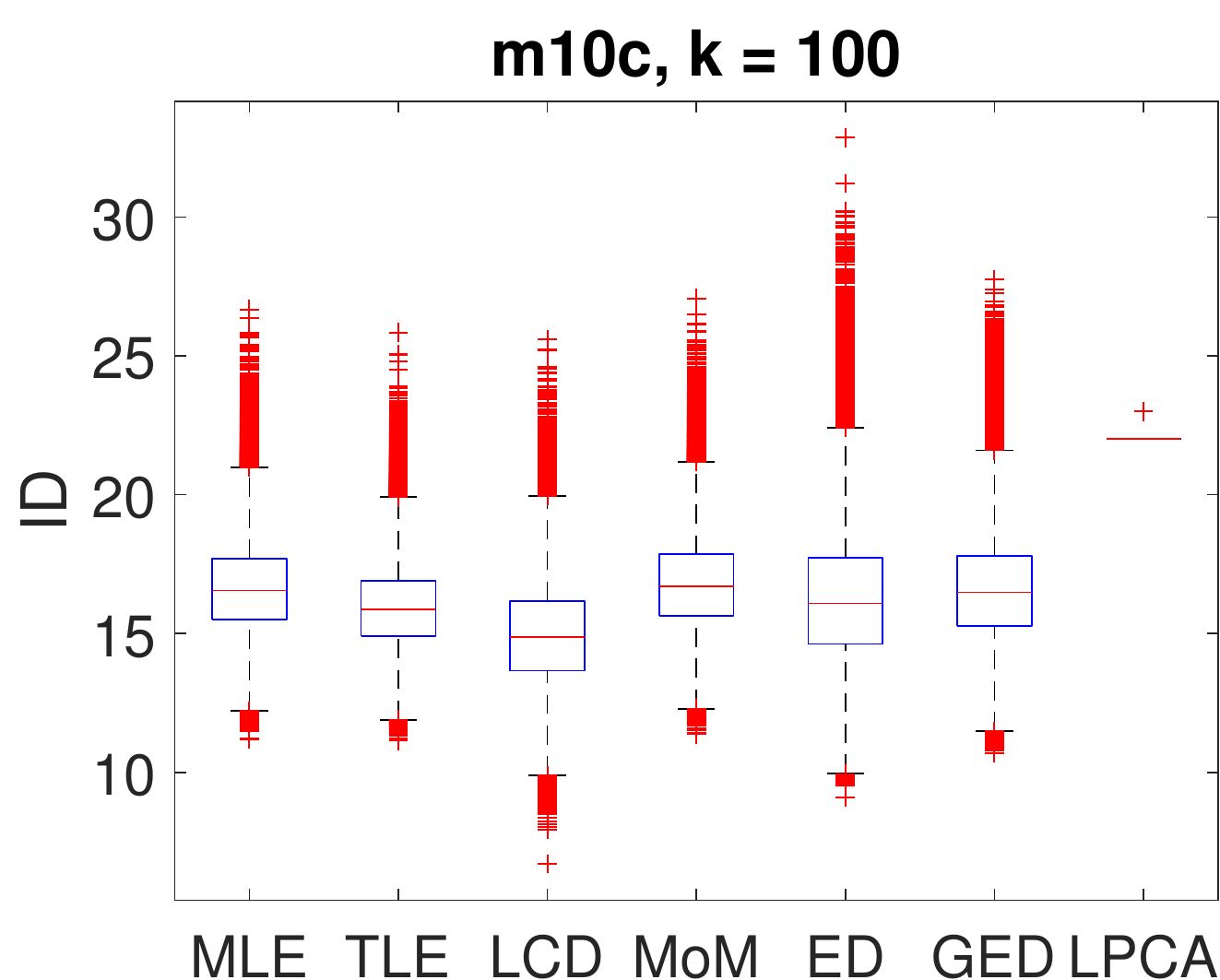}\\
\includegraphics[width=.30\textwidth]{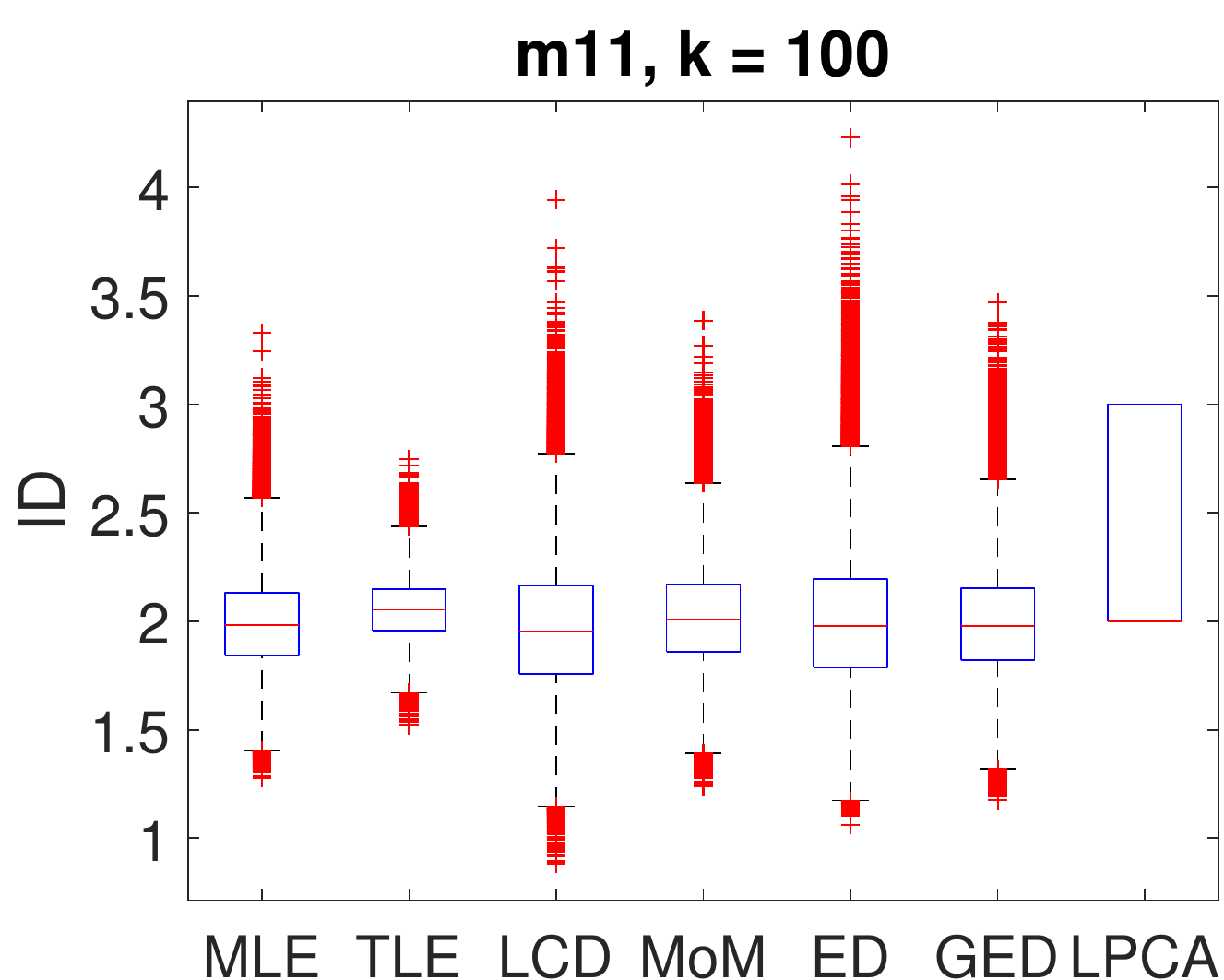}
\includegraphics[width=.30\textwidth]{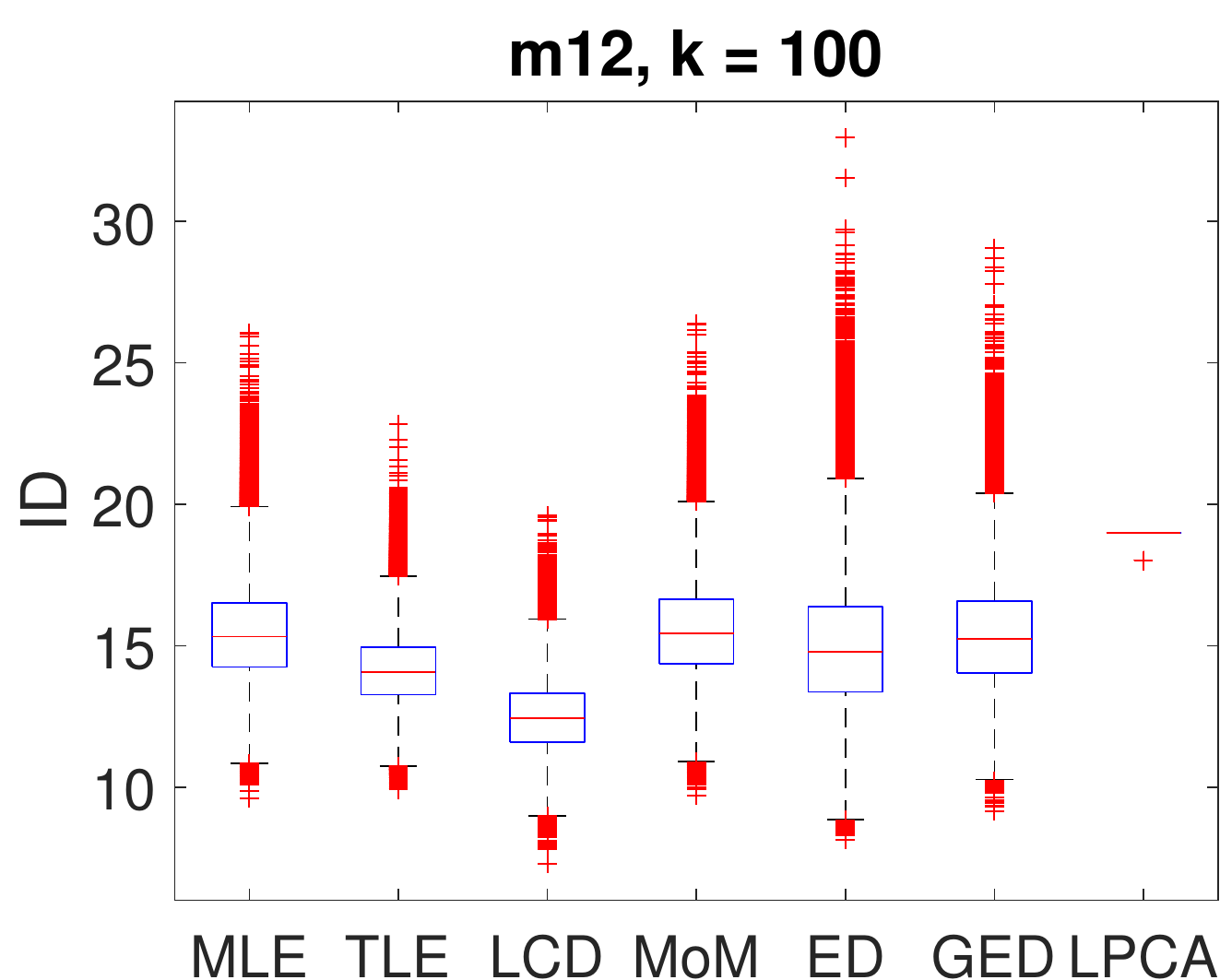}
\includegraphics[width=.30\textwidth]{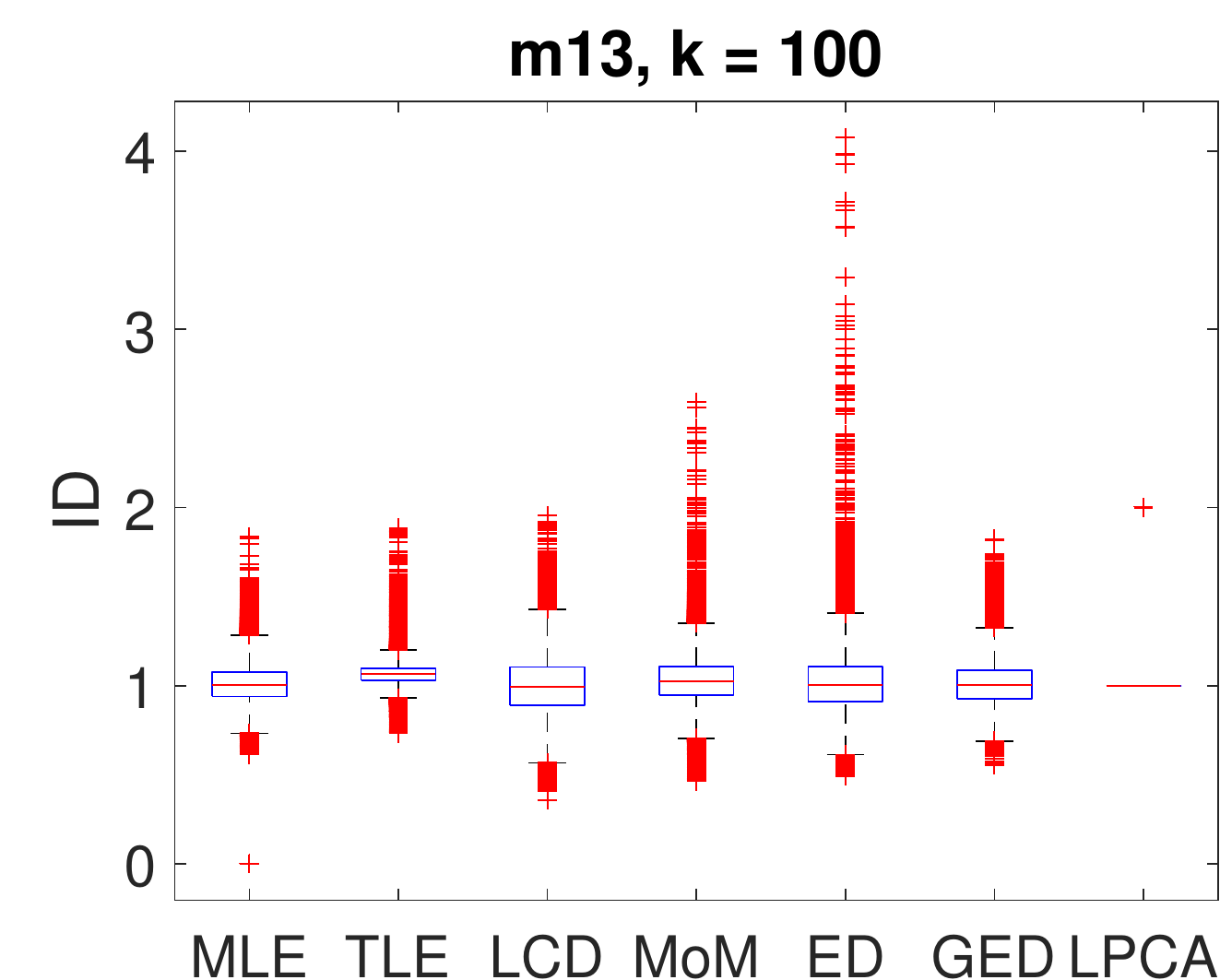}\\
\end{centering}
\caption{Box plots of estimated ID values, for neighborhood size 100, on various synthetic data.}\label{fig:synthm-boxplots-k100}
\end{figure*}

Figures~\ref{fig:synthm-boxplots-k20} and~\ref{fig:synthm-boxplots-k100}
support the above observations on various other synthetic data distributed within manifolds drawn from the
m-family. Box plots are used here to illustrate in more detail the
distributions of ID estimates. The neighborhood size is set to $k = 20$ and $k =
100$, respectively, with analogous trends observed for other $k$ values.
Besides emphasizing that TLE exhibits comparable bias and small
variance, the box plots also show that TLE tends to produce a smaller number
of outlying ID estimates, which are usually less pronounced. It is
interesting to note that these advantages of TLE are more visible for lower
values of~$k$, which underscores our observations of the successful use of TLE
in scenarios involving smaller neighborhood size.

\subsection{Real Data.}

%Figure~\ref{fig:real-boxplots-mutik} shows the distributions of ID estimates
%for real data sets, with two box plots for each data set, using neighborhood
%sizes $k = 20, 50$. In addition to supporting the observations made on
%synthetic data sets, it can be stressed that variance of TLE for $k = 20$ is
%usually as good as or superior to the variance of other methods for larger
%values of~$k$ (see supplement for box plots with $k = 100$). In the case when
%some other method (e.g.\ LCD) exhibits smaller variance, it usually has much
%worse bias, as we have also seen on synthetic data. An extreme example of
%this behavior is LPCA, which on real data often exhibits tight variance but
%positive bias that increases significantly with neighborhood size~$k$.

\begin{figure*}
\begin{centering}
\includegraphics[width=.32\textwidth]{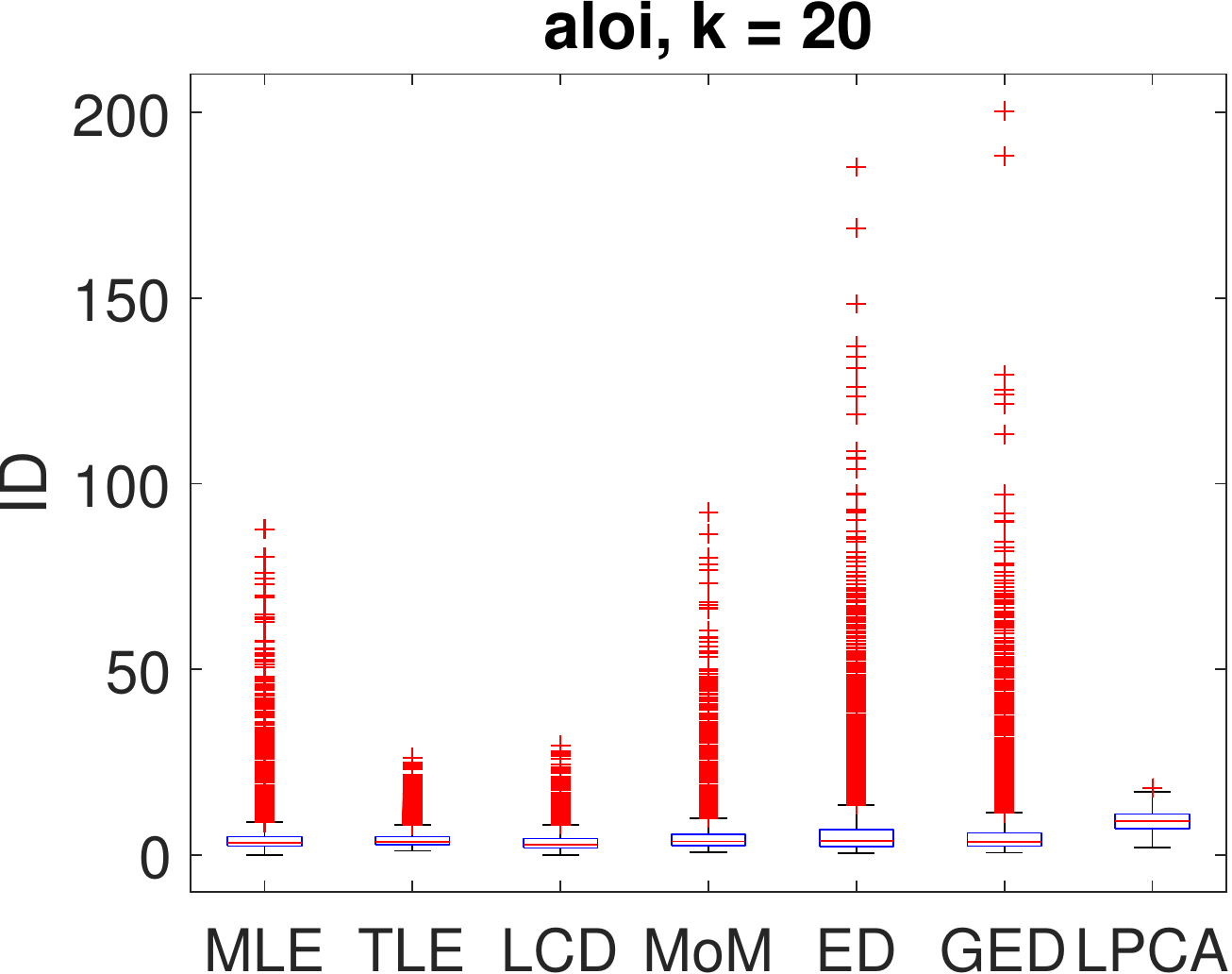}
\includegraphics[width=.32\textwidth]{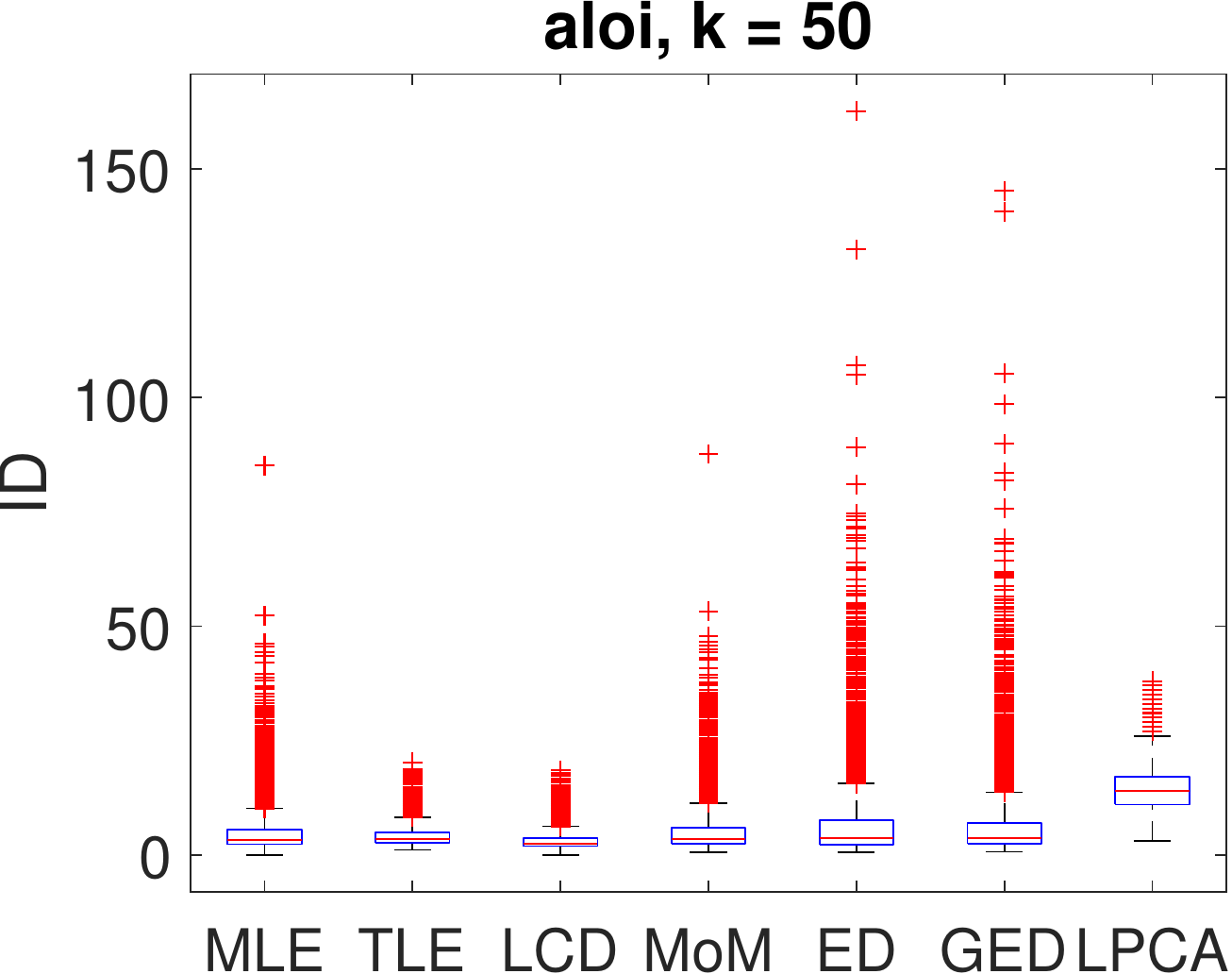}
\includegraphics[width=.32\textwidth]{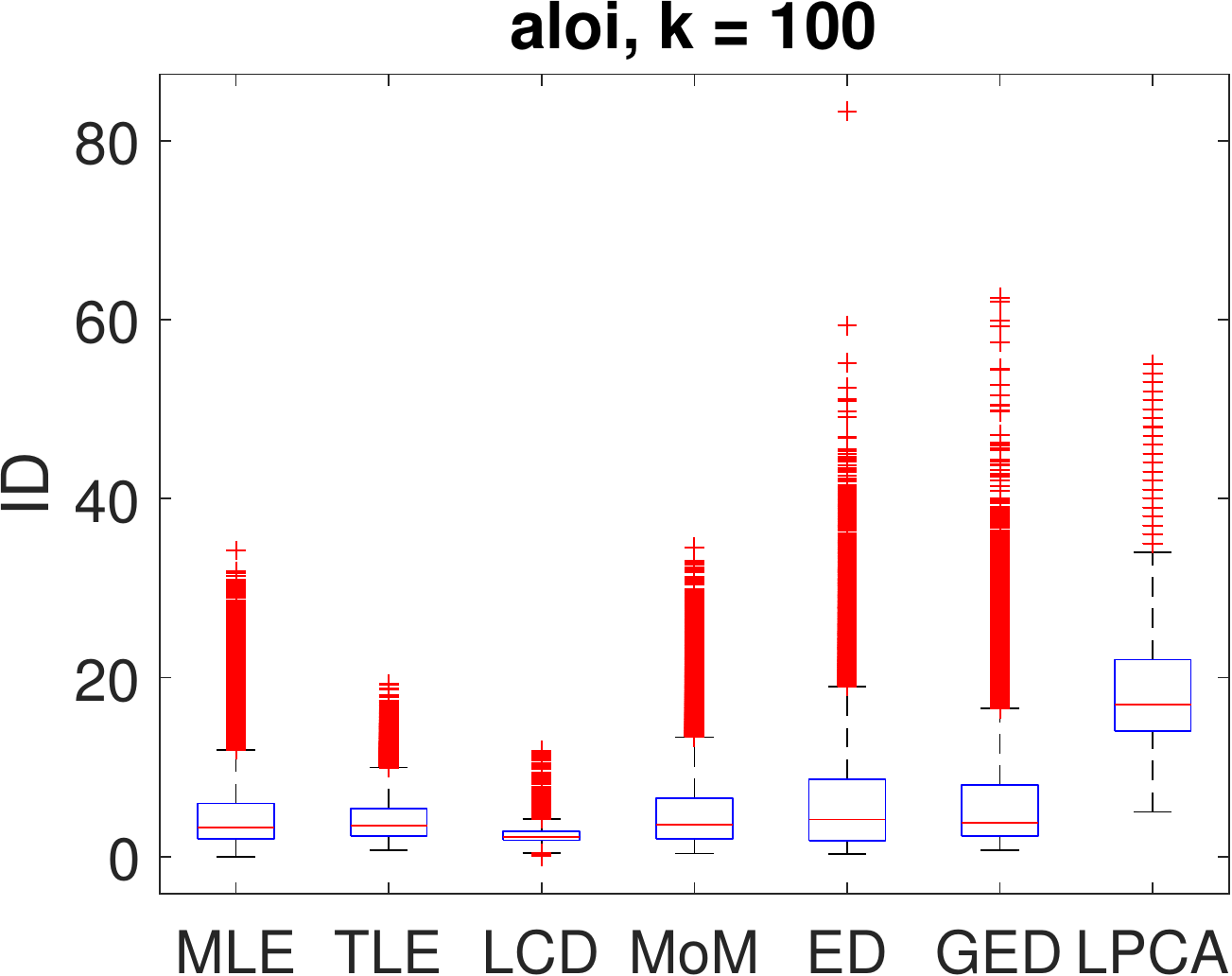}\\
\includegraphics[width=.32\textwidth]{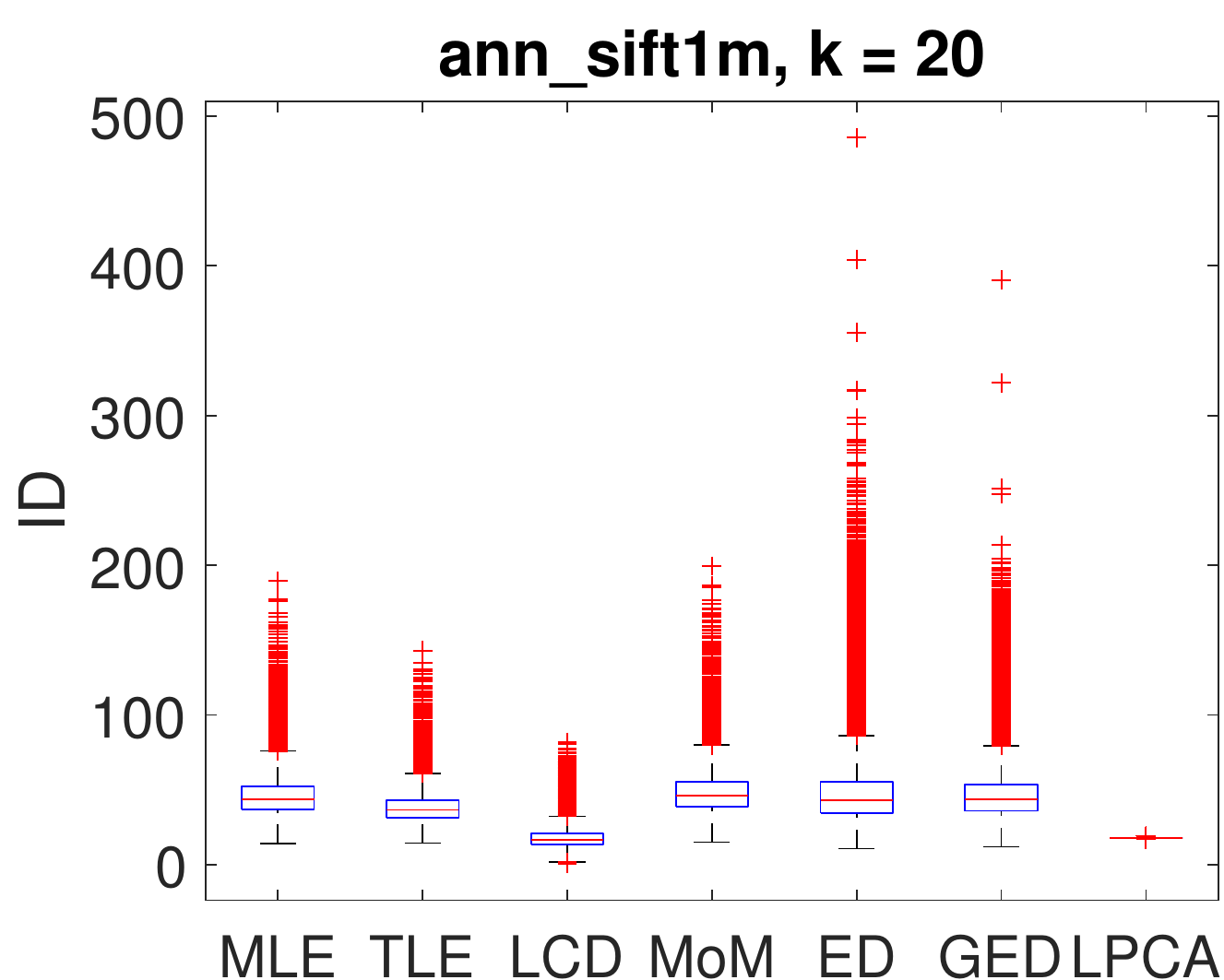}
\includegraphics[width=.32\textwidth]{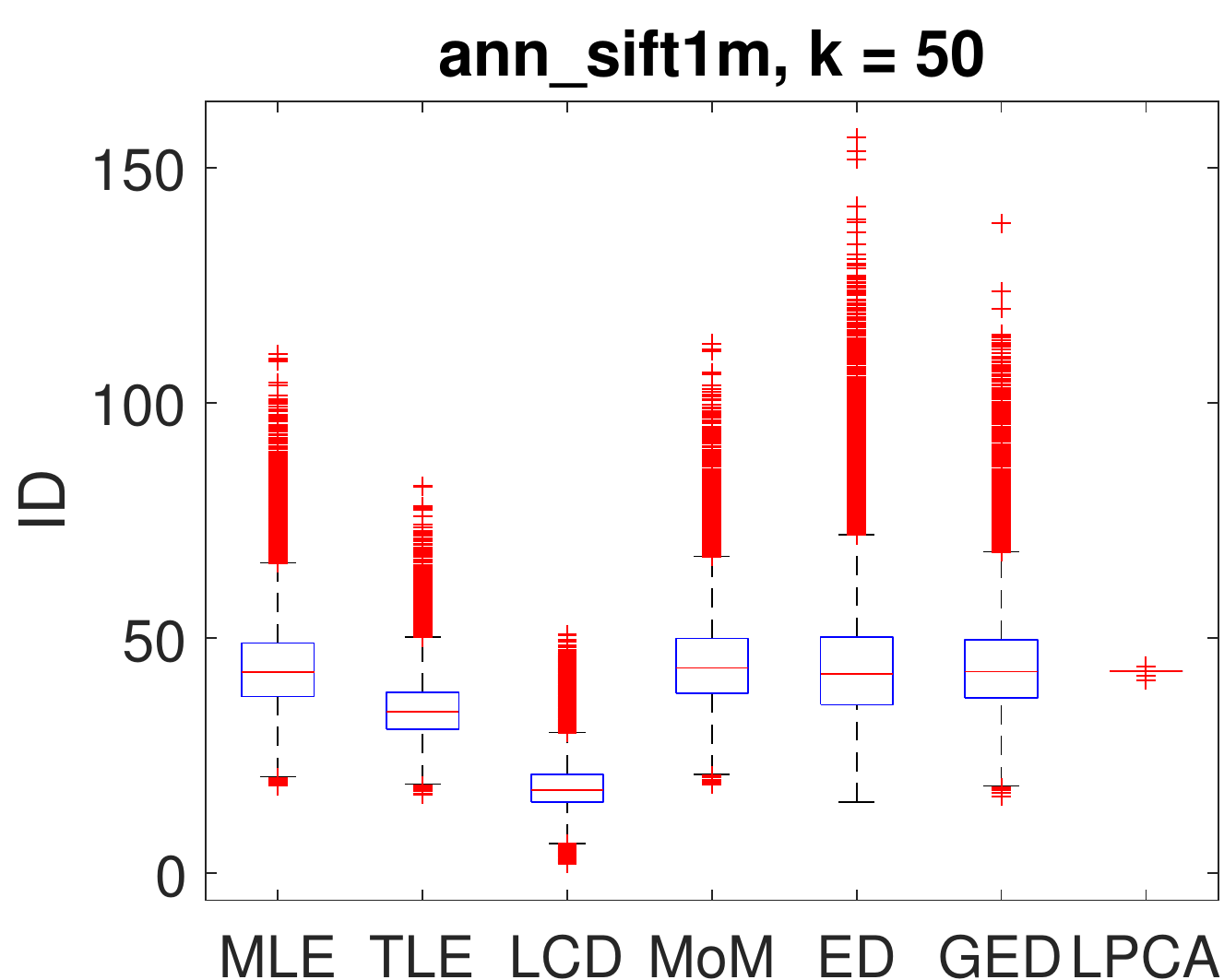}
\includegraphics[width=.32\textwidth]{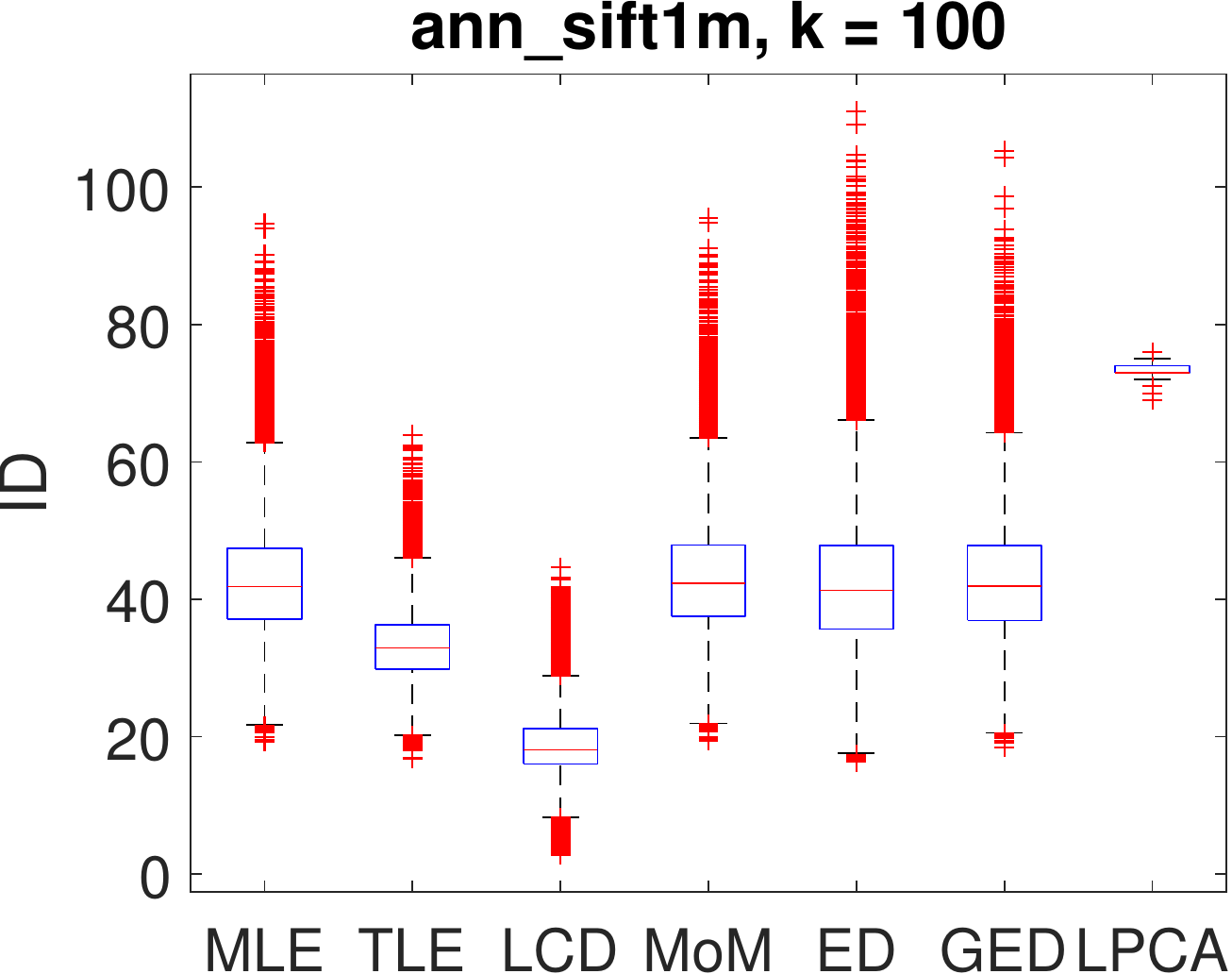}\\
\includegraphics[width=.32\textwidth]{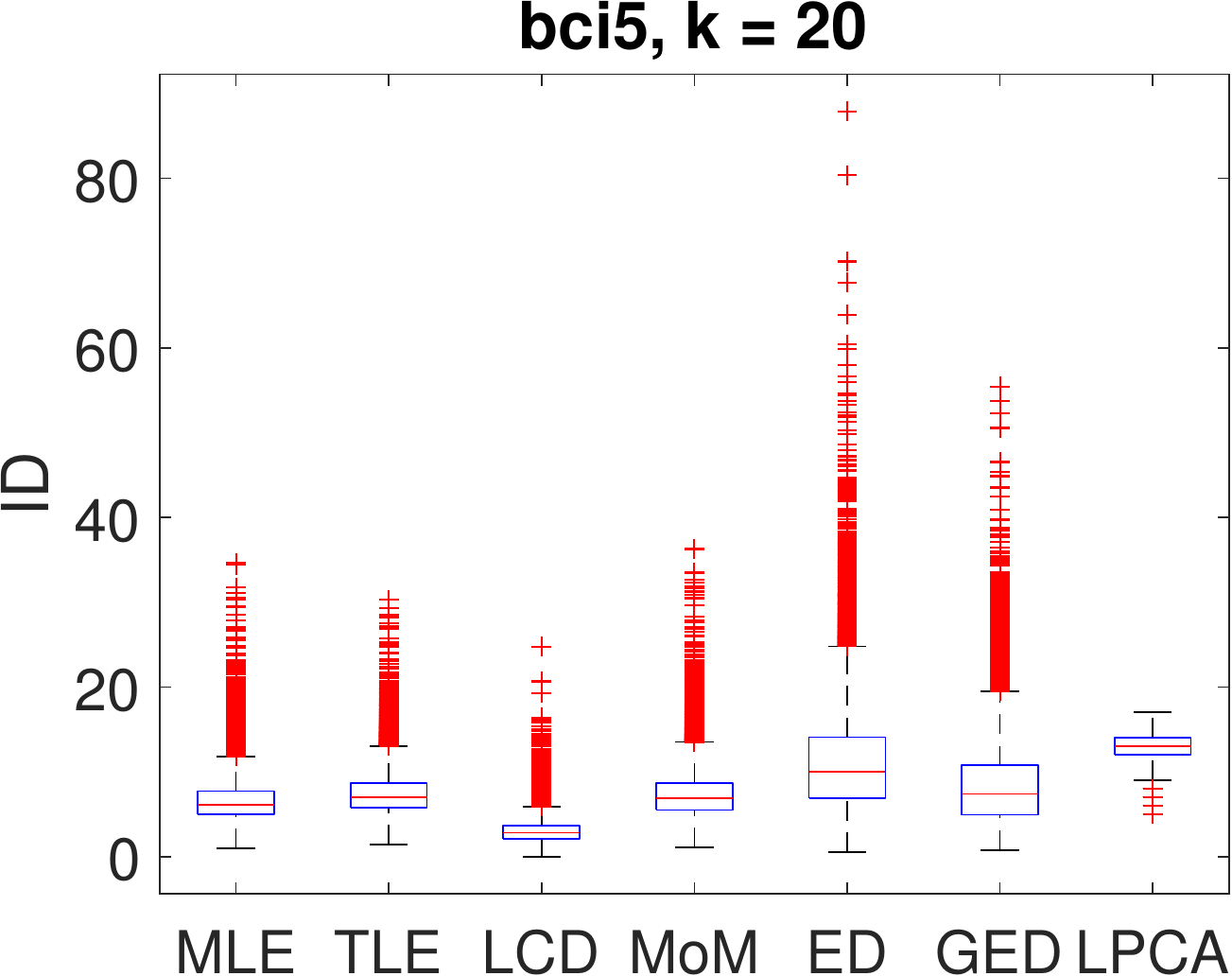}
\includegraphics[width=.32\textwidth]{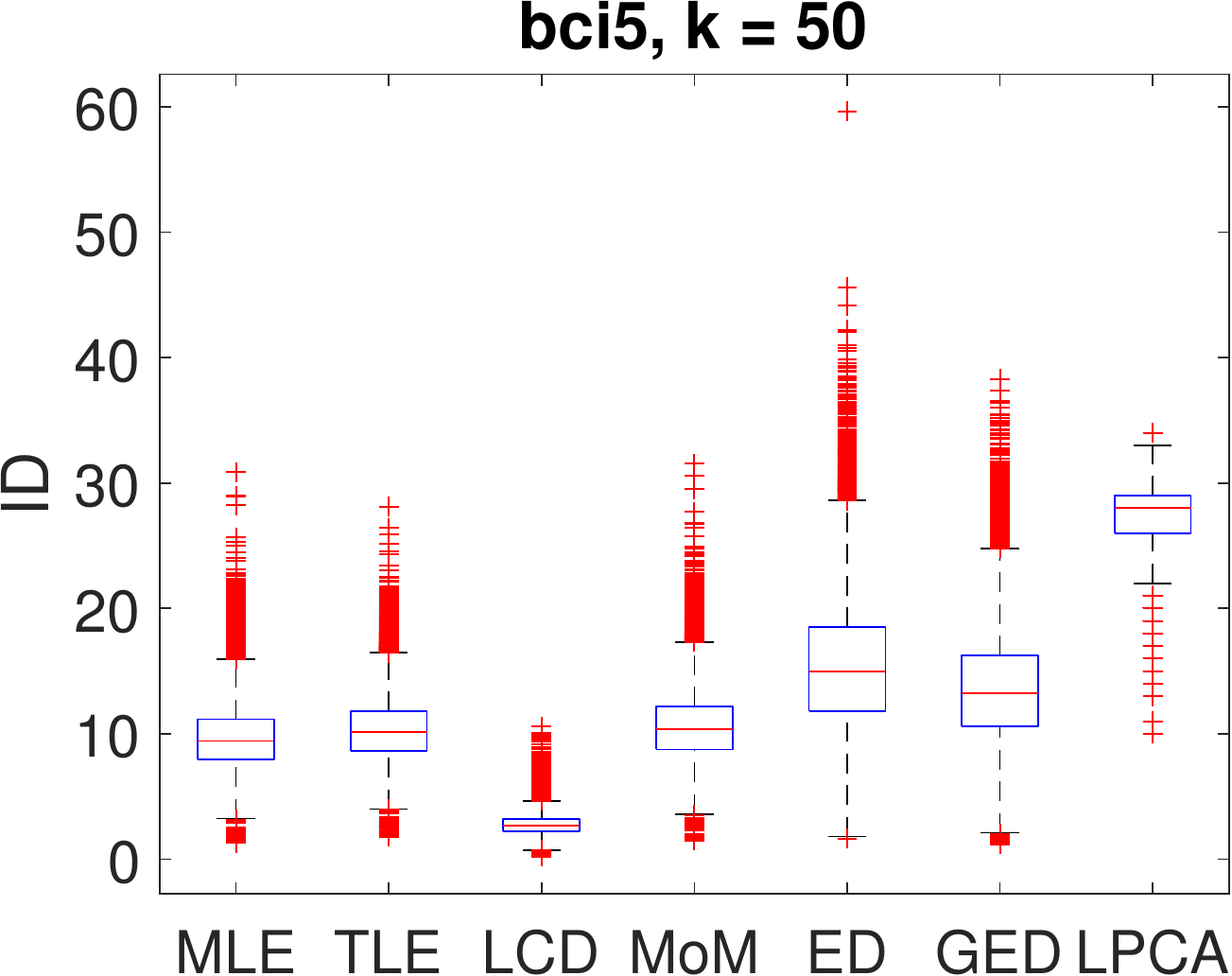}
\includegraphics[width=.32\textwidth]{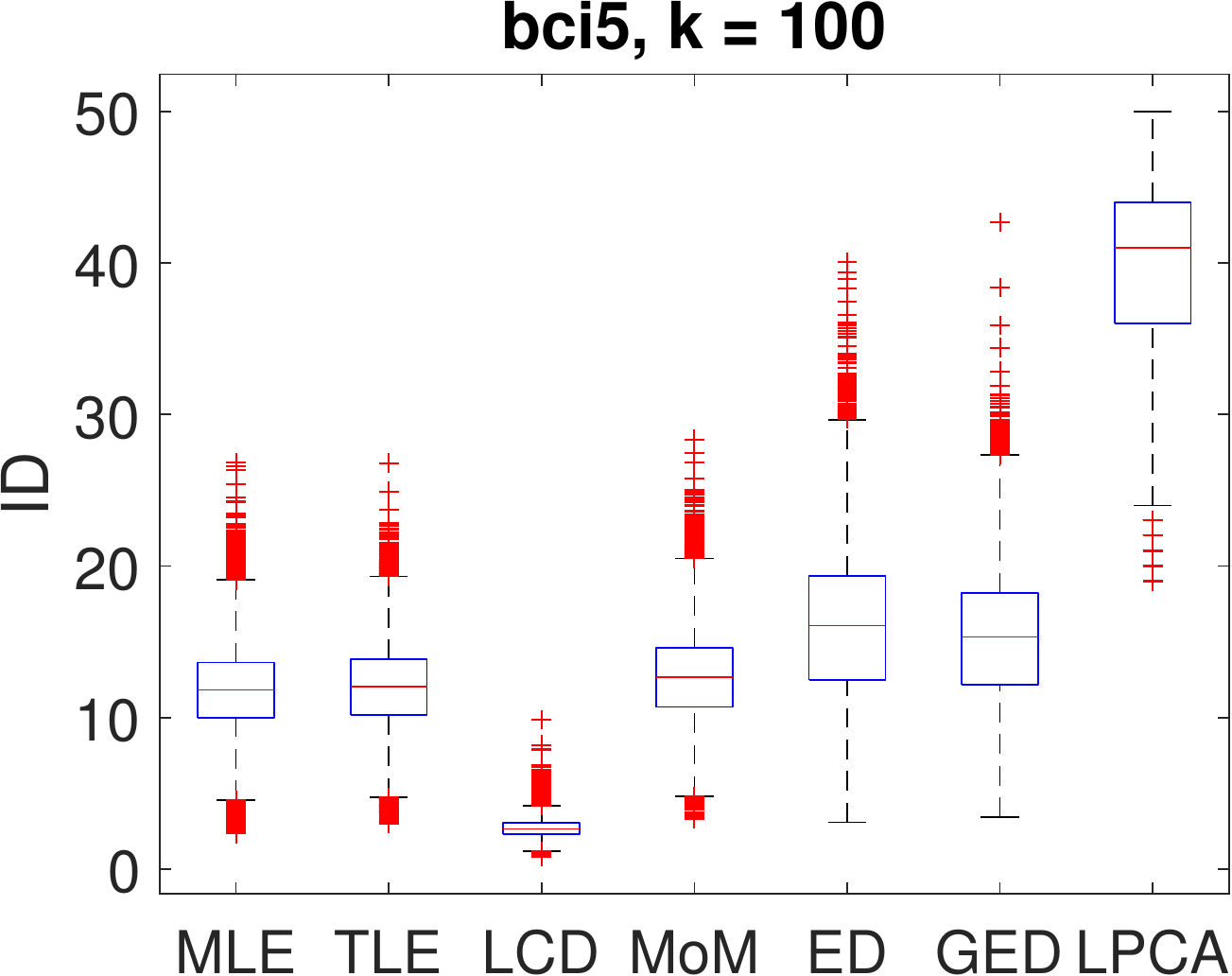}\\
\includegraphics[width=.32\textwidth]{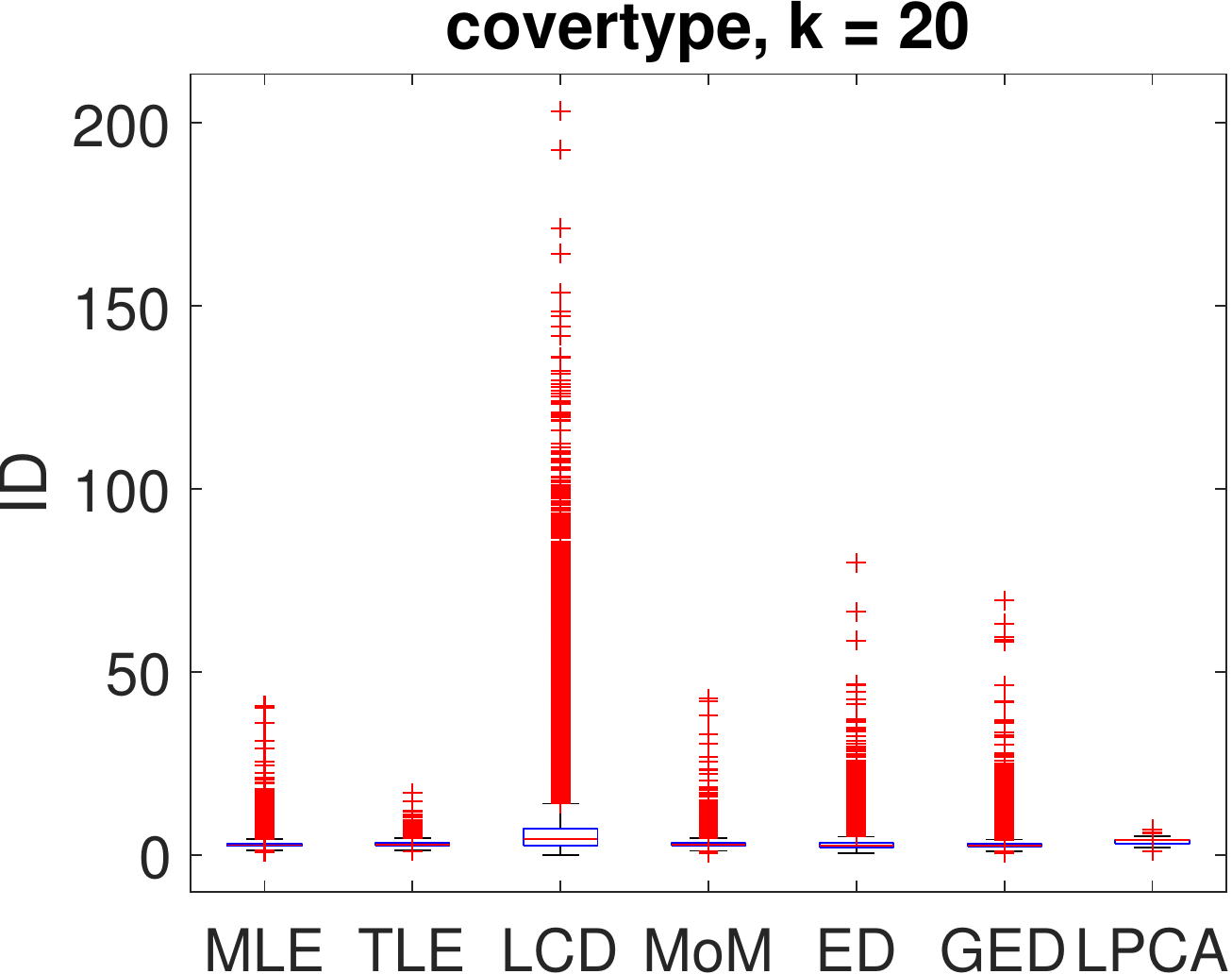}
\includegraphics[width=.32\textwidth]{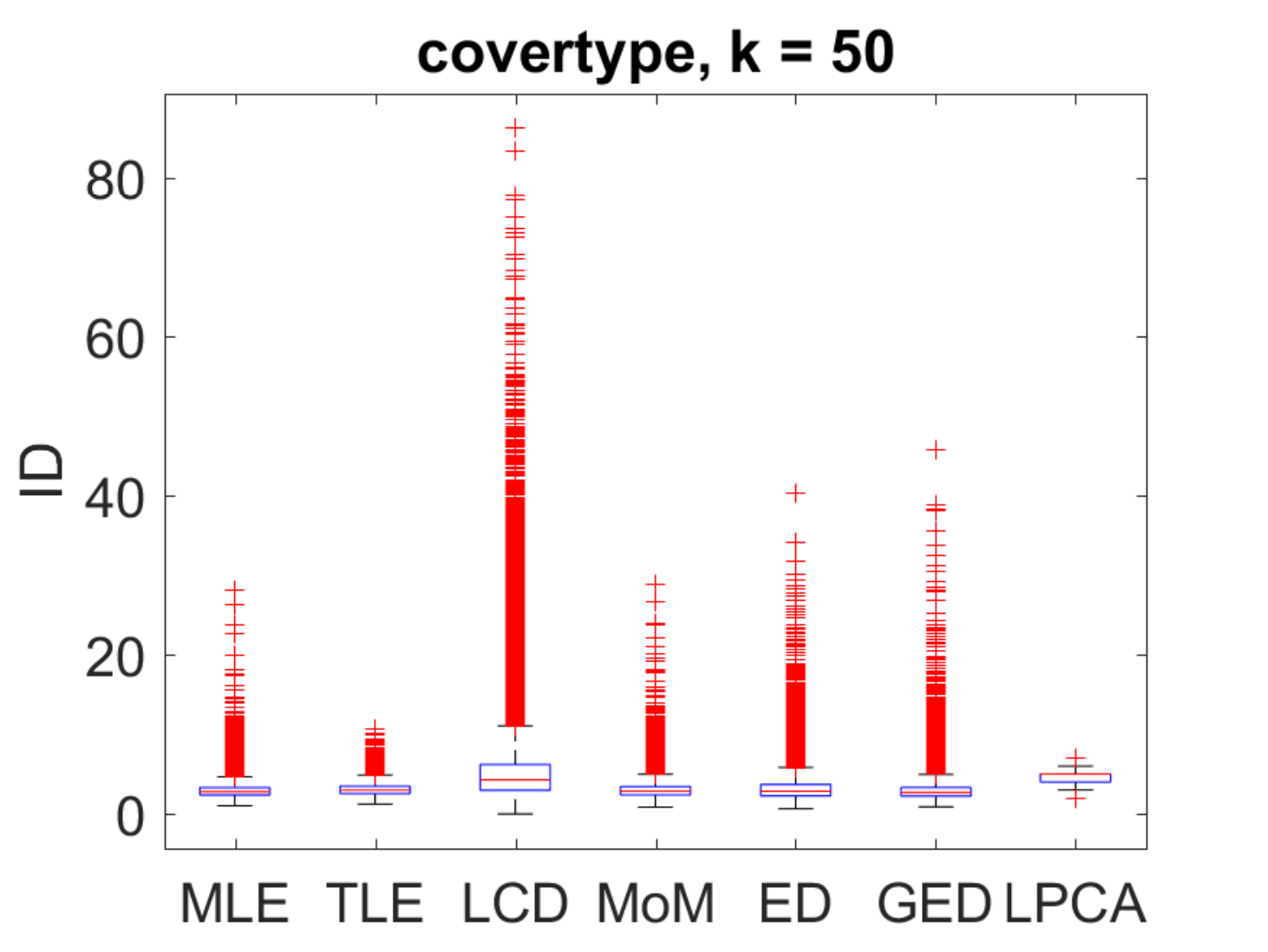}
\includegraphics[width=.32\textwidth]{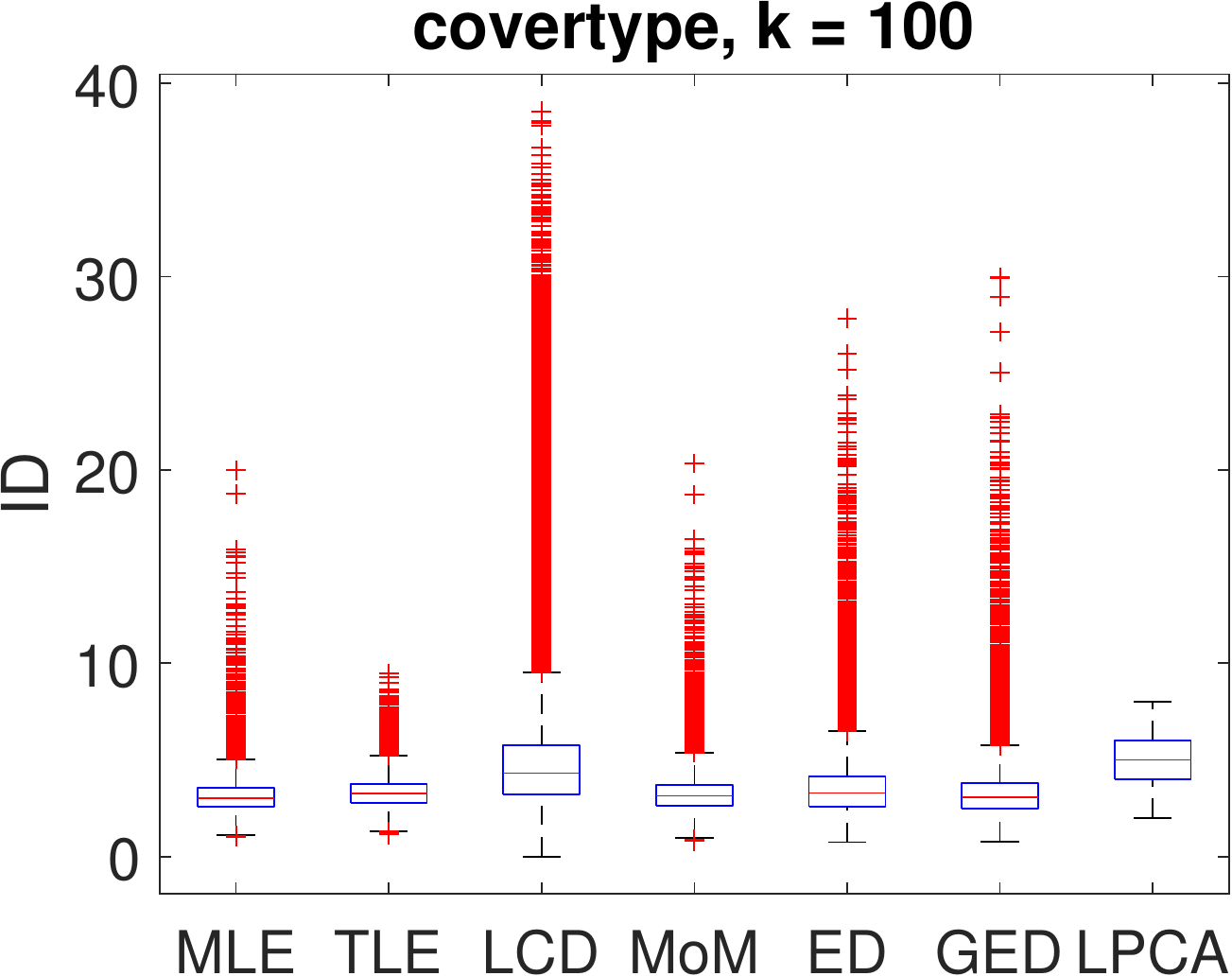}\\
\end{centering}
\caption{Box plots of estimated ID values, for neighborhood sizes 20, 50 and 100, on real data (part~1).}
\label{fig:real-boxplots-mutik-p1}
\end{figure*}

\begin{figure*}
\begin{centering}
\includegraphics[width=.32\textwidth]{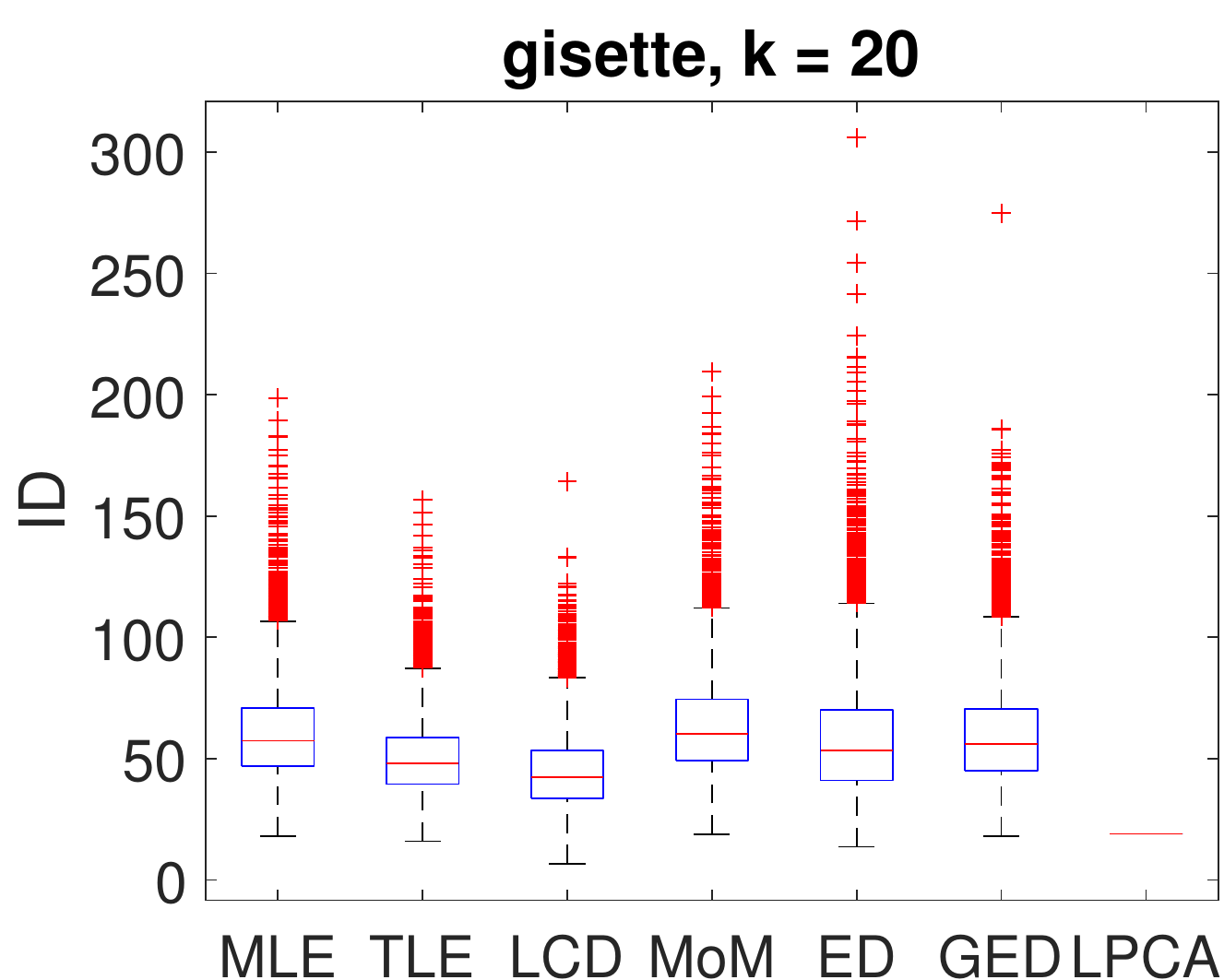}
\includegraphics[width=.32\textwidth]{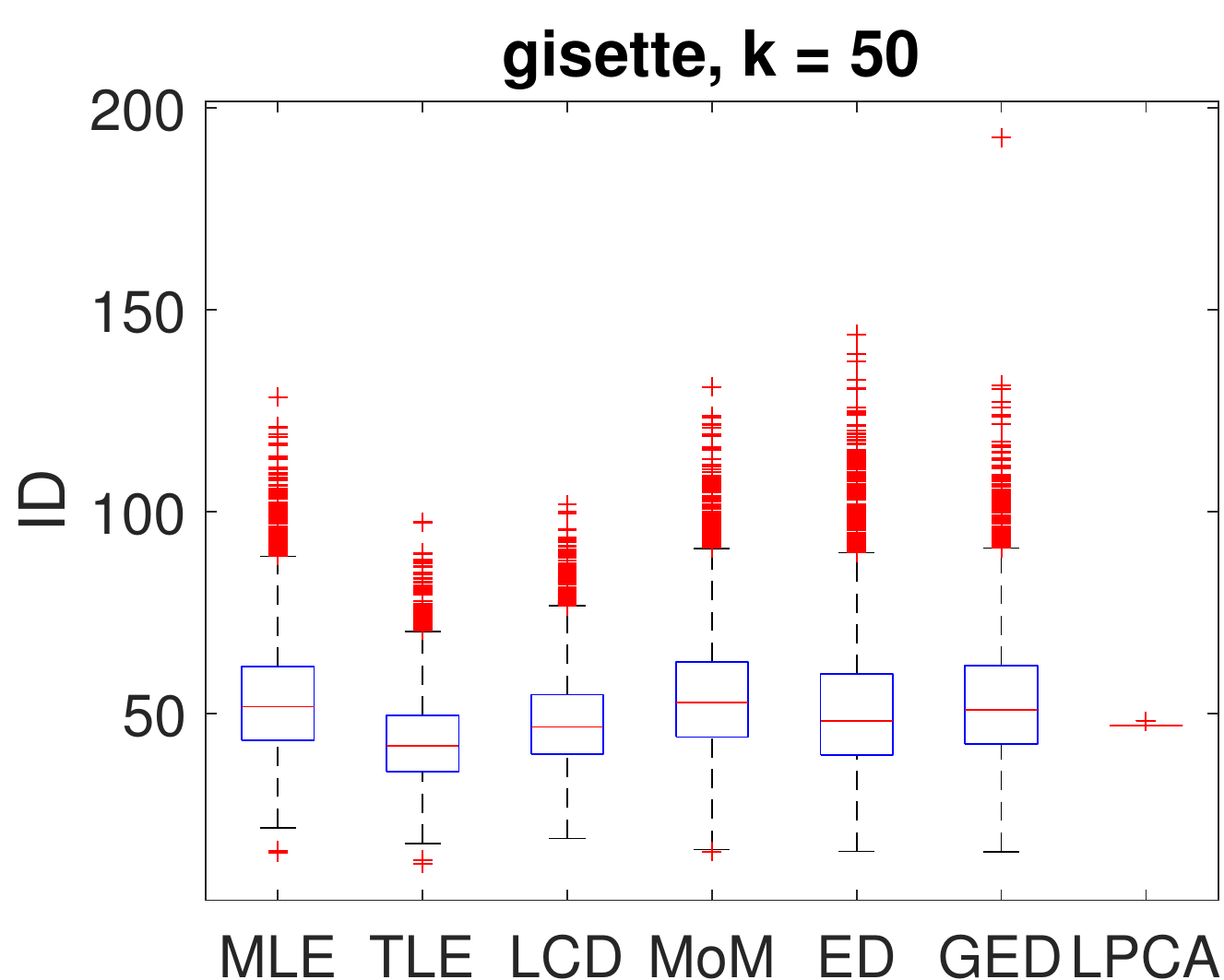}
\includegraphics[width=.32\textwidth]{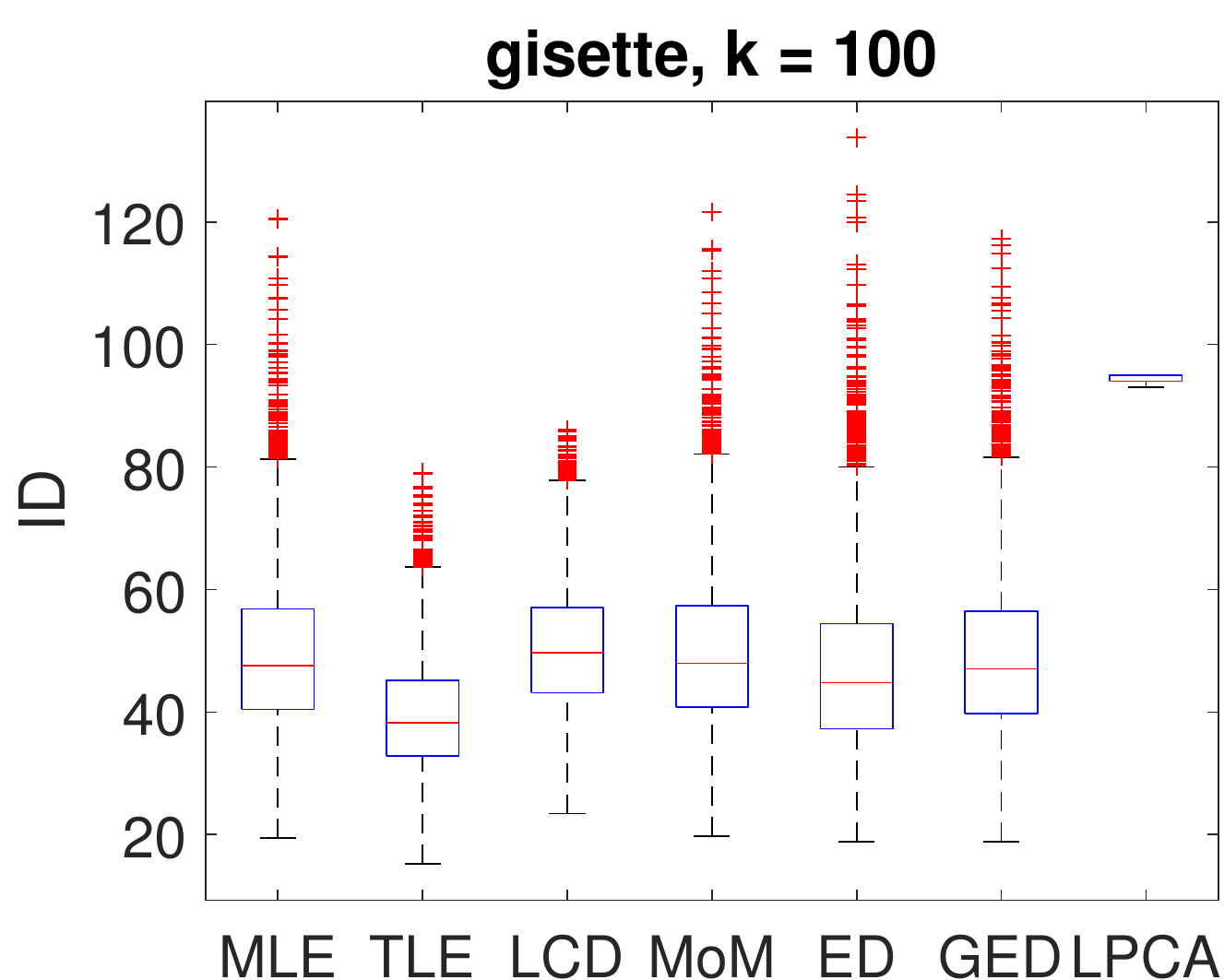}\\
\includegraphics[width=.32\textwidth]{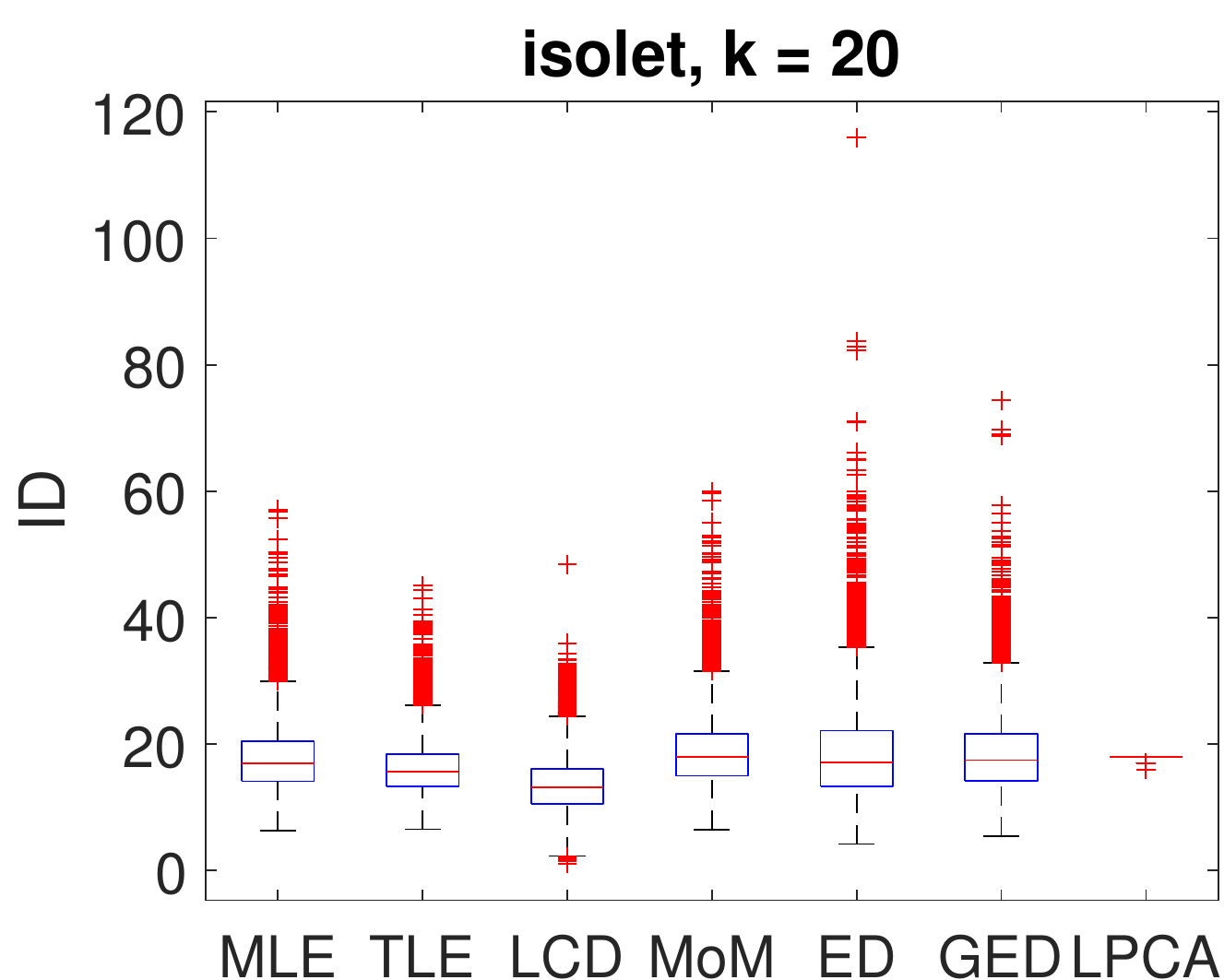}
\includegraphics[width=.32\textwidth]{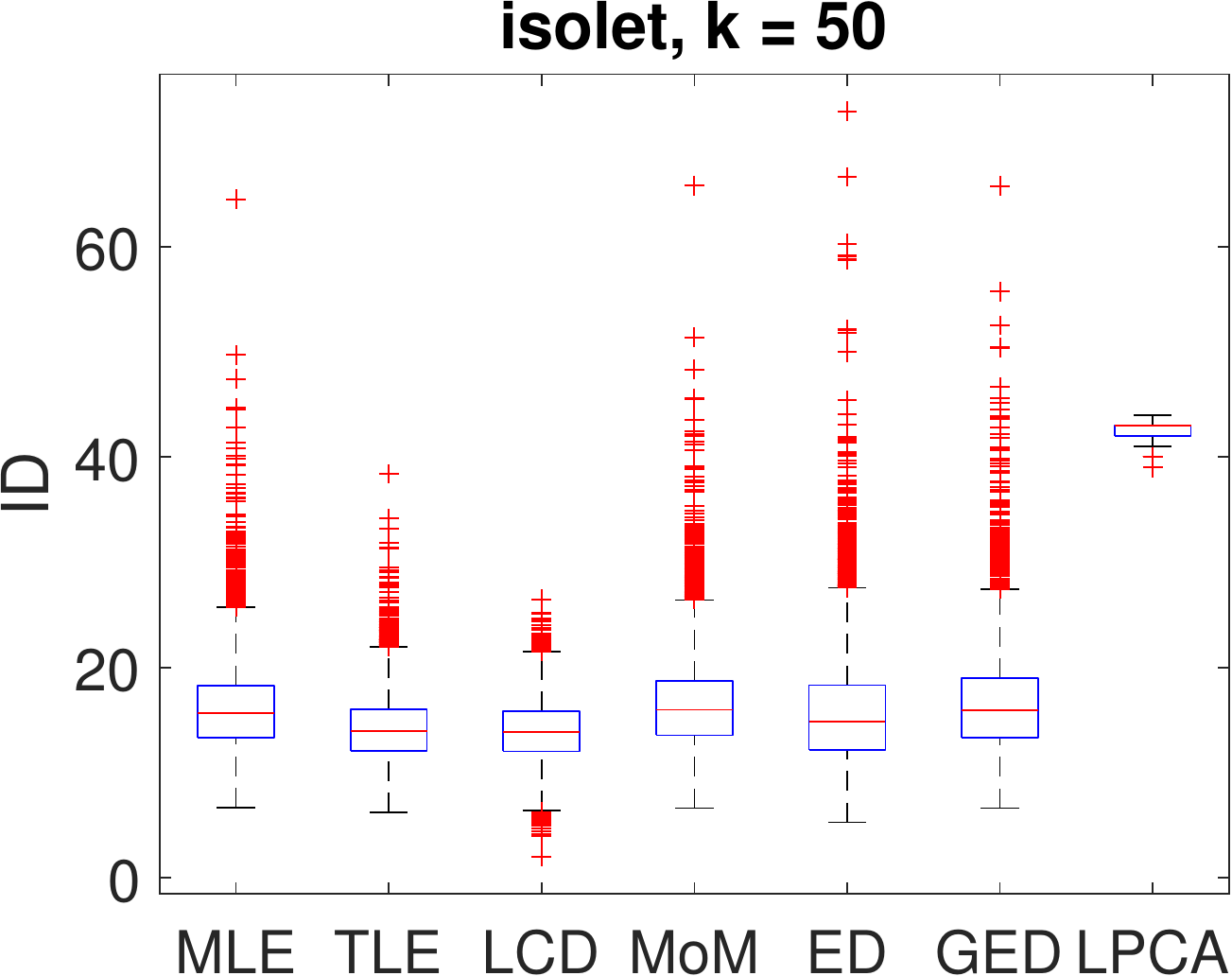}
\includegraphics[width=.32\textwidth]{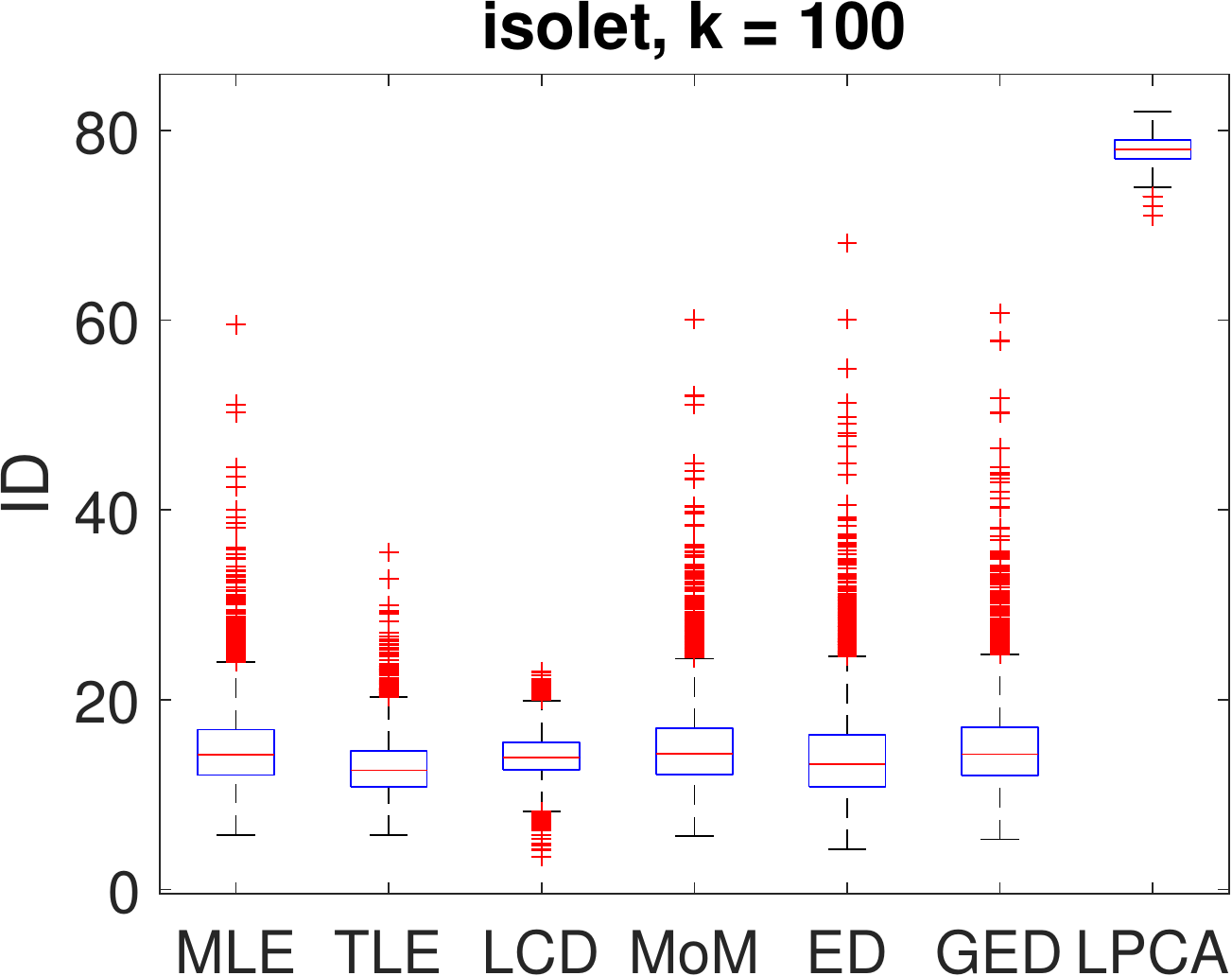}\\
\includegraphics[width=.32\textwidth]{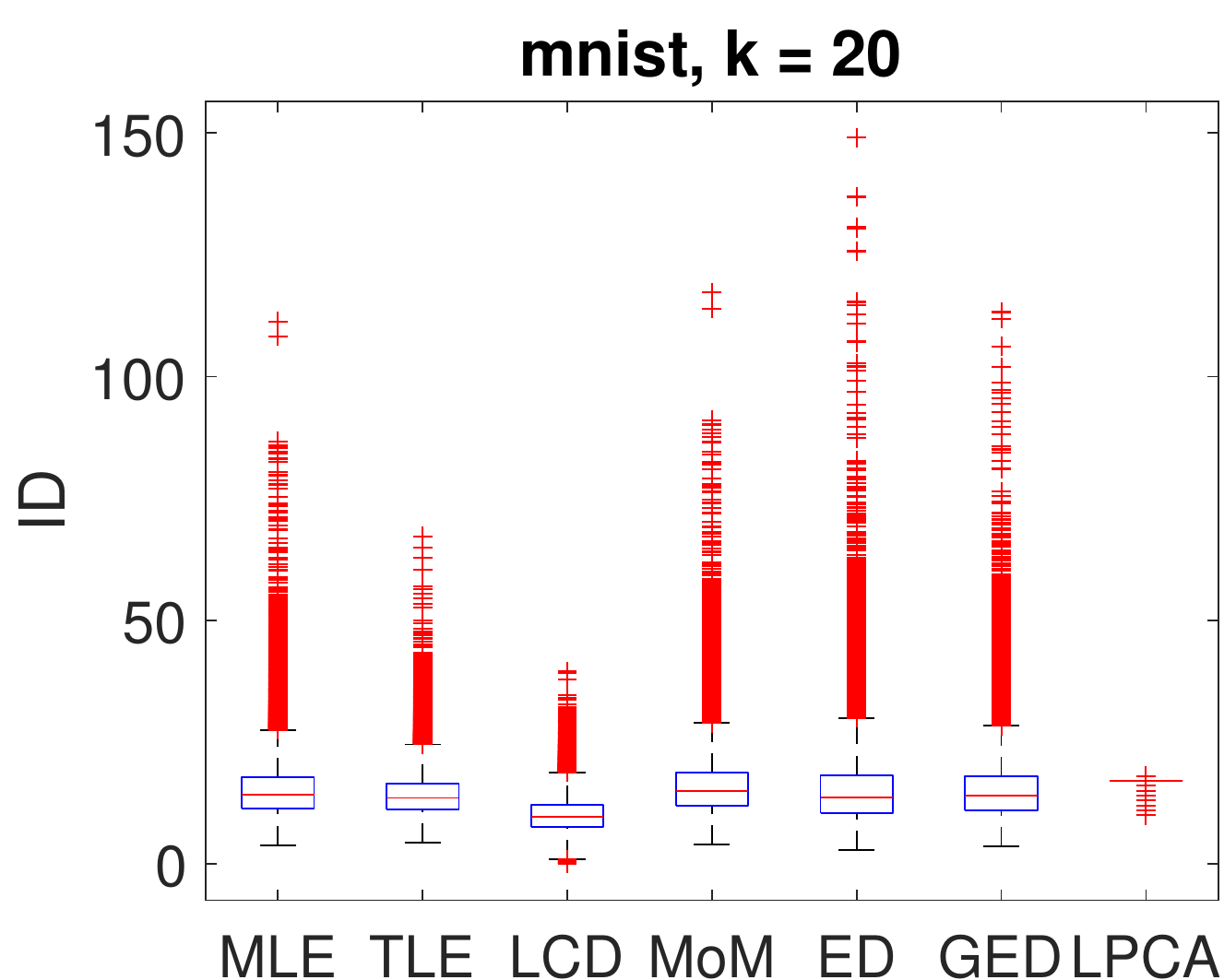}
\includegraphics[width=.32\textwidth]{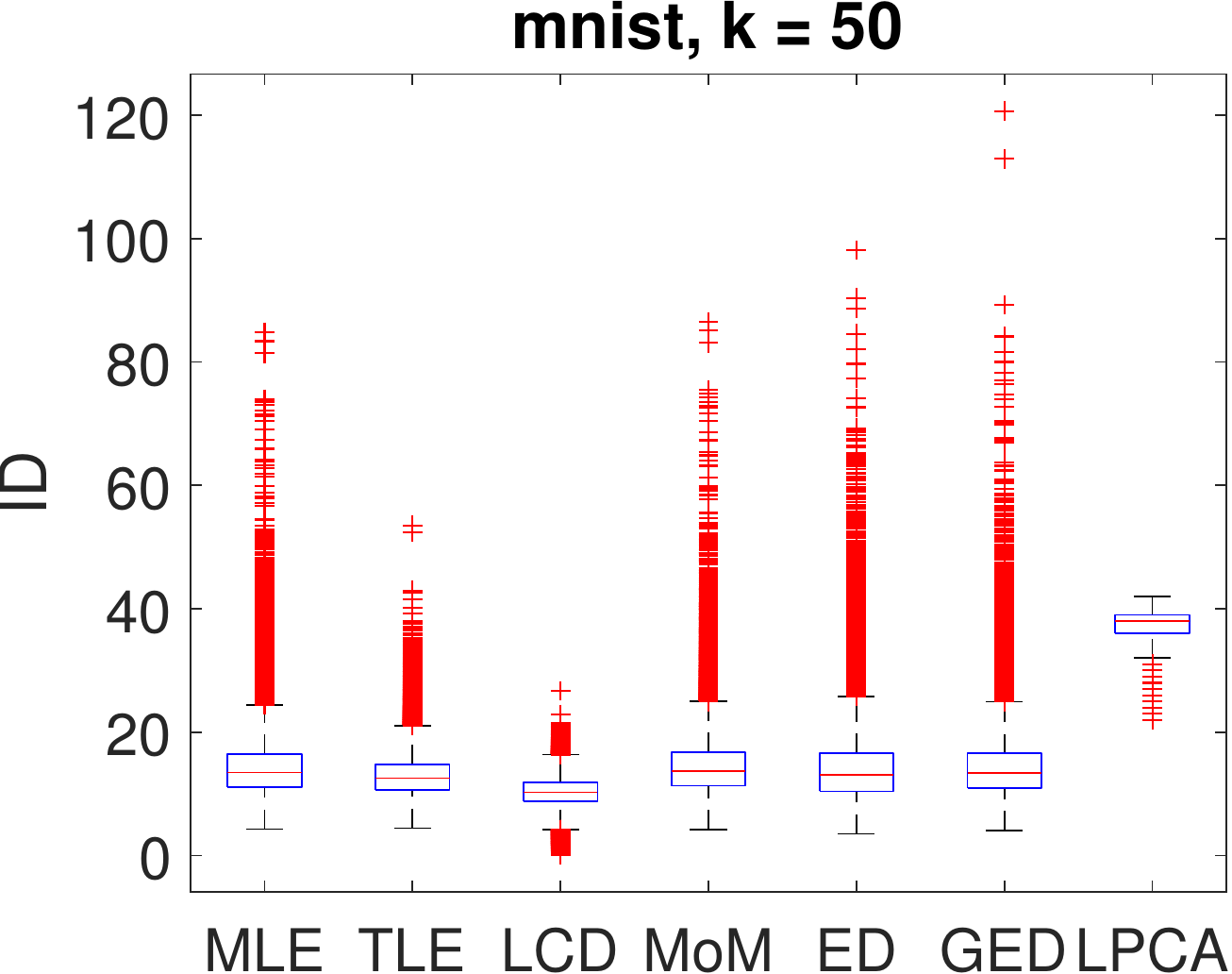}
\includegraphics[width=.32\textwidth]{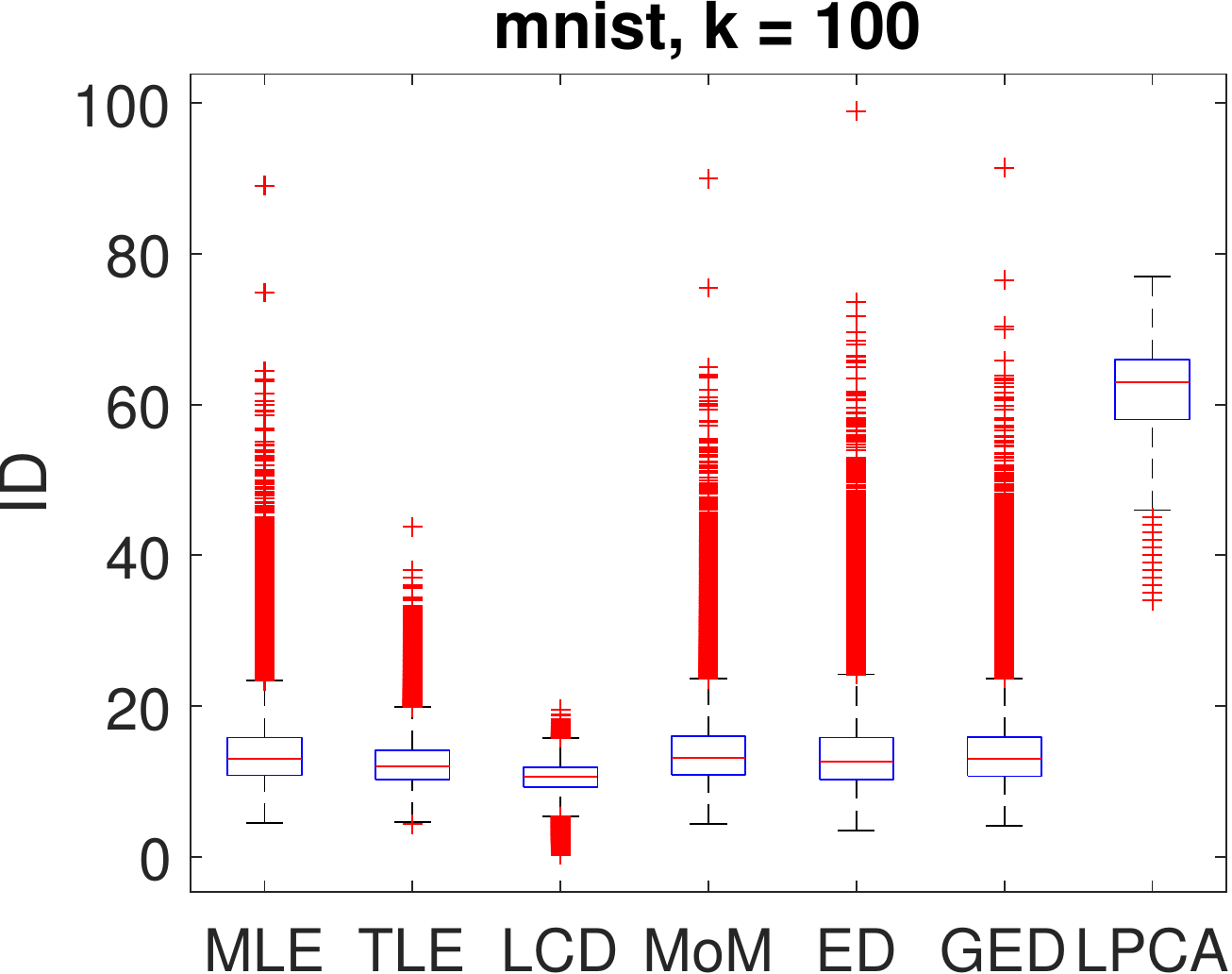}\\
\includegraphics[width=.32\textwidth]{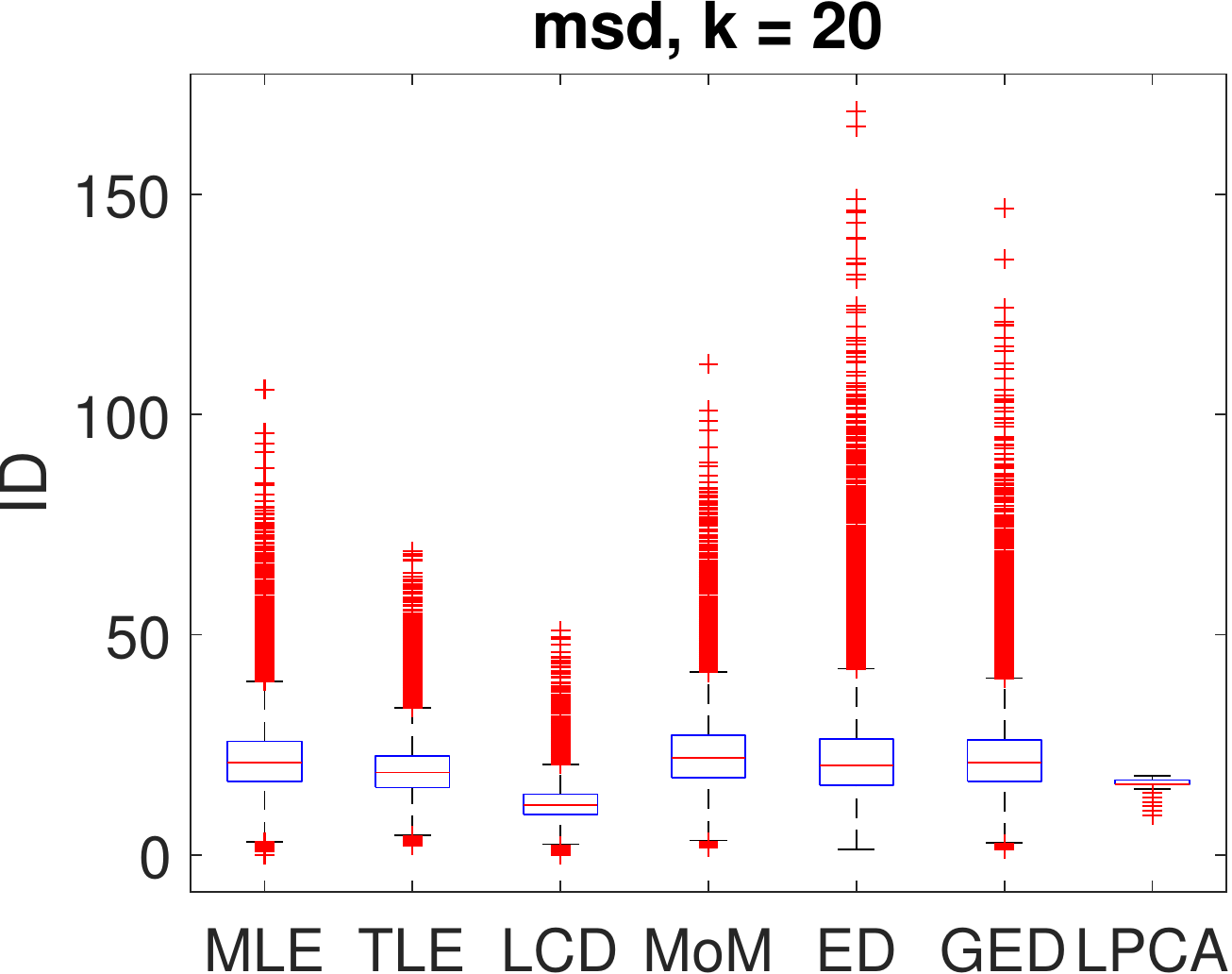}
\includegraphics[width=.32\textwidth]{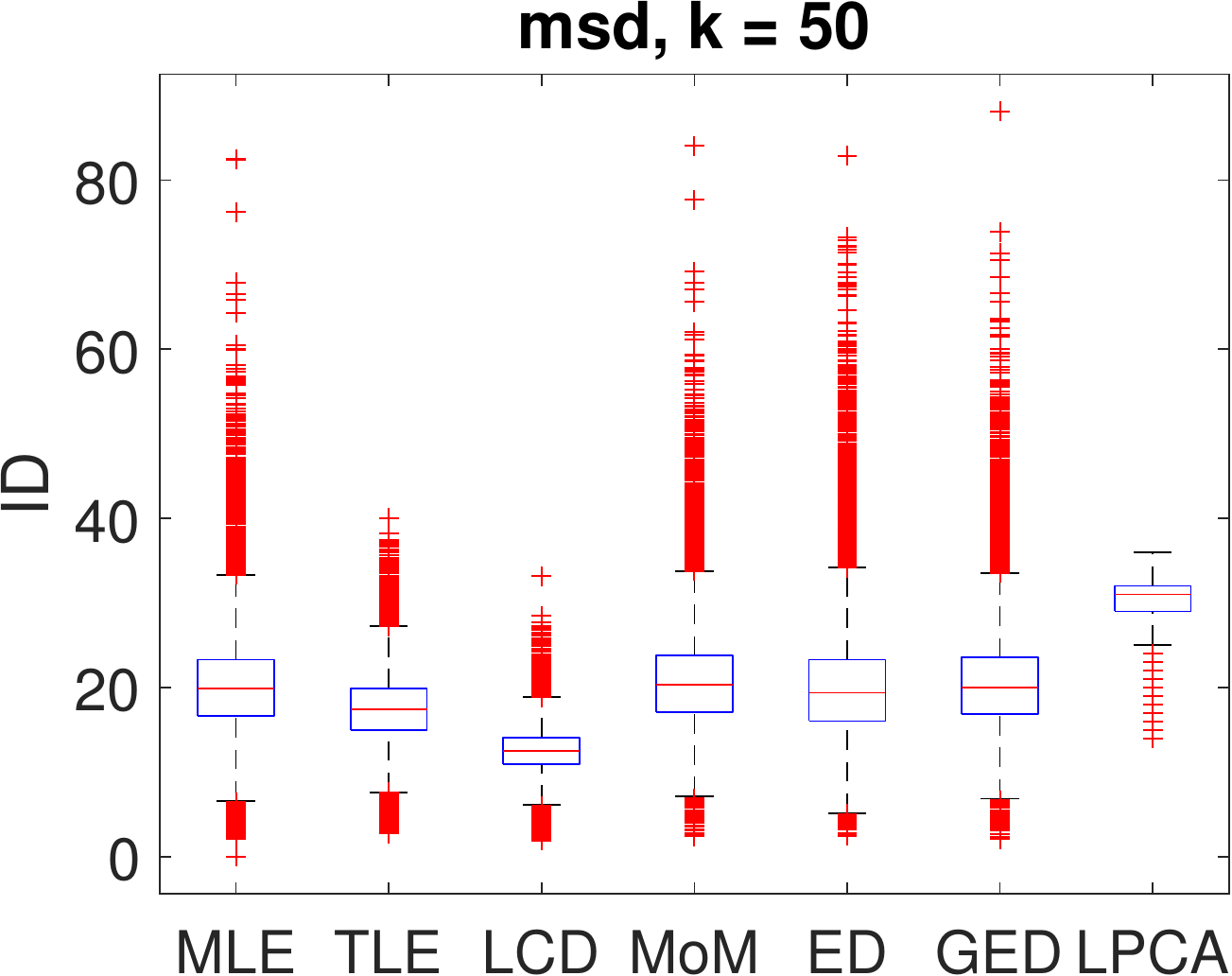}
\includegraphics[width=.32\textwidth]{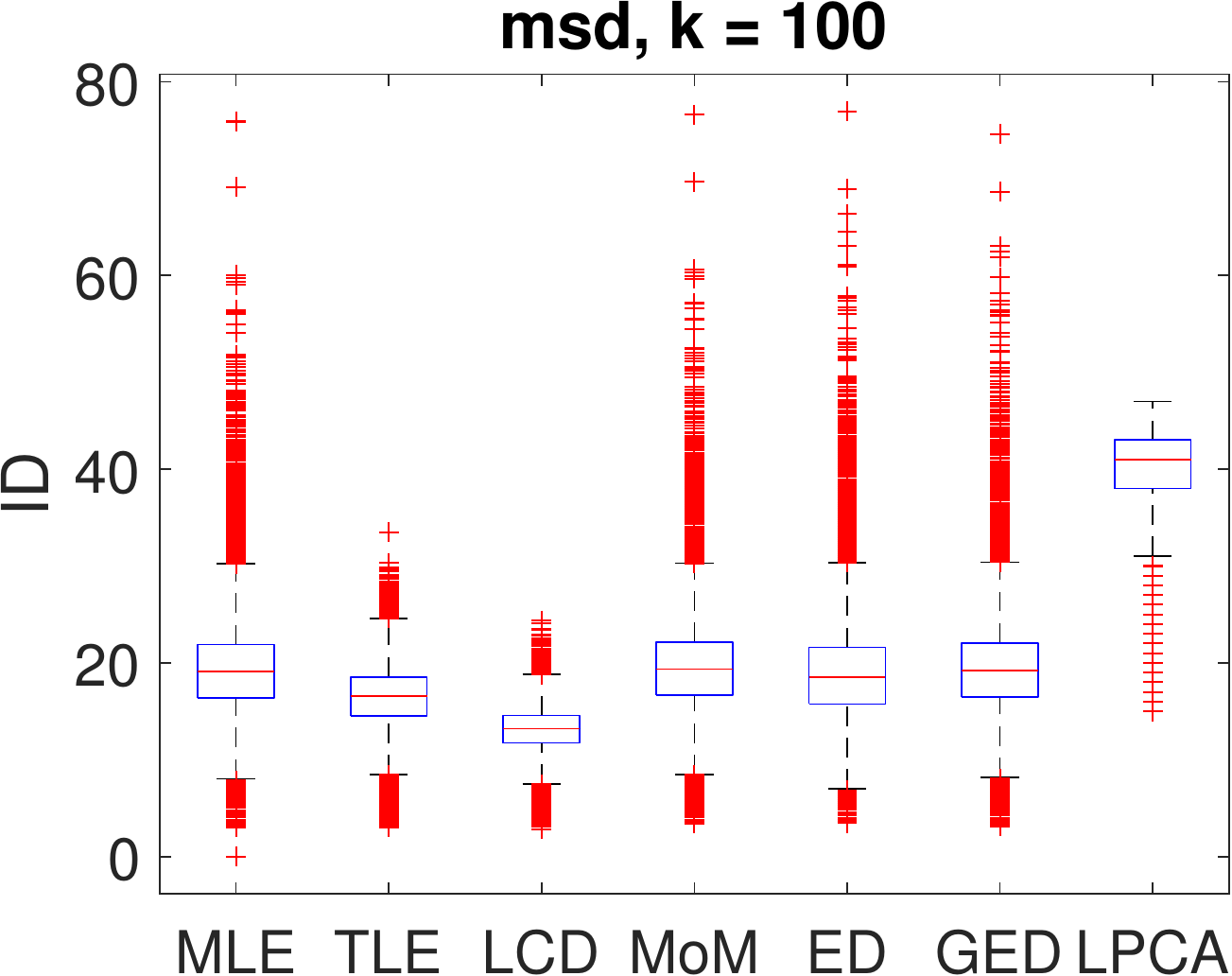}\\
\end{centering}
\caption{Box plots of estimated ID values, for neighborhood sizes 20, 50 and 100, on real data (part~2).}
\label{fig:real-boxplots-mutik-p2}
\end{figure*}

Figures~\ref{fig:real-boxplots-mutik-p1} and~\ref{fig:real-boxplots-mutik-p2}
show the distributions of ID estimates for real data sets, with three box
plots in a row for each data set, using neighborhood sizes $k = 20$, $50$, and $100$.
In addition to supporting the observations made on synthetic data sets, it
can be stressed that variance of TLE for $k = 20$ is usually as good as or
superior to the variance of other methods for larger values of~$k$. In the
case when some other method (such as LCD) exhibits smaller variance, it usually
has much worse bias, as we have also seen on synthetic data. An extreme
example of this behavior is LPCA, which on real data often exhibits tight
variance but positive bias that increases significantly with neighborhood
size~$k$.

%%%%%%%%%%%
\section{Conclusion}
\label{S:conclusion}

In models such as the Correlation Dimension,
pairwise distance measurements have been successfully used in order
to estimate global intrinsic dimensionality.
However, to the best of our knowledge,
none of the existing models of local intrinsic dimensionality
take advantage of distances other than those from a test point to
the members of its neighborhood.
Here we have shown that estimating the Correlation Dimension on small neighborhoods
does not lead to a correct ID estimation
if all available pairwise distances are used
without accounting for the clipping of data to the respective localities.

Our proposed estimation strategy makes use of
a subset of the available intra-neighborhood distances to achieve
faster convergence with fewer samples, and
can thus be used on applications in which the data consists of
many natural groups of small size.
There is also evidence of better robustness when estimating the intrinsic dimensionality over neighborhoods based on outlier data.
Moreover, it has a smaller bias and variance than state-of-the-art estimators,
especially on nonlinear subspaces.
Consequently,
the estimator can achieve more accurate ID estimates
within a smaller locality than the traditional estimators.
This has the potential to improve the quality
of algorithms where locality is an important factor,
such as subspace clustering and subspace outlier detection.
\bibliographystyle{plain}
%\small
\bibliography{references,milos-bib}

\end{document}